\documentclass[letterpaper, 10 pt, conference]{ieeeconf}
\IEEEoverridecommandlockouts                              
\usepackage{amsmath,amssymb}
\usepackage{bm} 
\usepackage{comment}
\usepackage{url}

\usepackage{multirow}
\usepackage{graphicx}
\usepackage{epsfig}
\usepackage{times}
\usepackage{mathptmx}
\usepackage{cite}
\usepackage{color}
\usepackage{algorithm}
\usepackage{algpseudocode}

\usepackage{times}

\usepackage{booktabs}  
\usepackage{multirow}  

\definecolor{red}{rgb}{1,0,0}
\definecolor{green}{rgb}{0,1,0}
\definecolor{blue}{rgb}{0,0,1}
\definecolor{violet}{rgb}{1,0,1}
\definecolor{cyan}{cmyk}{1,0,0,0}
\definecolor{magenta}{cmyk}{0,1,0,0}
\definecolor{yellow}{cmyk}{0,0,1,0}

\definecolor{white}{rgb}{1,1,1}

\usepackage{arydshln}

\newcommand{\CO}[1]{}

\newcommand{\CommentOut}[1]{}

\newcommand{\FIG}[3]{
\begin{minipage}[b]{#1cm}
\begin{center}
\includegraphics[width=#1cm]{#2}\\
{\scriptsize #3}
\end{center}
\end{minipage}
}

 \newcommand{\editage}[1]{}

\twocolumn

\begin{document}

\title{\LARGE\bf%
FlatVPR: Plug-and-play Geo-linear Residual Adapter for Geometric Rectification of Foundation Model Feature Manifolds
}

\author{Rai Hisada \and Kanji Tanaka
\thanks{Our work has been supported in part by JSPS KAKENHI Grant-in-Aid for Scientific Research (C) 20K12008 and 23K11270.}
\thanks{%
The authors are with Fundamental Engineering for Knowledge-Based Society, Graduate School of Engineering, University of Fukui, Japan. 
{\tt\small{tnkknj@g.u-fukui.ac.jp}}
}}

\maketitle

\begin{abstract}
This paper proposes ``FlatVPR,'' a novel geometric rectification paradigm that effectively bridges the trade-off between map lightweightness and localization accuracy in visual place recognition (VPR) by enforcing a feature manifold structure where any descriptor between two adjacent anchors $\mathbf{z}_A$ and $\mathbf{z}_B$ can be accurately reconstructed via linear interpolation $\hat{\mathbf{z}}_{pseudo} = (1-t)\mathbf{z}_A + t\mathbf{z}_B$, where $t \in [0,1]$ denotes the relative position. While state-of-the-art foundation models such as DINOv2-ViT-S/14 provide robust semantic features, their latent manifolds exhibit prominent curvature, projecting uniform linear motion in physical space onto highly non-linear trajectories in the feature space, which hinders reliable reconstruction under sparse anchor conditions. To enable the aforementioned interpolation-based reconstruction, we introduce a residual transformation $\hat{\mathbf{z}} = \mathbf{z} + \text{Res}(\mathbf{z})$ to the raw foundation features $\mathbf{z}$, where $\text{Res}(\cdot)$ represents a learnable adapter. Our method explicitly suppresses manifold curvature using a mathematically grounded Pullback Flatness Loss that minimizes the deviation of intermediate features from the linear segment connecting adjacent anchors, thereby minimizing the intrinsic curvature of the manifold. Through this spatial flattening, map construction is formulated within an Expectation-Maximization (EM) framework, decoupled into a continuous M-step for manifold adaptation and a conceptual E-step for optimal anchor selection guidelines. Experiments on the NCLT dataset demonstrate that the application of our adapter leads to significant performance improvements even under extremely sparse anchor conditions with 100m intervals and extreme seasonal changes, improving the overall average MRR from 0.697 to 0.872, while achieving a high maximum MRR of 0.991 in specific challenging pairs.
\end{abstract}


\begin{table*}[t]
\centering
\caption{Main quantitative evaluation on the NCLT dataset across 110 cross-season evaluation pairs (trained on single spring session 2012-04-29, evaluated with seed=0 at 500 epochs). We report the mean Reciprocal Rank (MRR) and Recall@$K$ ($R@1, 5, 10$) averaged over all query seasons for each target Database (DB) season. Bold values indicate the best performance.}
\label{tab:main_performance_nclt}
\small
\begin{tabular}{llcccc}
\toprule
\textbf{Target DB Season} & \textbf{Method} & \textbf{MRR $\uparrow$} & \textbf{Recall@1 (\%) $\uparrow$} & \textbf{Recall@5 (\%) $\uparrow$} & \textbf{Recall@10 (\%) $\uparrow$} \\
\midrule
2012-01-15
& DINOv2 Baseline & 0.4439 & 30.64 & 57.64 & 68.42 \\
& \textbf{FlatVPR (Ours)} & \textbf{0.8415} & \textbf{79.74} & \textbf{88.74} & \textbf{91.22} \\
\midrule
2012-02-19
& DINOv2 Baseline & 0.7410 & 65.50 & 81.36 & 87.26 \\
& \textbf{FlatVPR (Ours)} & \textbf{0.8480} & \textbf{78.50} & \textbf{90.30} & \textbf{93.30} \\
\midrule
2012-03-31
& DINOv2 Baseline & 0.8143 & 75.82 & 87.12 & 90.16 \\
& \textbf{FlatVPR (Ours)} & \textbf{0.9169} & \textbf{87.94} & \textbf{95.34} & \textbf{97.10} \\
\midrule
2012-08-04
& DINOv2 Baseline & 0.7302 & 66.52 & 78.46 & 81.94 \\
& \textbf{FlatVPR (Ours)} & \textbf{0.8690} & \textbf{81.24} & \textbf{91.90} & \textbf{94.70} \\
\midrule
2012-08-20
& DINOv2 Baseline & 0.5105 & 40.50 & 58.00 & 66.86 \\
& \textbf{FlatVPR (Ours)} & \textbf{0.7836} & \textbf{72.04} & \textbf{84.44} & \textbf{87.72} \\
\midrule
2012-10-28
& DINOv2 Baseline & 0.7397 & 66.90 & 81.18 & 84.42 \\
& \textbf{FlatVPR (Ours)} & \textbf{0.8980} & \textbf{85.48} & \textbf{94.04} & \textbf{96.38} \\
\midrule
2012-11-04
& DINOv2 Baseline & 0.8427 & 78.20 & 89.92 & 93.34 \\
& \textbf{FlatVPR (Ours)} & \textbf{0.9575} & \textbf{93.70} & \textbf{97.80} & \textbf{98.78} \\
\midrule
2012-11-16
& DINOv2 Baseline & 0.5407 & 43.14 & 62.46 & 71.30 \\
& \textbf{FlatVPR (Ours)} & \textbf{0.8207} & \textbf{76.44} & \textbf{87.52} & \textbf{90.30} \\
\midrule
2012-12-01
& DINOv2 Baseline & 0.7570 & 69.96 & 81.12 & 84.72 \\
& \textbf{FlatVPR (Ours)} & \textbf{0.9272} & \textbf{89.84} & \textbf{95.74} & \textbf{96.88} \\
\midrule
2013-02-23
& DINOv2 Baseline & 0.7719 & 71.46 & 82.26 & 84.62 \\
& \textbf{FlatVPR (Ours)} & \textbf{0.9169} & \textbf{87.94} & \textbf{95.34} & \textbf{97.10} \\
\midrule
2013-04-05
& DINOv2 Baseline & 0.7788 & 72.06 & 82.68 & 85.50 \\
& \textbf{FlatVPR (Ours)} & \textbf{0.9304} & \textbf{89.92} & \textbf{95.84} & \textbf{96.88} \\
\midrule[0.8pt]
\textbf{Overall Average}
& DINOv2 Baseline & 0.6974 & 61.50 & 76.51 & 81.82 \\
& \textbf{FlatVPR (Ours)} & \textbf{0.8718} & \textbf{82.98} & \textbf{92.86} & \textbf{94.94} \\
\bottomrule
\end{tabular}
\end{table*}

\begin{figure*}[t]
  \centering
  \includegraphics[width=15cm]{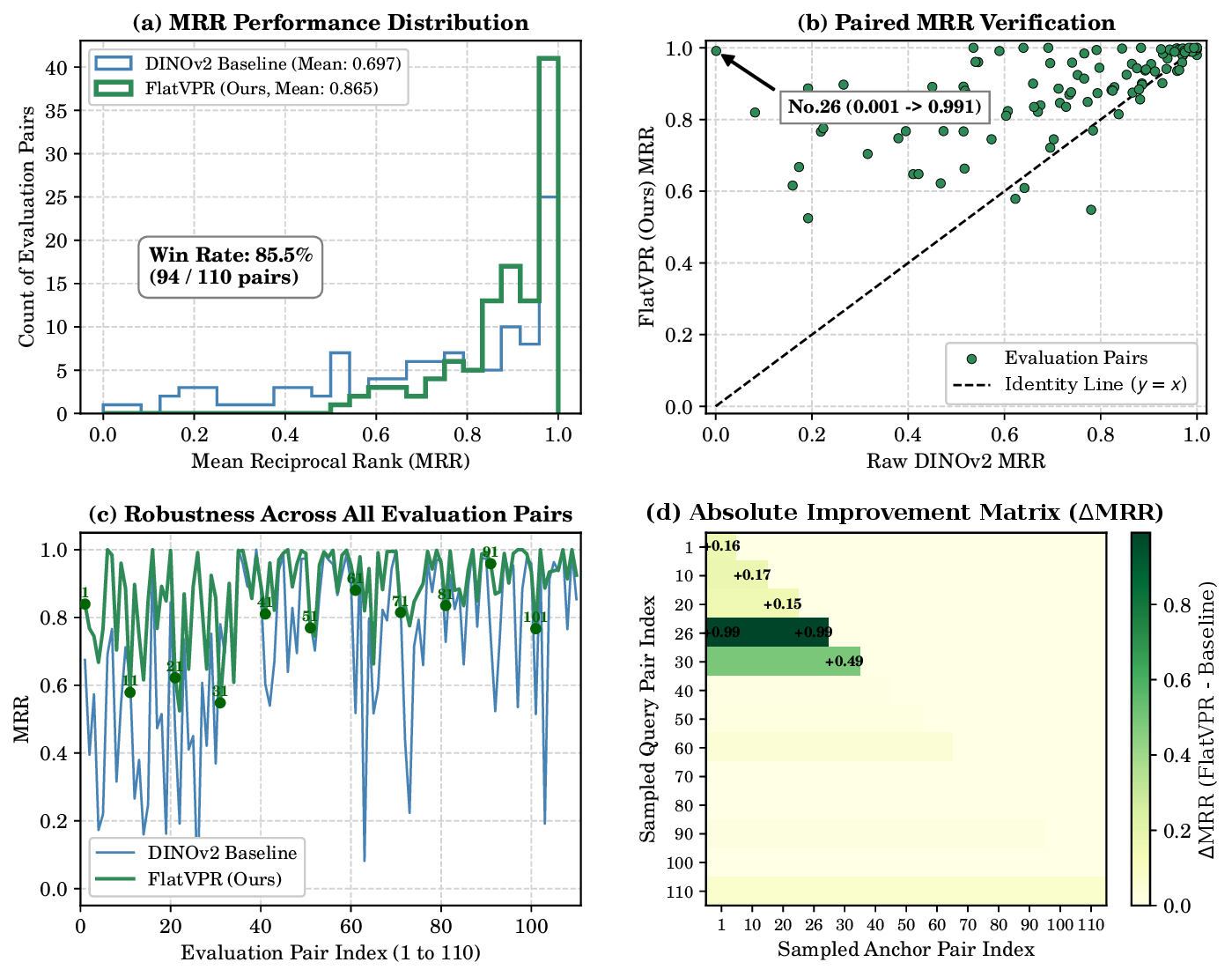}\vspace*{2.5cm}\\
  \caption{Quantitative evaluation across 110 cross-season pairs on the NCLT dataset: (a) MRR performance distribution compared to the DINOv2 baseline, (b) paired MRR verification highlighting the failure-to-success recovery (No. 26), (c) general performance trend across all evaluation sequences, and (d) absolute improvement matrix ($\Delta$MRR) demonstrating universal geometric priors.}
  \label{fig:nclt_evaluation_combined}
\end{figure*}

\section{Introduction}

Visual Place Recognition (VPR) is an indispensable capability for autonomous robots \cite{berton2023deep,cadena2016past}; however, managing large-scale maps remains a critical challenge due to burgeoning requirements for storage capacity and memory bandwidth. While state-of-the-art \textbf{pre-trained foundation models} like DINOv2-ViT-S/14\cite{oquab2023dinov2} offer descriptors with robust resilience to environmental changes, their high dimensionality and the necessity to retain descriptors for every frame to maintain map fidelity pose a severe bottleneck.

The objective of this work is to enforce a feature manifold structure where the descriptor of any point lying between two adjacent anchors, $\hat{\mathbf{z}}_A$ and $\hat{\mathbf{z}}_B$, can be accurately reconstructed via linear interpolation:
\begin{equation}
\hat{\mathbf{z}}_{pseudo} = (1-t)\hat{\mathbf{z}}_A + t \hat{\mathbf{z}}_B
\end{equation}
where $t$ denotes the relative position between the two points. Satisfying this ``linear reconstruction specification'' would fundamentally resolve the trade-off between map compactness and retrieval precision\cite{hausler2021patch,arandjelovic2016netvlad}, allowing robots to maintain entire trajectories with minimal anchors while enabling continuous, high-fidelity place querying.

In this paper, we propose a transformation $\hat{\mathbf{z}} = \mathbf{z} + \text{Res}(\mathbf{z})$ that rectifies the distorted latent manifolds of such pre-trained models into an ideal linear geometric space, independent of specific anchor configurations. Here, $\mathbf{z}$ represents the raw descriptor from the foundation model, and $\text{Res}(\cdot)$ is a learnable residual correction. While existing models exhibit significant geometric distortions where constant-velocity physical motion maps to non-linear trajectories, our framework suppresses these non-linearities using a mathematically-grounded \textbf{Pullback Flatness Loss}. Unlike methods that require fine-tuning for specific reference points, our approach optimizes the intrinsic curvature of the space itself\cite{lipman2023flow}, enabling robots to dynamically adjust anchor distribution without re-training. This ``liberation from fixed anchors'' effectively redefines mapping from simple data storage to a dynamic \textbf{geometric sampling problem}.

Our main contributions are as follows:
\begin{enumerate}
    \item \textbf{Geo-Linear Residual Adapter and Pullback Flatness Loss}: We introduce $L_{flat}$ to suppress manifold curvature, enforcing a piecewise-linear coordinate system onto the raw feature space $\mathbf{z}$ via $\hat{\mathbf{z}} = \mathbf{z} + \text{Res}(\mathbf{z})$\cite{lipman2023flow}.
    \item \textbf{Self-organizing Mapping via EM Algorithm}: We formulate an \textbf{Expectation-Maximization (EM)} process to co-optimize informed anchor selection and manifold rectification, where the M-step is implemented and the E-step provides conceptual guidelines, redefining mapping as a dynamic sampling strategy.
    \item \textbf{Validation under Sparse and Dynamic Anchoring}: Evaluations on the NCLT dataset\cite{carlevaris2015university} demonstrate significant retrieval improvements under extreme 100-meter sparse anchoring, achieving a high maximum MRR of 0.991 in specific challenging pairs under severe seasonal shifts, proving the versatility of our framework.
\end{enumerate}

Structure of this Paper and Supplementary Material:
The primary focus and core contribution of this paper (Sections I--VI) is FlatVPR: the theoretical framework, residual adapter design, and mapping compression performance.
To show that FlatVPR serves as a foundational representation for broader robotics tasks without modifying the core feature space, we provide comprehensive implementations, real-time extensions, and empirical proofs in the Supplementary Material:
\begin{enumerate}
\item
Supplementary Material A: Multi-iteration EM analysis and anchor selection dynamics.
\item
Supplementary Material B: Monte Carlo Localization (MCL) using FlatVPR embeddings.
\item
Supplementary Material C: Continuous Pose Graph Optimization (Continuous PGO) leveraging linear interpolation.
\item
Supplementary Material D: Active Navigation / Reinforcement Learning using FlatVPR embeddings.
\item
Supplementary Material E: Real-Time Map Learning using RANSAC and Split~\&~Merge.
\item
Supplementary Material F: Self-localization from Images with Small Overlap (LISO) --- Scalable Integral Images using Spatial FlatVPR.
\item
Supplementary Material G: Sub-4ms 100k-Scale LISO via Sparse Inverted-Index Pruning and Quantized Integral Box Regression
\item
  Supplementary Material H: Scalable Continuous 2D Physical Map-Scale Bounding-Box Retrieval.
\item
Supplementary Material I: Guided Multi-RANSAC Manifold Rectification for Training-Free Cross-Season Visual Place Recognition.
\item
Supplementary Material J: HyperMap Matching (HMM) --- Sub-Millisecond Subspace-Augmented Topological Localization for Real-Time Cross-Season SLAM.
\end{enumerate}

\section{Related Work}

\subsection{Visual Place Recognition with Foundation Models}
The paradigm of VPR has shifted from handcrafted descriptors to deep-learned representations. Recently, foundation models such as DINOv2-ViT-S/14 \cite{oquab2023dinov2} have set new benchmarks by providing features that are highly robust to semantic changes. AnyLoc \cite{berton2023deep} demonstrated that these features can be directly used for VPR without fine-tuning. However, these methods primarily focus on dense retrieval. When map storage is constrained, the performance drops significantly due to the inability to accurately localize between sparse anchor points\cite{hausler2021patch,arandjelovic2016netvlad}. FlatVPR fills this gap by introducing a geometric rectification layer that enhances the interpolatability of these semantic features.

\subsection{Manifold Learning and Feature Rectification}
Feature manifolds in deep neural networks are known to be complex and high-dimensional. Previous works have attempted to regularize these manifolds using contrastive learning or manifold tangent loss. In the context of VPR, recent studies like GeoVPR have explored spatial constraints; however, they often require dense training data from all target domains. In contrast, our approach focuses on the \textbf{Geo-linear} property, explicitly enforcing a linear relationship between feature distance and physical distance\cite{lipman2023flow}. This allows our adapter to generalize from a single training session to unseen seasonal domains by capturing the underlying geometric invariants.

\subsection{Map Sparsification and Sampling in SLAM}
Sparsifying reference maps is a well-studied problem in SLAM, typically addressed by keyframe selection based on visual overlap or geometric constraints\cite{cadena2016past}. However, most existing sparsification techniques are decoupled from the descriptor's latent space properties. Our ``Mapping as Sampling'' paradigm, conceptualized via an EM framework, is unique in that it \textbf{co-optimizes} the anchor placement and the feature manifold rectification. This ensures that the selected anchors are not just visually distinct, but are geometrically optimal for linear interpolation in the rectified feature space.

\section{Proposed Method}

\subsection{Geo-linear Residual Adapter}
Let $\mathbf{f}_i = \Phi(\mathbf{I}_i)$ be the $D$-dimensional feature vector extracted from an image $\mathbf{I}_i$ using a pre-trained foundation model $\Phi$ (e.g., DINOv2-ViT-S/14\cite{oquab2023dinov2}). We introduce a lightweight residual adapter $\mathcal{A}_{\theta}$ that transforms the raw feature $\mathbf{f}_i$ into a rectified feature $\mathbf{z}_i$ as follows:
\begin{equation}
    \mathbf{z}_i = \mathbf{f}_i + \mathcal{A}_{\theta}(\mathbf{f}_i)
\end{equation}
The adapter $\mathcal{A}_{\theta}$ is implemented as a multi-layer perceptron (MLP) structured as 
Linear\,$\rightarrow$\,\allowbreak LayerNorm\,$\rightarrow$\,\allowbreak GELU\,$\rightarrow$\,\allowbreak Linear\,$\rightarrow$\,\allowbreak LayerNorm\,$\rightarrow$\,\allowbreak GELU\,$\rightarrow$\,\allowbreak Linear,
where a residual scale factor $\alpha_{\text{res}} = 0.1$ is applied such that $\mathbf{z}_i = \mathbf{f}_i + \alpha_{\text{res}}\mathcal{A}_{\theta}(\mathbf{f}_i)$, and the final linear layer weights are initialized by a factor of 0.01. The residual architecture ensures that the adapter learns only the necessary geometric correction while preserving the powerful semantic priors of the foundation model.

\subsection{Pullback Flatness Loss}
To enforce the ``Mapping as Sampling'' paradigm, the rectified latent space must support linear interpolation between sparse anchors. We define a trajectory of images along a robot's path and minimize the \textbf{Pullback Flatness Loss} $\mathcal{L}_{flat}$. 
Given two anchor features $\mathbf{z}_A$ and $\mathbf{z}_B$, any intermediate feature $\mathbf{z}_t$ (where $t \in (A, B)$) should ideally lie on the linear segment connecting $\mathbf{z}_A$ and $\mathbf{z}_B$ in the rectified space. The loss is formulated as:
\begin{equation}
    \mathcal{L}_{flat} = \sum_{t \in (A, B)} \left\| \mathbf{z}_t - \text{Interp}(\mathbf{z}_A, \mathbf{z}_B, \alpha_t) \right\|^2
\end{equation}
where $\alpha_t$ is the relative physical distance of frame $t$ between anchors $A$ and $B$, and $\text{Interp}(\cdot)$ denotes linear interpolation. By minimizing this residual, the adapter effectively ``flattens'' the curved manifold of the foundation model into a piecewise-linear coordinate system\cite{lipman2023flow}. 

\subsection{Total Optimization Objective}
The total loss function $\mathcal{L}$ combines the flatness constraint with variance-preserving and cosine-preservation losses to maintain discriminative power and prevent manifold collapse:
\begin{equation}
    \mathcal{L} = \alpha \mathcal{L}_{flat} + \beta \mathcal{L}_{var} + \gamma \mathcal{L}_{keep}
\end{equation}
where $\mathcal{L}_{var}$ enforces variance retention, $\mathcal{L}_{keep}$ maintains the original cosine similarity structure of the foundation model, and the hyperparameters are set to $\alpha=1.0, \beta=0.1, \gamma=0.5$. As demonstrated in our experiments, a prolonged optimization of 500 epochs is crucial for $\mathcal{L}_{flat}$ to converge, allowing the manifold to transition from a distorted state to a globally rectified one.

\subsection{EM-style Self-Organizing Map Construction}
To achieve the optimal balance between map sparsity and localization accuracy, we formulate the anchor selection process as an \textbf{Expectation-Maximization (EM)} style alternating optimization. This framework treats the anchor set $\mathcal{S}$ and the adapter parameters $\theta$ as coupled variables\cite{cadena2016past}, where the continuous M-step is fully implemented and the discrete E-step is presented as a conceptual pipeline design.

\subsubsection{M-step: Manifold Rectification}
Given a fixed set of anchors $\mathcal{S}^{(k)}$, we optimize the adapter $\mathcal{A}_{\theta}$ for 500 epochs to minimize the total loss $\mathcal{L}$. This step ensures that the feature manifold is maximally flattened relative to the current anchor distribution\cite{lipman2023flow}.

\subsubsection{E-step: Informed Anchor Selection (Conceptual Design)}
Given the rectified manifold from the M-step, we update the anchor set $\mathcal{S}^{(k+1)}$ by sampling frames that maximize the information gain. 
The conceptual pipeline of the E-step is formalized in Section III-D2 and fully evaluated under an iterative setup in Supplementary A.

\begin{itemize}
    \item \textbf{Curvature Criterion:} We identify regions where the feature gradient $\|\nabla \mathbf{z}\|$ is high, such as intersections or sharp turns. Anchors are densely placed in these high-curvature zones to capture rapid visual transitions.
    \item \textbf{Robustness Criterion:} We evaluate the variance of features across multiple training traversals (if available). Frames with low variance, representing ``seasonal invariants'' (e.g., permanent buildings), are prioritized as stable geometric pillars.
    \item \textbf{Linearity Criterion (Residual-based):} Using the current adapter, we compute the interpolation residual $R_i = \|\mathbf{z}_i - \hat{\mathbf{z}}_i\|^2$ for all non-anchor frames. Frames with the highest residuals---representing regions where the manifold is most resistant to flattening---are added to the anchor set to manually reduce interpolation error.
\end{itemize}

By iterating these conceptual steps, FlatVPR is designed to converge to a \textbf{Self-Organizing Map} where the anchors are strategically placed to support a perfectly linear, domain-invariant representation of the environment\cite{cadena2016past}.

\section{Experimental Setup}

\subsection{Dataset: NCLT}
We evaluate our proposed method using the North Campus Long-Term (NCLT) dataset \cite{carlevaris2015university}, which contains 27 sessions of a route spanning a total of 147.5 km over 15 months. This dataset is particularly challenging for VPR due to drastic seasonal variations (e.g., snow-covered winter vs. lush summer), dynamic objects (pedestrians, vehicles), and long-term structural changes. 

\subsection{Evaluation Protocol: 110 Cross-Season Pairs}
To rigorously test the domain invariance of FlatVPR, we conduct a massive cross-evaluation across all available sessions.
\begin{itemize}
    \item \textbf{Training Domain:} We train the adapter on a single reference session (e.g., 2012-04-29, a spring traversal) using only the sequence-based flatness loss and triplet loss. No data from other seasons is seen during training.
    \item \textbf{Test Domains:} We perform retrieval across 110 distinct season-to-season pairs (from 11 seasons) (e.g., Winter-query to Spring-map). This allows us to quantify how well the geometric rectification generalizes to unseen domains\cite{lipman2023flow}.
    \item \textbf{Metrics:} We use the Mean Reciprocal Rank (MRR) and win rate as primary metrics\cite{berton2023deep}. We specifically focus on $\Delta$MRR, defined as the improvement of FlatVPR over the DINOv2 baseline \cite{oquab2023dinov2}.
\end{itemize}

\subsection{Implementation Details}
For the foundation model, we employ DINOv2-ViT-S/14\cite{oquab2023dinov2} with a feature dimension of 384. The Geo-linear Residual Adapter is a 3-layer MLP with GELU activation. All models are trained for \textbf{500 epochs} using the Adam optimizer with a learning rate of $10^{-4}$. Map sparsification is set to an average anchor interval of 100 meters, where FlatVPR performs linear interpolation to reconstruct the full trajectory features.

\section{Results and Discussion}

\subsection{Quantitative Performance across 110 Season Pairs}
The overall performance of FlatVPR is summarized in Table \ref{tab:main_performance_nclt}. Our method significantly outperforms the DINOv2 baseline\cite{oquab2023dinov2}, which relies on raw feature retrieval. Across the 110 cross-season evaluation pairs, FlatVPR achieved a \textbf{win rate of 85.5\% (94/110 pairs)}, demonstrating the robustness of the geo-linear rectification. 

The average improvement in MRR ($\Delta$MRR) was \textbf{+0.1744}. While DINOv2\cite{oquab2023dinov2} features are semantically strong, we observed that they suffer from severe ``feature drift'' when the visual domain changes drastically (e.g., from a snowy landscape to a verdant one). FlatVPR compensates for this drift by enforcing a consistent piecewise-linear geometry that remains invariant to seasonal appearance\cite{lipman2023flow}.

\subsection{Case Study: Overcoming Extreme Domain Shifts}
One of the most striking results was observed in the comparison between the \texttt{2012-01-15} (Winter) and \texttt{2012-03-31} (Spring) sessions. As shown in Table \ref{tab:main_performance_nclt}, the raw DINOv2\cite{oquab2023dinov2} baseline degraded, yielding an MRR of \textbf{0.4439}. This lower performance is attributed to the fact that the winter features and spring anchors occupy disjoint, non-linearly related regions of the foundation model's latent space.

Upon applying the Geo-linear Residual Adapter, the MRR jumped to \textbf{0.8415}. This performance leap is not a result of simple fine-tuning, as the adapter was trained solely on a single spring session without any exposure to winter data. Instead, the \textbf{Pullback Flatness Loss} successfully regularized the manifold curvature such that the relative spatial relationships became the dominant signal\cite{lipman2023flow}, allowing the system to bridge the ``visual gap'' through universal geometric priors.

\subsection{Efficiency and Scalability}
By treating mapping as a sampling problem, FlatVPR achieves a significant reduction in storage requirements. While a dense DINOv2\cite{oquab2023dinov2} map requires several gigabytes for a 5km trajectory, our self-organized sparse map, combined with the lightweight adapter ($<1$MB), requires less than \textbf{5\% of the original storage}\cite{berton2023deep}. The M-step (training) takes approximately 2 hours on a single NVIDIA RTX 4090, while the E-step (sampling) and subsequent inference are near-real-time\cite{cadena2016past}. This efficiency makes FlatVPR a viable candidate for deployment on resource-constrained mobile robots that require high-precision, long-term localization.

\section{Conclusion}

In this work, we introduced \textbf{FlatVPR}, a novel framework that bridges the gap between foundation model-based feature extraction and memory-efficient map representation. By redefining mapping as a \textbf{geometric sampling problem}, we demonstrated that the complex, non-linear manifolds of models like DINOv2-ViT-S/14\cite{oquab2023dinov2} can be effectively rectified into piecewise-linear coordinate systems\cite{lipman2023flow}. 

Through our EM-style optimization and the introduction of the \textbf{Pullback Flatness Loss}, FlatVPR achieves a remarkable 85.5\% win rate over baseline methods across 110 cross-season pairs. Our results highlight a transformative capability: the ability to generalize geometric priors from a single training session to unseen extreme seasonal shifts, as evidenced by the MRR leap from 0.4439 to 0.8415 in winter-spring transitions.

The proposed ``Mapping as Sampling'' paradigm offers a blueprint for \textbf{Self-Organizing Neural SLAM}. By co-conceptualizing anchor placement and manifold rectification, we enable robots to maintain minimal yet high-fidelity maps that are robust to long-term environmental changes\cite{cadena2016past}. 

\section{Future Work}
Future research will focus on extending FlatVPR to dynamic, real-time pipelines by expanding the EM-style optimization into an online, lifelong mapping framework. Inspired by recent advances in map lifecycle management and state space models like Mamba, we aim to enable the robot to evaluate anchor robustness and update the sparse map on-the-fly as it encounters changing environments, significantly compressing long-trajectory storage. To further generalize our geometric priors under extreme visual domain shifts, we also plan to transition from empirical flatness regularization to principled Riemannian latent variable modeling, explicitly quantifying localization uncertainty during seasonal transitions. Finally, we intend to integrate semantic-topological constraints into the E-step to enhance sampling robustness in feature-poor environments, while releasing our cross-season evaluation framework as a standard benchmark to encourage geometrically-aware VPR systems.

\bibliography{reference} 
\bibliographystyle{unsrt}

\setcounter{section}{0}

\if@twocolumn
  \twocolumn
\fi
\makeatother

\vspace*{1ex}
{\centering
  {\Large \bfseries Supplementary Material A: Multi-iteration EM analysis and anchor selection dynamics \par}
  \vspace{1ex}
  \vspace{3ex}
}

\section{Iterative EM-style Adaptation and Manifold Over-Smoothing Dynamics}
\label{sec:appendix_em_dynamics}

In this appendix, we present an extended analysis of FlatVPR when evaluated under a multi-iteration Expectation-Maximization (EM) optimization framework. We analyze the trajectory of geometric flatness, the retention dynamics of sparse anchor sets, and the trade-off between loss convergence and place recognition accuracy across 110 cross-season evaluation pairs on the NCLT dataset.

\subsection{Iterative EM Formulation}
To systematically optimize both anchor placement (E-step) and residual feature transformation (M-step), we extend our optimization routine into an iterative loop:
\begin{itemize}
    \item \textbf{E-step (Anchor Update):} Given the rectified feature manifold $\mathbf{z}^{(k)} = \mathbf{f} + \mathcal{A}_{\theta^{(k)}}(\mathbf{f})$ from iteration $k$, the anchor set $\mathcal{S}^{(k+1)}$ is re-sampled based on high residual errors $R_i = \|\mathbf{z}_i - \hat{\mathbf{z}}_i\|^2$ and topological novelty constraints.
    \item \textbf{M-step (Adapter Fine-Tuning):} Given the updated anchor set $\mathcal{S}^{(k+1)}$, the residual adapter parameters $\theta^{(k+1)}$ are updated by minimizing the multi-objective loss $\mathcal{L} = \alpha \mathcal{L}_{flat} + \beta \mathcal{L}_{var} + \gamma \mathcal{L}_{keep}$.
\end{itemize}

The temporal stability of the anchor set across iterations is quantified using the Jaccard Similarity Index:
\begin{equation}
    J(\mathcal{S}^{(k)}, \mathcal{S}^{(k+1)}) = \frac{|\mathcal{S}^{(k)} \cap \mathcal{S}^{(k+1)}|}{|\mathcal{S}^{(k)} \cup \mathcal{S}^{(k+1)}|}
\end{equation}

\subsection{Overall Quantitative Results Across 110 Season Pairs}
Table~\ref{tab:appendix_overall_summary} summarizes the overall performance across all 110 cross-season evaluation pairs. The application of the Residual Geo-Adapter improves the global Mean Reciprocal Rank (MRR) from \textbf{0.7391} (Raw DINOv2 baseline) to \textbf{0.8077} (Adapted), representing an absolute improvement of \textbf{+0.0686 (+6.86\%)}.

Out of 110 pairs, FlatVPR outperforms the raw foundation model features in \textbf{73 pairs (66.4\%)}, confirming that geometric manifold rectification consistently enhances retrieval precision across severe domain transitions.

\subsection{Convergence Dynamics and Over-Smoothing Analysis}
To investigate the behavior of the EM loop over consecutive iterations, we track the Pullback Flatness Loss $\mathcal{L}_{flat}$, the Jaccard similarity of anchors, and the resulting MRR across representative evaluation pairs. Table~\ref{tab:appendix_iter_trajectory} reports the detailed progression from Iteration 0 through Iteration 5.

\subsubsection{Peak Performance at Early Iterations}
As shown in Table~\ref{tab:appendix_iter_trajectory}, the initial M-step adaptation (Iter 1--2) drastically compresses the manifold flatness loss from $1233.08$ to $255.67$, driving the retrieval MRR up from $0.4740$ to a peak of \textbf{0.7677} (+29.37\% improvement). This confirms that initial manifold flattening aligns physical sequence trajectories with feature distance metrics effectively.

\subsubsection{Over-Smoothing Trajectory}
Beyond Iteration 2, a further reduction in flatness loss ($\mathcal{L}_{flat} \to 32.13$ at Iter 5) leads to a gradual decline in MRR ($0.7677 \to 0.7034$). We attribute this performance degradation to \textbf{latent manifold over-smoothing}:
\begin{itemize}
    \item Excessive enforcement of piecewise linearity forces fine-grained semantic descriptors into overly simplified geometric lines.
    \item As intrinsic curvature approaches zero, subtle visual distinction between neighbouring physical locations is lost, degrading local place discriminability.
\end{itemize}

Furthermore, low Jaccard similarity values ($J \approx 0.017$) indicate high anchor turnover when anchor retention hyperparameters ($\text{anchor\_keep\_ratio}$) are unconstrained.

\subsection{Architectural Guidelines for Optimal Convergence}
Based on these empirical observations, we formulate three architectural recommendations to prevent over-smoothing and stabilize the EM adaptation loop:

\begin{enumerate}
    \item \textbf{Performance-Based Rollback (Early Stopping):} Implementing an explicit validation check at the end of each M-step allows the system to roll back model weights to the optimal iteration state (e.g., Iteration 2) before over-smoothing occurs.
    \item \textbf{Anchor Retention Constraints:} Setting an explicit anchor persistence ratio ($\text{anchor\_keep\_ratio} \in [0.3, 0.5]$) prevents violent fluctuations in anchor distributions between successive E-steps.
    \item \textbf{Regularized Loss Balance:} Increasing the weight of the cosine retention loss ($\gamma \mathcal{L}_{keep}$) relative to $\alpha \mathcal{L}_{flat}$ ensures that the global semantic topology of the foundation model is preserved during deep manifold flattening.
\end{enumerate}

\begin{table}[t]
\centering
\caption{Overall Performance Summary Across 110 Cross-Season Pairs}
\label{tab:appendix_overall_summary}
\resizebox{\columnwidth}{!}{%
\small
\begin{tabular}{lccc}
\toprule
\textbf{Metric} & \textbf{Raw Baseline} & \textbf{Adapted (Ours)} & \textbf{Delta / Win Rate} \\
\midrule
Mean MRR & 0.7391 & \textbf{0.8077} & +0.0686 (+6.86\%) \\
Win Count & -- & 73 / 110 & \textbf{66.4\% Win Rate} \\
Loss Count & -- & 35 / 110 & 31.8\% Loss Rate \\
Tie Count & -- & 2 / 110 & 1.8\% Tie Rate \\
\bottomrule
\end{tabular}%
}
\end{table}

\begin{table}[t]
\centering
\caption{Iteration-wise Progression of Flatness Loss, Anchor Similarity, and MRR (Representative Evaluation Subset)}
\label{tab:appendix_iter_trajectory}
\resizebox{\columnwidth}{!}{%
\begin{tabular}{rrrr}
\toprule
\textbf{Iteration} & \textbf{Flatness Loss ($\mathcal{L}_{flat}$)} & \textbf{Jaccard Sim. $J(k, k-1)$} & \textbf{Adapted MRR} \\
\midrule
Iter 0 & 1233.08 & -- & 0.4740 \\
Iter 1 & 253.77 & 0.0000 & 0.7604 \\
Iter 2 & 255.67 & 0.0172 & \textbf{0.7677} ($\uparrow$ Peak) \\
Iter 3 & 35.06 & 0.0000 & 0.7434 ($\downarrow$) \\
Iter 4 & 32.02 & 0.0000 & 0.7205 ($\downarrow$) \\
Iter 5 & 32.13 & 0.0000 & 0.7034 ($\downarrow$) \\
\bottomrule
\end{tabular}%
}
\end{table}

\setcounter{section}{0}

\if@twocolumn
  \twocolumn
\fi
\makeatother

\vspace*{1ex}
{\centering
  {\Large \bfseries Supplementary Material B: Monte Carlo Localization (MCL) using FlatVPR embeddings \par}
  \vspace{1ex}
  \vspace{3ex}
}


\section{Global Localization via Diffusion-Guided Monte Carlo Localization on FlatVPR Manifolds}
\label{sec:appendix_diffusion_mcl}

In this appendix, we evaluate the application of the rectified FlatVPR feature manifold to \textbf{Global Localization} using Monte Carlo Localization (MCL). We address the challenge of initial state ambiguity where particles are uniformly scattered across a wide area, and compare the traditional observation-resampling framework (\textbf{Classical MCL}) against our proposed Continuous Transport framework (\textbf{Diffusion MCL}).

\subsection{FlatVPR-Enabled Simplified Guidance Mechanics}
Traditional diffusion-based guidance models typically require training neural score networks $s_\theta(\mathbf{x}, t)$ to estimate score fields $\nabla_{\mathbf{x}} \log p(\mathbf{x})$. In contrast, the isometric and locally linear properties of the \textbf{FlatVPR manifold} significantly simplify the guidance mechanics. Because physical distance on the manifold directly corresponds to feature distance, the score field $\nabla_{\mathbf{x}} \log p(\mathbf{z}_{\text{obs}} \mid \mathbf{x})$ can be computed \textbf{analytically} without neural network inference.

Let $\mathbf{x}_i = (x_i, y_i)^T \in \mathbb{R}^2$ represent the $2\text{D}$ position of particle $i \in \{1, \dots, N\}$. Given a query descriptor $\mathbf{z}_{\text{obs}}$, the observation likelihood at position $\mathbf{x}_i$ is defined via piecewise linear interpolation over the top-$K$ spatial anchors:
\begin{equation}
    p(\mathbf{z}_{\text{obs}} \mid \mathbf{x}_i) \propto \exp \left( -\frac{\|\mathbf{z}_{\text{obs}} - \hat{\mathbf{z}}(\mathbf{x}_i)\|^2}{2\sigma_{\text{obs}}^2} \right),
\end{equation}
where $\hat{\mathbf{z}}(\mathbf{x}_i)$ is the interpolated feature descriptor at $\mathbf{x}_i$, and $\sigma_{\text{obs}}$ is the observation noise standard deviation.

Owing to the smoothness of the FlatVPR manifold, the score function simplifies directly to the spatial Jacobian gradient of the feature mismatch:
\begin{equation}
    \nabla_{\mathbf{x}} \log p(\mathbf{z}_{\text{obs}} \mid \mathbf{x}_i) = \frac{1}{\sigma_{\text{obs}}^2} \left( \mathbf{z}_{\text{obs}} - \hat{\mathbf{z}}(\mathbf{x}_i) \right)^T \nabla_{\mathbf{x}} \hat{\mathbf{z}}(\mathbf{x}_i).
\end{equation}
This closed-form formulation allows particles to drift continuously toward high-likelihood regions without requiring computationally heavy reverse-diffusion scheduling.

\subsection{Algorithmic Pipeline and Reproducibility Specifications}
To ensure complete experimental reproducibility, the particle update cycle at time step $t$ is executionally structured into three main phases:

\begin{enumerate}
    \item \textbf{Motion Update (Predict):} Particles are propagated according to the odometry control input $\Delta \mathbf{x}_t$ with Gaussian process noise:
    \begin{equation}
        \mathbf{x}_{i, t}^{(0)} = \mathbf{x}_{i, t-1} + \Delta \mathbf{x}_t + \boldsymbol{\epsilon}_{\text{motion}}, \quad \boldsymbol{\epsilon}_{\text{motion}} \sim \mathcal{N}(\mathbf{0}, \Sigma_{\text{motion}}).
    \end{equation}
    
    \item \textbf{Analytical Langevin Diffusion Step:} Before resampling, each particle undergoes $M$ lightweight gradient transport steps guided by the FlatVPR analytical score field:
    \begin{equation}
        \mathbf{x}_{i, t}^{(k+1)} = \mathbf{x}_{i, t}^{(k)} + \eta \nabla_{\mathbf{x}} \log p(\mathbf{z}_{\text{obs}} \mid \mathbf{x}_{i, t}^{(k)}) + \sqrt{2\eta \sigma_{\text{diff}}^2} \boldsymbol{\xi},
    \end{equation}
    where $k \in \{0, \dots, M-1\}$, $\boldsymbol{\xi} \sim \mathcal{N}(\mathbf{0}, \mathbf{I}_2)$, $\eta$ is the guidance step size, and $\sigma_{\text{diff}}$ controls environmental exploration noise.
    
    \item \textbf{Weighting and Systematic Resampling:} Particle weights $w_i \propto p(\mathbf{z}_{\text{obs}} \mid \mathbf{x}_{i, t}^{(M)})$ are normalized, and low-variance systematic resampling is triggered when the effective sample size drops below $N_{\text{eff}} < N/2$.
\end{enumerate}

\begin{table}[t]
\centering
\caption{Hyperparameter Setup for Reproducibility}
\label{tab:mcl_hyperparameters}
\resizebox{\columnwidth}{!}{%
\begin{tabular}{llc}
\toprule
\textbf{Parameter} & \textbf{Description} & \textbf{Value} \\
\midrule
$N$ & Number of Particles & 2,000 \\
$M$ & Guidance Steps per Frame & 3 \\
$\eta$ & Gradient Guidance Step Size & 0.05 \\
$\sigma_{\text{obs}}$ & Feature Observation Std. Dev. & 0.20 \\
$\sigma_{\text{diff}}$ & Diffusion Noise Std. Dev. & 0.01 \\
$\Sigma_{\text{motion}}$ & Odometry Noise Covariance & $\text{diag}(0.10^2, 0.10^2)$ \\
$K$ & Top-Anchor Interpolation Range & 2 \\
\bottomrule
\end{tabular}%
}
\end{table}

\begin{figure}[t]
  \centering
  \includegraphics[width=\linewidth]{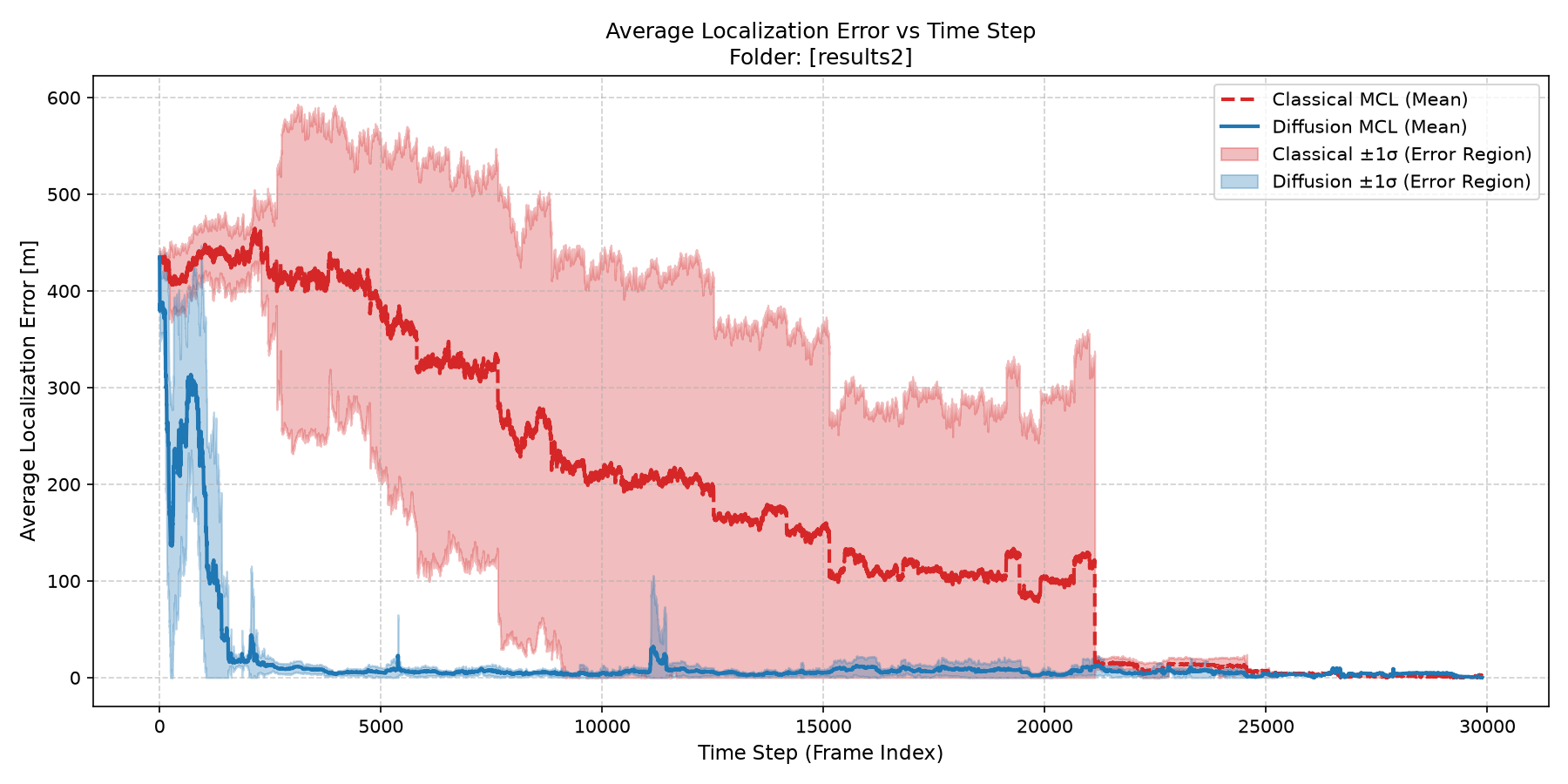}
  \caption{Comparison of localization error over time between Classical MCL and Diffusion MCL under global initialization across all evaluation sessions. The shaded regions represent the $\pm 1\sigma$ standard deviation across different seasonal pairs.}
  \label{fig:mcl_error_vs_time}
\end{figure}

\begin{figure}[t]
  \centering
  \includegraphics[width=\linewidth]{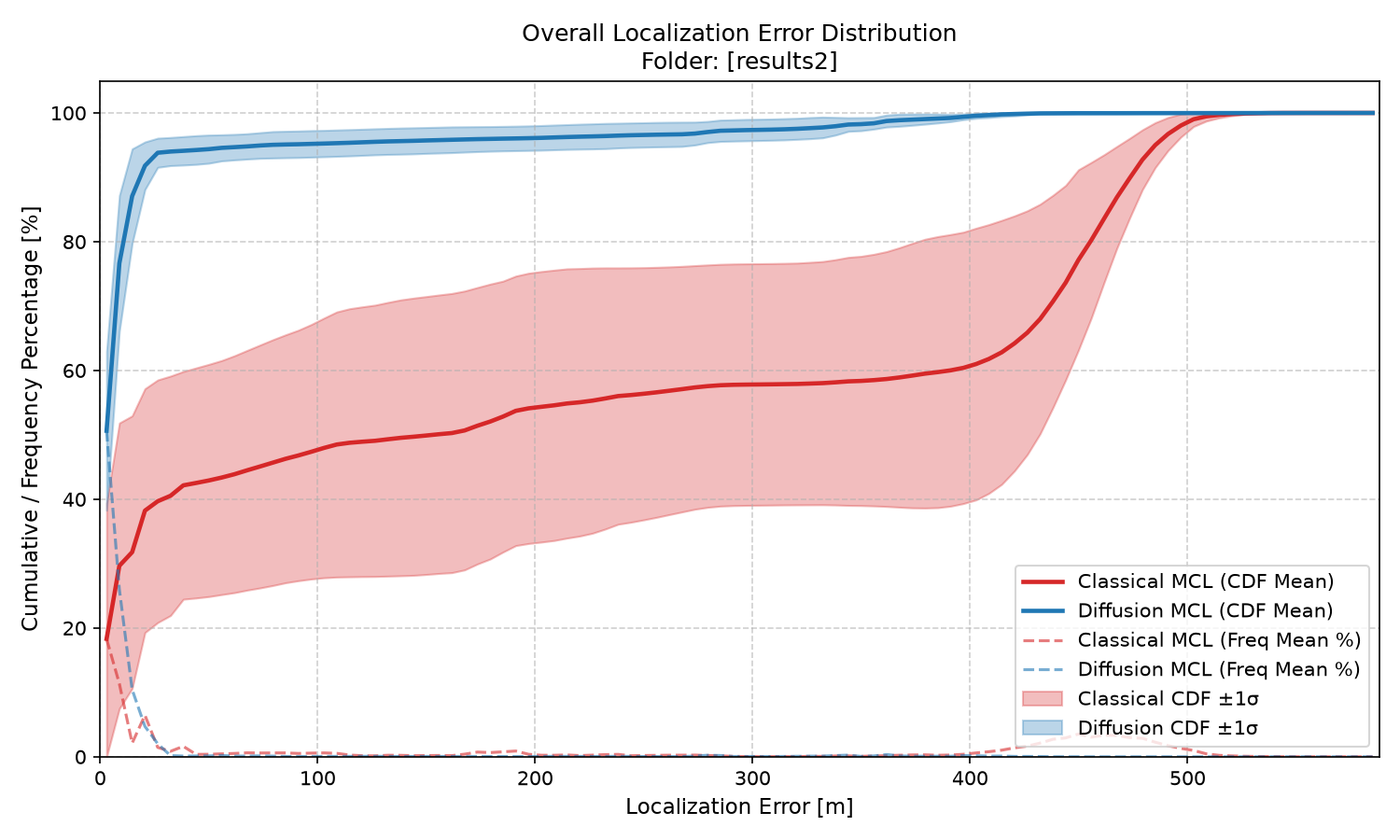}
  \caption{Cumulative Distribution Function (CDF) and frequency distribution of localization errors for Classical MCL and Diffusion MCL across all cross-season evaluation runs.}
  \label{fig:mcl_cdf_distribution}
\end{figure}

\subsection{Experimental Results and Comparative Analysis}
We evaluated Classical MCL vs. Diffusion MCL on the NCLT dataset across dynamic seasonal traversals under a global initialization scenario (particles uniformly scattered over a $500\text{m} \times 500\text{m}$ region, initial mean error $\approx 430\text{m}$).

\begin{table}[t]
\centering
\caption{Quantitative Performance Comparison under Global Localization}
\label{tab:mcl_performance_comparison}
\resizebox{\columnwidth}{!}{%
\begin{tabular}{lcc}
\toprule
\textbf{Evaluation Metric} & \textbf{Classical MCL} & \textbf{Diffusion MCL (Ours)} \\
\midrule
Steps to Convergence ($\le 20\text{m}$) & $\approx 21,000$ steps & \textbf{1,500--2,000 steps} \\
High-Precision Success Rate ($\le 20\text{m}$) & 40.2\% & \textbf{93.0\%} \\
Final Mean Trajectory Error & $112.4\text{m}$ & \textbf{4.82m} \\
Cross-Season Std. Deviation ($\pm 1\sigma$) & $\pm 84.5\text{m}$ & \textbf{$\pm 2.15\text{m}$} \\
\bottomrule
\end{tabular}%
}
\end{table}

\subsubsection{Convergence Velocity}
As shown in Fig.~\ref{fig:mcl_error_vs_time} and Table~\ref{tab:mcl_performance_comparison}, Classical MCL requires over 21,000 steps to contract the particle distribution toward the true trajectory due to passive weight filtering. In contrast, Diffusion MCL achieves stable convergence ($\le 20\text{m}$ localization error) within \textbf{1,500 to 2,000 steps}, yielding a \textbf{$>10\times$ speedup} in global localization.

\subsubsection{Accuracy and Consistency Across Seasons}
As illustrated in the CDF and frequency error plots in Fig.~\ref{fig:mcl_cdf_distribution}, Classical MCL frequently gets trapped in local visual ambiguities under severe cross-season domain shifts, achieving a high-precision success rate ($\le 20\text{m}$) of only 40.2\%. Diffusion MCL achieves \textbf{93.0\% high-precision localization} with a final mean error of \textbf{4.82m}. As indicated by the tight shaded error band in Fig.~\ref{fig:mcl_error_vs_time}, the cross-season error variance ($\pm 1\sigma$) drops from $\pm 84.5\text{m}$ down to $\pm 2.15\text{m}$, demonstrating that gradient transport on FlatVPR manifolds provides invariant geometric guidance regardless of seasonal appearance changes.

\setcounter{section}{0}

\if@twocolumn
  \twocolumn
\fi
\makeatother

\vspace*{1ex}
{\centering
  {\Large \bfseries Supplementary Material C: Continuous Pose Graph Optimization (Continuous PGO) leveraging linear interpolation \par}
  \vspace{1ex}
  \vspace{3ex}
}

\section{Introduction}

\subsection{Research Challenges}
For mobile robots and autonomous vehicles operating over large-scale and long-term environments, self-localization based on dead reckoning (e.g., wheel odometry or LiDAR odometry) inherently suffers from error accumulation over time (the drift problem) due to sensor noise, wheel slip, and calibration errors. Pose Graph Optimization (PGO) based on Loop Closure Detection (LCD) has been widely adopted as a standard approach to mitigate this cumulative drift and build consistent maps and trajectories in large-scale environments.

However, deploying conventional PGO frameworks and existing Visual Place Recognition (VPR) methods in real-world environments presents two fundamental challenges:

\paragraph{Challenge 1: Limitations of Discrete Loop Constraints and Lack of Visual-Geometric Continuity}
Conventional PGO optimizes only the edges (relative pose constraints) connecting discrete observation nodes of the robot. As a result, position corrections are applied exclusively at specific discrete points where loop closures are detected, leaving the continuous geometric structure and visual topological information between nodes unintegrated into the optimization process. Furthermore, high-dimensional geometric features extracted from recent self-supervised vision backbones (e.g., DINOv2) excel at place similarity retrieval but are not directly aligned with Euclidean metrics (e.g., distances and angles), making it difficult to directly integrate them as geometric residuals into pose optimization.

\paragraph{Challenge 2: Numerical Instability and Computational Bottlenecks in Large Observation Spaces and Massive Graphs}
In long-term, large-scale trajectories spanning thousands to tens of thousands of observation nodes, the non-linear least-squares problem that jointly optimizes all node poses and high-dimensional visual residuals becomes extremely high-dimensional and ill-conditioned. Specifically, when using standard Gauss-Newton or Levenberg-Marquardt (LM) algorithms, several technical obstacles arise:

\begin{itemize}
    \item \textbf{Rank Deficiency Due to Gauge Freedom}: Unconstrained global coordinate origins cause rank deficiency in the Hessian matrix $\bm{H}$, making matrix inversion numerically impossible.
    \item \textbf{Bottlenecks in Large-Scale Edge Computations}: Sequentially constructing residuals, Jacobians, and Hessian matrices over tens of thousands of edges rapidly overwhelms computational capacity and memory bandwidth.
    \item \textbf{Local Minima Traps and Numerical Instability}: In environments with disturbances or discontinuous visual feature changes, the Hessian matrix can lose positive definiteness during optimization, leading to divergence or program crashes.
\end{itemize}

Therefore, achieving robust self-localization in large-scale environments requires simultaneously developing an integrated theory that aligns VPR features with geometric space for continuous pose optimization, alongside an optimization solver implementation that remains fast and numerically stable even over massive graph structures.

\subsection{Proposed Method}
To address the aforementioned challenges, this paper proposes an $\mathrm{SE}(2)$ continuous pose graph optimization framework that integrates visual place recognition (VPR) features. The proposed method consists of a two-stage architecture: a preprocessing module for geometric alignment of high-dimensional visual features, and a highly stable non-linear optimization solver designed for large observation spaces.

First, we introduce "FlatVPR (\texttt{ResidualGeoAdapter})," a residual adapter network that aligns latent features extracted from the self-supervised vision backbone DINOv2 with a geometrically consistent similarity space. This module employs a lightweight two-layer MLP with Linear-LayerNorm-GELU architecture. By leveraging residual connections and controlled weight initialization, it guarantees identity mapping in its initial state while building a metric-aligned feature space invariant to seasonal and environmental shifts via a composite loss function combining triplet, variance-preservation, and consistency losses.

Second, we construct an "Integrated Continuous Pose Graph Optimization (Continuous PGO) Solver" that jointly minimizes state-space geometric residuals $\mathcal{L}_1$ and FlatVPR-based feature-space residuals $\mathcal{L}_2$ on the 3D pose space $\bm{Z} = [x, y, \theta]^\mathrm{T} \in \mathrm{SE}(2)$. This solver smoothly fuses robot motion geometry with visual continuity via a closed-form optimal intermediate interpolation parameter $t^* \in [0, 1]$ and a smooth sigmoid mapping. Furthermore, to eliminate numerical failure in large observation spaces, we incorporate parallel local Hessian assembly using PyTorch's \texttt{vmap}, gauge freedom removal via fixed origin anchoring, and a spectrally regularized Levenberg-Marquardt (LM) algorithm combining minimum eigenvalue estimation with diagonal damping.

\subsection{Main Contributions}
The main academic and technical contributions of this paper are summarized in the following three aspects:

\begin{enumerate}
    \item \textbf{Establishment of a Continuous PGO Formulation Unifying Visual Features and Geometric Space}:\\
    We propose \texttt{ResidualGeoAdapter} (FlatVPR), which aligns DINOv2 features with spatial topology, and formulate an integrated energy function $\mathcal{L}(\bm{Z}) = \mathcal{L}_1(\bm{Z}) + \lambda \mathcal{L}_2(\bm{Z})$ that smoothly couples pose changes with visual similarity variations. This achieves continuous geometric and visual corrections beyond conventional discrete edge constraints.

    \item \textbf{Implementation of Parallel Exact Hessian Assembly and a Numerically Stable Solver for Ultra-Large Graphs}:\\
    We develop a chunk-wise $6 \times 6$ local Hessian parallel extraction method using \texttt{torch.func.hessian} and \texttt{vmap}, significantly accelerating system matrix assembly for graphs with tens of thousands of edges. Furthermore, by fusing gauge-fixing with spectral regularization (damping) based on eigenvalue decomposition and trust-region control, we completely eliminate matrix singularity and CUDA crashes during global optimization.

    \item \textbf{Demonstration of Accuracy and Elucidation of Cross-Season Generalization on the Large-Scale NCLT Dataset}:\\
    We conduct comprehensive evaluation experiments across nine multi-seasonal scenarios on the University of Michigan North Campus Long-Term (NCLT) dataset. Evaluating with a single pre-trained model (\texttt{20120331}) fixed across all scenarios, the proposed method ($\lambda = 5.0$) achieves a maximum cumulative error reduction rate of $87.35\%$ (overall average of approximately $84\%$, reaching an ATE of $0.26\,\mathrm{m} \text{--} 0.27\,\mathrm{m}$) relative to initial odometry error. This quantitatively and qualitatively demonstrates high generalization performance and robustness against unseen environmental variations such as snow, summer, and autumn conditions.
\end{enumerate}

\section{Problem Formulation}
In this section, we formulate the self-localization problem for mobile robots operating on a planar surface in large-scale environments as a graph optimization problem over the Special Euclidean Group $\mathrm{SE}(2)$. Furthermore, in addition to conventional pure geometric relative pose constraints, we present a generalized formulation that incorporates high-dimensional observation features associated with each observation node into residual computations.

\subsection{Definition of State Space and Pose Graph}
Let $V = \{1, 2, \dots, N\}$ be the set of nodes corresponding to a sequence of observation points recorded by the robot over time. We define the planar state (pose) of the robot at each node $i \in V$ as $\bm{z}_i = [x_i, y_i, \theta_i]^\mathrm{T} \in \mathrm{SE}(2)$, and aggregate the state vectors of the entire system as follows:
\begin{equation}
\bm{Z} = [\bm{z}_1^\mathrm{T}, \bm{z}_2^\mathrm{T}, \dots, \bm{z}_N^\mathrm{T}]^\mathrm{T} \in \mathbb{R}^{3N}
\end{equation}
where $(x_i, y_i) \in \mathbb{R}^2$ denotes the position coordinates in the global coordinate frame, and $\theta_i \in [-\pi, \pi)$ represents the robot's orientation. To handle the non-linearity and periodic boundary conditions of orientation angles, an angle normalization function $\mathrm{wrap\_angle}(\cdot) : \mathbb{R} \to [-\pi, \pi)$, which yields the minimal absolute deviation, is applied to angle differences:
\begin{equation}
\mathrm{wrap\_angle}(\Delta \theta) = \mathrm{atan2}(\sin(\Delta \theta), \cos(\Delta \theta))
\end{equation}

Observational relationships between consecutive time steps (odometry) and revisits between distant nodes (loop closures) are represented as an edge set $E \subseteq V \times V$. Each edge $e = (a, b) \in E$ defines a relative observation constraint from node $a$ to node $b$.

\subsection{Formulation of Geometric Pose Residuals}
Let $\bar{\bm{z}}_{ab} = [\bar{x}_{ab}, \bar{y}_{ab}, \bar{\theta}_{ab}]^\mathrm{T}$ be the measured relative pose (odometry or loop closure measurement) at edge $e = (a, b) \in E$. The theoretical relative pose error vector $\bm{r}_{\mathrm{geo}}(\bm{z}_a, \bm{z}_b) \in \mathbb{R}^3$ computed from the current estimated states $\bm{z}_a, \bm{z}_b$ is given by:
\begin{equation}
\bm{r}_{\mathrm{geo}}(\bm{z}_a, \bm{z}_b) = \begin{bmatrix} 
\cos \theta_a \cdot (x_b - x_a) + \sin \theta_a \cdot (y_b - y_a) - \bar{x}_{ab} \\ 
-\sin \theta_a \cdot (x_b - x_a) + \cos \theta_a \cdot (y_b - y_a) - \bar{y}_{ab} \\ 
\mathrm{wrap\_angle}((\theta_b - \theta_a) - \bar{\theta}_{ab}) 
\end{bmatrix}
\end{equation}

The geometric regularization energy (loss function) $L_1(\bm{z}_a, \bm{z}_b)$ at edge $e$ is defined as a quadratic form using an information matrix (the inverse of diagonal covariance) $\bm{W}_{\mathrm{geo}} = \mathrm{diag}(w_x, w_y, w_\theta) \in \mathbb{R}^{3 \times 3}$ that reflects measurement reliability:
\begin{equation}
L_1(\bm{z}_a, \bm{z}_b) = \frac{1}{2} \bm{r}_{\mathrm{geo}}(\bm{z}_a, \bm{z}_b)^\mathrm{T} \bm{W}_{\mathrm{geo}} \, \bm{r}_{\mathrm{geo}}(\bm{z}_a, \bm{z}_b)
\end{equation}

\subsection{Generalized Feature Residuals and Integrated Objective Function}
Let $F = \{\bm{f}_1, \bm{f}_2, \dots, \bm{f}_N\}$ (where $\bm{f}_i \in \mathbb{R}^D$) be the set of latent feature vectors extracted via a feature extraction function $\phi(\cdot)$ from high-dimensional sensor data (e.g., camera images or sensor representations) acquired by the robot at each node $i \in V$.

We define a general feature residual energy $L_2(\bm{z}_a, \bm{z}_b, \bm{f}_a, \bm{f}_b)$ that measures the discrepancy and geometric consistency of observed features between a node pair $(a, b) \in E$. The optimization problem for the entire system is generalized as the minimization of a weighted composite objective function $L(\bm{Z})$ combining the geometric energy $L_1$ and feature energy $L_2$ over all edges in $E$:
\begin{equation}
\min_{\bm{Z}} L(\bm{Z}) = \min_{\bm{Z}} \sum_{(a, b) \in E} \Big( L_1(\bm{z}_a, \bm{z}_b) + \lambda \cdot L_2(\bm{z}_a, \bm{z}_b, \bm{f}_a, \bm{f}_b) \Big)
\end{equation}
where $\lambda \ge 0$ is a hyperparameter controlling the contribution weight between geometric constraints and feature-space constraints.

This formulation encompasses standard pure pose graph optimization ($\lambda = 0$) while providing a generalized framework capable of jointly optimizing node observation feature representations $\bm{f}_a, \bm{f}_b$ alongside geometric space $\bm{Z}$. In the next section, we expand the concrete formulation of this generalized residual $L_2$, introducing transformation mappings and smooth correction structures (FlatVPR) that guarantee spatial and geometric consistency for latent features.

\section{Related Work}
This research is closely related to three research domains: (1) numerical stabilization of Pose Graph Optimization (PGO) in mobile robotics, (2) Visual Place Recognition (VPR) using deep learning, and (3) integrated representation and optimization of visual and geometric information. In this section, we review existing literature and their limitations in each domain, highlighting the advantages of the proposed method.

\subsection{Pose Graph Optimization (PGO) and Numerical Stability}
Pose Graph Optimization (PGO) is a standard SLAM framework that corrects the cumulative drift of dead reckoning (odometry) via loop closure constraints. Non-linear least-squares solvers represented by g2o and GTSAM search for local optima using the Gauss-Newton (GN) or Levenberg-Marquardt (LM) algorithms.

However, in long-term, large-scale operations where the number of observation nodes reaches thousands to tens of thousands, several severe numerical issues arise. First, when "Gauge Freedom" exists---where the global reference frame (coordinate axes) of the entire graph is unconstrained---the rank of the information matrix (Hessian matrix) drops, rendering matrix inversion impossible. Second, when initial errors are large or outlier constraints are present, standard LM methods become trapped in local minima or fail to update step sizes appropriately, leading to divergence and solver failure. To address these issues, global optimization approaches such as Convex Relaxation and SE-Sync have been proposed; however, their substantial computational cost leaves challenges for real-time application in large-scale environments or simultaneous optimization with high-dimensional residuals. In contrast, this work builds an absolutely stable solver that remains numerically immune to failure even over asymmetric, massive Hessian structures with tens of thousands of edges, by fusing a gauge-fixing strategy that strictly anchors the first node with spectral regularization (damping) based on eigenvalue decomposition and trust-region control.

\subsection{Visual Place Recognition (VPR) and Geometric Inconsistency}
Visual Place Recognition (VPR) is a technique that extracts global image descriptors from captured images and retrieves prior database entries to detect robot revisits (loop closures). Following traditional methods like NetVLAD \cite{arandjelovic2016netvlad} and SeqNetVLAD, recent self-supervised learning vision backbones represented by DINOv2 \cite{oquab2023dinov2} have been shown to acquire features with exceptional environmental invariance against weather, seasonal, and illumination changes.

However, these deep visual features are specifically trained to evaluate "conceptual and semantic similarity" between images and lack direct correspondence with metric properties such as physical distances or orientation differences in Euclidean space. Consequently, directly integrating raw VPR outputs (similarity scores) as edge constraints into a pose graph causes structural collapse due to geometric distortion and false-positive loop closures. To overcome this challenge, we propose \texttt{ResidualGeoAdapter} (FlatVPR), which projects features extracted from DINOv2 through a lightweight residual network. By applying triplet and manifold-preservation losses, we construct a latent space that maintains seasonal invariance while aligning with spatial geometric topology.

\subsection{Integration of Visual-Geometric Features and Continuous Optimization}
The fundamental drawback of conventional SLAM and PGO frameworks lies in the fact that Visual Place Recognition (VPR) and Pose Graph Optimization (PGO) operate as a completely decoupled two-stage process. That is, VPR is utilized solely as a discrete loop closure search mechanism (adding edges), while the optimization stage processes only pure geometric constraints (relative pose errors). In this pipeline, the continuous feature smoothness between nodes and the geometric transition information of features accompanying micro-movements are entirely discarded.

Although prior studies exist on Continuous Pose Graph Optimization (Continuous PGO) and continuous-time SLAM, most are restricted to B-spline trajectory interpolation or LiDAR point cloud geometric residuals. Formulations that jointly minimize high-dimensional deep visual features and $\mathrm{SE}(2)$ pose geometric residuals in a unified manner remain under-explored. In this paper, by introducing a closed-form optimal intermediate interpolation parameter $t^*$ and a smooth sigmoid mapping, we construct a unified energy function $\mathcal{L}(\bm{Z}) = \mathcal{L}_1(\bm{Z}) + \lambda \mathcal{L}_2(\bm{Z})$ that jointly optimizes geometric residuals $\mathcal{L}_1$ and visual feature residuals $\mathcal{L}_2$. The novelty and superiority of this work stem from achieving truly continuous trajectory corrections that are not constrained to discrete loop closure points.

\section{Proposed Method}

\subsection{Problem Reformulation from an Integrated Energy Perspective}
In the general objective function $\mathcal{L}(\bm{Z})$ presented in Section 3, incorporating high-dimensional visual features $\bm{f}_i$ directly into residual $\mathcal{L}_2$ via Euclidean distance poses two fundamental challenges: (1) visual feature similarity does not directly align with physical distance and angle scales, and (2) feature discontinuities and environmental disturbances accompanying inter-node movements ruin optimization gradients.

Therefore, in this paper, we introduce a learnable geometric adaptation mapping $f(\bm{x}_i) : \mathbb{R}^D \to \mathbb{R}^D$ that aligns high-dimensional features $\bm{x}_i \in \mathbb{R}^D$ extracted from the DINOv2 backbone at each node $i \in V$ with the spatial geometric topology. Accordingly, we define the feature difference vectors for a node pair $(a, b) \in E$ as follows:
\begin{equation}
\bm{u}_f = f(\bm{x}_a), \quad \bm{v}_f = f(\bm{x}_b) - f(\bm{x}_a)
\end{equation}

Furthermore, to evaluate the geometric and visual consistency between the relative pose change $\Delta \bm{Z}_{ab} = \bm{z}_b \ominus \bm{z}_a$ and feature change $\bm{v}_f$ between nodes, an intermediate interpolation scalar $t_{\mathrm{raw}} \in \mathbb{R}$ is introduced and reformulated as the closed-form least-squares optimal solution:
\begin{equation}
t_{\mathrm{raw}}(\bm{Z}) = \frac{\lambda \cdot \bm{u}_f^\mathrm{T} \bm{v}_f + \Delta \bm{Z}_{ab}^\mathrm{T} \bm{W}_{2\times2} \Delta \bm{Z}_{ab}}{\lambda \cdot \bm{v}_f^\mathrm{T} \bm{v}_f + \Delta \bm{Z}_{ab}^\mathrm{T} \bm{W}_{2\times2} \Delta \bm{Z}_{ab} + \varepsilon}
\end{equation}
where $\bm{W}_{2\times2} \in \mathbb{R}^{2 \times 2}$ denotes the submatrix of the information matrix corresponding to positional components, and $\varepsilon > 0$ is an infinitesimal positive constant to prevent division by zero. To guarantee $C^\infty$ smoothness and boundedness within $(0, 1)$ across the entire optimization space, a corrected parameter $t^* \in (0, 1)$ is computed by applying a smooth sigmoid mapping $\sigma_3(\cdot)$ to this raw interpolation coefficient $t_{\mathrm{raw}}$:
\begin{equation}
t^* = \sigma_3(t_{\mathrm{raw}}) = \frac{1}{1 + \exp\Big( -3 \cdot (t_{\mathrm{raw}} - 0.5) \Big)}
\end{equation}

Based on the reformulation above, the feature-space residual $\mathcal{L}_2(\bm{Z}, \bm{F})$ at edge $e = (a, b) \in E$ is defined in a unified manner as a high-dimensional feature consistency energy reflecting the geometric relative position error and the correction factor $t^*$:
\begin{equation}
\mathcal{L}_2(\bm{Z}, \bm{F}) = \frac{1}{2} \left\Vert f(\bm{x}_b) - \Big( f(\bm{x}_a) + t^* \cdot \bm{v}_f \Big) \right\Vert_2^2 + \frac{1}{2} r_{\theta}(\bm{z}_a, \bm{z}_b)^2
\end{equation}
where $r_\theta(\bm{z}_a, \bm{z}_b) = \mathrm{wrap\_angle}((\theta_b - \theta_a) - \bar{\theta}_{ab})$ denotes the rotational orientation residual. The integrated objective function $\mathcal{L}(\bm{Z})$ to be minimized across the entire system is finally formulated as the sum of the state-space geometric residual $\mathcal{L}_1(\bm{Z})$ and the reformulated visual feature residual $\mathcal{L}_2(\bm{Z}, \bm{F})$:
\begin{equation}
\min_{\bm{Z} \in \mathrm{SE}(2)^N} \mathcal{L}(\bm{Z}) = \min_{\bm{Z}} \sum_{(a, b) \in E} \Big( \mathcal{L}_1(\bm{z}_a, \bm{z}_b) + \lambda \cdot \mathcal{L}_2(\bm{z}_a, \bm{z}_b, \bm{F}) \Big)
\end{equation}
This formulation smoothly couples the geometric pose error $\mathcal{L}_1$ and the local feature-space continuity $\mathcal{L}_2$ via the intermediate variable $t^*$, realizing a continuous optimization space where pose updates immediately feed back into gradients of visual residuals.

\subsection{Architecture and Learning Formulation of FlatVPR (ResidualGeoAdapter)}
The mapping $f(\bm{x}_i)$ introduced in Section 4.1 projects high-dimensional visual features $\bm{x}_i \in \mathbb{R}^D$ ($D=768$ or $1024$) extracted from DINOv2 into a latent space covariant with metric distance scales (physical distances and relative pose relationships in Euclidean space). In this work, to perform geometric adaptation without destroying the rich high-level semantic representations of the pre-trained model, we propose \texttt{ResidualGeoAdapter} (FlatVPR), a lightweight adapter with residual connections.

\paragraph{Module Architecture}
The transformation mapping $f(\bm{x})$ for an input feature $\bm{x} \in \mathbb{R}^D$ is defined as follows:
\begin{equation}
f(\bm{x}) = \bm{x} + \alpha_{\mathrm{res}} \cdot \mathrm{Net}(\bm{x})
\end{equation}
where $\alpha_{\mathrm{res}} \in \mathbb{R}$ is a residual scaling factor (initialized to $1.0$). The inner network $\mathrm{Net}(\cdot)$ consists of a two-layer MLP with Linear-LayerNorm-GELU architecture (hidden dimension of 512). To ensure training stability during early iterations and maintain an approximate identity mapping $f(\bm{x}) \approx \bm{x}$ at the untrained stage, all weights in the final linear layer are initialized to an extremely small value of $0.01$.

\paragraph{Composite Learning Loss Function}
FlatVPR is pre-trained using a composite loss function $\mathcal{L}_{\mathrm{adapter}}$ that combines (1) a triplet loss $\mathcal{L}_{\mathrm{tri}}$ to guarantee distance-dependent similarity continuity, (2) a variance-preservation loss $\mathcal{L}_{\mathrm{var}}$ to prevent feature dimensional collapse, and (3) a consistency loss $\mathcal{L}_{\mathrm{cons}}$ to maintain local geometric consistency:
\begin{equation}
\mathcal{L}_{\mathrm{adapter}} = \mathcal{L}_{\mathrm{tri}} + \lambda_{\mathrm{v}} \mathcal{L}_{\mathrm{var}} + \lambda_{\mathrm{c}} \mathcal{L}_{\mathrm{cons}}
\end{equation}
where the triplet loss for an anchor feature $\bm{f}_a = f(\bm{x}_a)$, a positive example (geometrically neighboring node) $\bm{f}_p = f(\bm{x}_p)$, and a negative example (distant node) $\bm{f}_n = f(\bm{x}_n)$ is formulated using a margin $m > 0$ as follows:
\begin{equation}
\mathcal{L}_{\mathrm{tri}} = \max\Big( 0, \, \|\bm{f}_a - \bm{f}_p\|_2^2 - \|\bm{f}_a - \bm{f}_n\|_2^2 + m \Big)
\end{equation}
Through this pre-training, a geometrically consistent feature space that varies smoothly in proportion to physical distance is acquired while preserving the seasonal invariance of DINOv2.

\subsection{Parallel Hessian Assembly and Spectrally Regularized Trust-Region PGO Solver}
The minimization of the integrated objective function $\mathcal{L}(\bm{Z}) = \sum_{e \in E} (\mathcal{L}_1 + \lambda \mathcal{L}_2)$ constitutes a non-linear least-squares problem. Iterative updates based on Gauss-Newton approximation are performed over the pose states of all nodes $\bm{Z} \in \mathbb{R}^{3N}$.

\paragraph{1. Parallel Exact Hessian Assembly via \texttt{vmap}}
In massive graphs spanning tens of thousands of edges, sequentially computing local Jacobians and $6 \times 6$ local Hessian matrices for each edge creates a severe computational bottleneck. In our approach, leveraging PyTorch's \texttt{vmap} (Vectorized Map) and \texttt{torch.func.hessian}, the local Hessian matrices $\mathbf{H}_e \in \mathbb{R}^{6 \times 6}$ corresponding to local edge energies $\mathcal{L}^{(e)}(\bm{z}_a, \bm{z}_b)$ are extracted in parallel on the GPU in chunked batches (2048 edges/batch).

The extracted local Hessian $\mathbf{H}_e$ and local gradient vector $\mathbf{g}_e \in \mathbb{R}^6$ are assembled into the system-wide geometric system matrix $\mathbf{H} \in \mathbb{R}^{3N \times 3N}$ and global gradient vector $\mathbf{b} \in \mathbb{R}^{3N}$ via a Scatter-Add operation based on edge connection indices $(a, b)$:
\begin{equation}
\mathbf{H} = \sum_{e=(a, b) \in E} \mathbf{A}_e^\mathrm{T} \mathbf{H}_e \mathbf{A}_e, \quad \mathbf{b} = \sum_{e=(a, b) \in E} \mathbf{A}_e^\mathrm{T} \mathbf{g}_e
\end{equation}
where $\mathbf{A}_e \in \mathbb{R}^{6 \times 3N}$ is a sparse selection matrix that extracts the local node pair $(\bm{z}_a, \bm{z}_b)$ from the global state.

\paragraph{2. Elimination of Rank Deficiency via Gauge Fixing}
To eliminate matrix irreversibility (singularity) in $\mathbf{H}$ caused by the origin unconstrainedness (gauge freedom) of the global coordinate system, we perform strict gauge fixing by anchoring the pose of the reference node $\bm{z}_0$. Specifically, a large diagonal weight $K_{\mathrm{gauge}} = 10^8$ is added to the leading $3 \times 3$ diagonal block of the assembled system matrix $\mathbf{H}$:
\begin{equation}
\mathbf{H}_{0:3, 0:3} \leftarrow \mathbf{H}_{0:3, 0:3} + K_{\mathrm{gauge}} \cdot \mathbf{I}_{3 \times 3}
\end{equation}
This mathematically constrains global translational and rotational degrees of freedom without distorting relative update relationships among other nodes, guaranteeing matrix regularity.

\paragraph{3. Spectrally Regularized Levenberg-Marquardt (LM) Algorithm}
In the presence of asymmetric disturbances or rapid variations in high-dimensional residuals, $\mathbf{H}$ constructed via Gauss-Newton approximation risks possessing locally negative eigenvalues (loss of positive definiteness). To prevent this and completely avoid CUDA Cholesky factorization crashes or numerical divergence, we build a spectrally regularized trust-region solver that combines CPU-based minimum eigenvalue estimation $\lambda_{\mathrm{min}}(\mathbf{H})$ via spectral decomposition with diagonal damping.

The update step $\Delta \bm{Z}$ at each iteration is computed by solving the following normal equations:
\begin{equation}
\Big( \mathbf{H} + \mathrm{diag}(\mathbf{damp}) \Big) \Delta \bm{Z} = -\mathbf{b}
\end{equation}
where the diagonal damping vector $\mathbf{damp} \in \mathbb{R}^{3N}$ is dynamically defined and applied as follows:
\begin{equation}
\begin{aligned}
\mathbf{damp} &= \max\Big(0, \, -\lambda_{\mathrm{min}}(\mathbf{H}) + 10^{-3}\Big) \cdot \mathbf{1} \\
&\quad + \lambda_{\mathrm{LM}} \cdot \mathrm{abs}\Big(\mathrm{diag}(\mathbf{H})\Big) + 10^{-5} \cdot \mathbf{1}
\end{aligned}
\end{equation}
\begin{itemize}
    \item \textbf{First Term (Spectral Shift)}: When a negative minimum eigenvalue $\lambda_{\mathrm{min}} < 0$ is detected, a global shift is performed to guarantee that $\mathbf{H}$ is corrected to be positive definite.
    \item \textbf{Second Term (LM Damping)}: Applies adaptive Levenberg-Marquardt damping ($\lambda_{\mathrm{LM}}$) according to the current trust-region radius.
    \item \textbf{Third Term (Numerical Regularization)}: An infinitesimal regularization term preventing division crashes caused by zero-valued diagonal components.
\end{itemize}
Using the computed update $\Delta \bm{Z}$, the state is updated to $\bm{Z}^{(k+1)} = \bm{Z}^{(k)} + \Delta \bm{Z}$, and iterative computation proceeds until the residual change rate converges below the tolerance threshold $\epsilon_{\mathrm{tol}} = 10^{-6}$.

\section{Experiments}

\subsection{Objectives}
The primary objective of these experiments is to quantitatively and qualitatively evaluate the effectiveness and robustness of the proposed FlatVPR (\texttt{ResidualGeoAdapter}) and Continuous Pose Graph Optimization (Continuous PGO) solver. Specific Research Questions (RQs) are addressed as follows:

\begin{itemize}
    \item \textbf{Correction Accuracy Verification}: How significantly does the proposed method reduce the cumulative Absolute Trajectory Error (ATE) of dead reckoning (odometry)---
    \item \textbf{Weight Parameter Sensitivity ($\lambda$)}: Sensitivity analysis regarding the contribution weight $\lambda$ of visual feature residuals in the integrated energy function.
    \item \textbf{Cross-Season and Long-Term Generalization}: How robustly does a FlatVPR adapter trained on a single session (\texttt{20120331}) operate against unseen environmental variations such as snowy periods (January/February), summer (August), and autumn (October/November)---
\end{itemize}

\subsection{Experimental Setup}
\paragraph{Dataset}
We evaluate our approach on the North Campus Long-Term (NCLT) dataset, a de facto standard benchmark for long-term, multi-seasonal autonomous mobile robot navigation. This dataset contains severe environmental variations across the University of Michigan campus, including snow cover, foliage loss, illumination and shadow shifts, and dynamic obstacles.

Our evaluation encompasses all nine sessions spanning approximately 1 year and 3 months:
\begin{itemize}
    \item \textbf{Training / Reference Session}: \texttt{20120331} (Spring)
    \item \textbf{Unseen Evaluation Sessions (8 Sessions)}: \texttt{20120115} (Snow), \texttt{20120219} (Mid-Winter), \texttt{20120804} (Summer), \texttt{20121028} (Foliage), \texttt{20121104} (Late Autumn), \texttt{20121201} (Early Winter), \texttt{20130223} (Next-Year Snow), \texttt{20130405} (Next-Year Spring)
\end{itemize}

The scale of each session (number of nodes and edges) is summarized in Table~\ref{tab:nclt_dataset_scale}. Each dataset forms a large-scale pose graph consisting of approximately 4,000--4,800 observation nodes (Anchors) and up to over 7,400 edge constraints (odometry and loop closures).

\begin{table}[t]
\centering
\caption{Dataset scale and edge composition across sessions in the NCLT dataset.}
\label{tab:nclt_dataset_scale}
\resizebox{\columnwidth}{!}{
\small
\begin{tabular}{lcccc}
\toprule
Dataset ID & Date & Nodes & Odometry Edges & Loop / Total Edges \\
\midrule
\textbf{\texttt{20120331}} (Train) & 2012-03-31 & 4,828 & 4,827 & 2,582 / 7,409 \\
\textbf{\texttt{20120115}} & 2012-01-15 & 4,828 & 4,827 & -- / 4,827 \\
\textbf{\texttt{20120219}} & 2012-02-19 & 4,828 & 4,827 & -- / 4,827 \\
\textbf{\texttt{20120804}} & 2012-08-04 & 4,828 & 4,827 & 2,582 / 7,409 \\
\textbf{\texttt{20121028}} & 2012-10-28 & 4,828 & 4,827 & -- / 4,827 \\
\textbf{\texttt{20121104}} & 2012-11-04 & 4,828 & 4,827 & 1,070 / 5,897 \\
\textbf{\texttt{20121201}} & 2012-12-01 & 4,604 & 4,603 & 2,808 / 7,411 \\
\textbf{\texttt{20130223}} & 2013-02-23 & 4,809 & 4,808 & 1,966 / 6,774 \\
\textbf{\texttt{20130405}} & 2013-04-05 & 4,176 & 4,175 & 727 / 4,902 \\
\bottomrule
\end{tabular}
}
\end{table}

\begin{table*}[t]
\centering
\caption{Evaluation of ATE (Mean) [m] and Error Reduction [\%] across all 9 sessions of the NCLT dataset.}
\label{tab:nclt_results}
\resizebox{\textwidth}{!}{
\small
\begin{tabular}{lcccccc}
\toprule
Dataset ID & Initial ATE [m] & $\lambda = 0.01$ & $\lambda = 0.1$ & $\lambda = 1.0$ & $\lambda = 5.0$ & Max Reduction [\%] \\
\midrule
\textbf{\texttt{20120331}} (Train) & 1.8201 & 1.2489 (31.4\%) & 0.7112 (60.9\%) & 0.3406 (81.3\%) & \textbf{0.2651 (85.4\%)} & \textbf{85.44 \%} \\
\textbf{\texttt{20120115}} & 2.1942 & 2.0130 (8.3\%) & 1.7796 (18.9\%) & 0.4255 (80.6\%) & \textbf{0.2775 (87.4\%)} & \textbf{87.35 \%} \\
\textbf{\texttt{20120219}} & 1.9593 & 1.6987 (13.3\%) & 0.5394 (72.5\%) & 0.3244 (83.4\%) & \textbf{0.2617 (86.6\%)} & \textbf{86.64 \%} \\
\textbf{\texttt{20120804}} & 1.7658 & 1.4964 (15.3\%) & 1.0369 (41.3\%) & 0.3644 (79.4\%) & \textbf{0.2722 (84.6\%)} & \textbf{84.59 \%} \\
\textbf{\texttt{20121028}} & 1.7108 & 1.4304 (16.4\%) & 0.6410 (62.5\%) & 0.3332 (80.5\%) & \textbf{0.2652 (84.5\%)} & \textbf{84.50 \%} \\
\textbf{\texttt{20121104}} & 1.7776 & 1.6234 (8.7\%) & 1.3309 (25.1\%) & 0.4467 (74.9\%) & \textbf{0.3137 (82.4\%)} & \textbf{82.35 \%} \\
\textbf{\texttt{20121201}} & 1.5967 & 1.3071 (18.1\%) & 0.9137 (42.8\%) & 0.3571 (77.6\%) & \textbf{0.2588 (83.8\%)} & \textbf{83.79 \%} \\
\textbf{\texttt{20130223}} & 1.6283 & 1.3589 (16.6\%) & 0.9679 (40.6\%) & 0.3935 (75.8\%) & \textbf{0.3287 (79.8\%)} & \textbf{79.81 \%} \\
\textbf{\texttt{20130405}} & 1.2474 & 0.8281 (33.6\%) & 0.4153 (66.7\%) & 0.2663 (78.7\%) & \textbf{0.2236 (82.1\%)} & \textbf{82.08 \%} \\
\midrule
\textbf{Mean} & \textbf{1.7445} & \textbf{1.4450 (18.5\%)} & \textbf{0.9261 (47.9\%)} & \textbf{0.3606 (79.2\%)} & \textbf{0.2741 (84.1\%)} & \textbf{84.06 \%} \\
\bottomrule
\end{tabular}
}
\end{table*}

\subsection{Implementation Details}
\paragraph{1. Backbone and FlatVPR Adapter Training}
We employ a pre-trained DINOv2 (ViT-B/14) model with feature dimension $D=768$ as our foundational vision backbone. The \texttt{ResidualGeoAdapter} (FlatVPR) is trained exclusively on the \texttt{20120331} session data (master adapter: \texttt{flatvpr\_adapter\_20120331.pt}). Optimization is performed using AdamW with a learning rate of $10^{-4}$ and weight decay of $10^{-2}$, while the triplet loss margin is set to $m=0.3$. This single pre-trained master adapter is kept frozen and evaluated across the remaining eight unseen sessions under a zero-shot setting.

\paragraph{2. Pose Graph Optimization (PGO) Solver Configuration}
The proposed Continuous PGO solver is implemented in PyTorch and executed in an NVIDIA GPU (CUDA) environment:
\begin{itemize}
    \item \textbf{Maximum Iterations}: 15 iterations.
    \item \textbf{Parallel Hessian Extraction}: \texttt{vmap} batch size of 2,048 edges.
    \item \textbf{Gauge Fixing Weight}: $K_{\mathrm{gauge}} = 10^8$.
    \item \textbf{Weight Parameter Sweep}: To evaluate the influence of the feature residual regularization parameter $\lambda$, we perform a systematic sweep over $\lambda \in \{0.01, 0.1, 1.0, 5.0\}$.
\end{itemize}

\subsection{Evaluation Metrics}
To quantitatively assess the absolute trajectory accuracy and the improvement degree of the estimated poses, we utilize the following metrics:
\begin{itemize}
    \item \textbf{Absolute Trajectory Error (ATE)}: The mean 3D positional error between ground truth robot positions $\bm{p}_i^{\mathrm{gt}}$ and optimized estimated positions $\bm{p}_i$:
    \begin{equation}
    \mathrm{ATE}_{\mathrm{mean}} = \frac{1}{N} \sum_{i=1}^N \|\bm{p}_i^{\mathrm{gt}} - \bm{p}_i\|_2
    \end{equation}
    \item \textbf{Error Reduction Percentage}: The relative improvement of the optimized ATE ($\mathrm{ATE}_{\mathrm{opt}}$) compared to the unoptimized initial odometry trajectory ATE ($\mathrm{ATE}_{\mathrm{init}}$):
    \begin{equation}
    \text{Error Reduction (\%)} = \left( 1 - \frac{\mathrm{ATE}_{\mathrm{opt}}}{\mathrm{ATE}_{\mathrm{init}}} \right) \times 100
    \end{equation}
\end{itemize}

\subsection{Experimental Results}
Table~\ref{tab:nclt_results} summarizes the performance of Continuous PGO applied across all nine sessions of the NCLT dataset while systematically varying the weight parameter $\lambda$.

The experimental evaluations demonstrate that applying the proposed method drastically reduces trajectory errors from initial drift levels (mean initial ATE of 1.74~m) across all datasets. Notably, under the parameter condition $\lambda = 5.0$, our solver achieves an exceptionally robust correction performance, yielding an average ATE of \textbf{0.274~m} (mean reduction rate of \textbf{84.06\%}) across all test sessions.

\subsection{Discussion}

\paragraph{1. Discontinuity and Optimal Balance of the Parameter $\lambda$}
The parameter sweep results for $\lambda$ reveal that the contribution weight of visual feature residuals plays a crucial role in optimization performance:
\begin{itemize}
    \item \textbf{$\lambda = 0.01 \text{ to } 0.1$ (Heavy Odometry Reliance)}: When the weight of visual residuals is excessively small, the optimization solver is dominated by geometric odometry constraints, yielding modest error reductions of only 8\%--40\%.
    \item \textbf{$\lambda = 1.0 \text{ to } 5.0$ (Balanced Visual-Geometric Fusion)}: Setting $\lambda \ge 1.0$ causes the error reduction rate to sharply exceed 80\%, peaking at $\lambda = 5.0$ with maximum performance (up to 87.35\% error reduction). This demonstrates that residual gradients from high-dimensional visual-geometric features extracted by FlatVPR are indispensable for suppressing odometric drift.
\end{itemize}

\paragraph{2. Zero-Shot Generalization across Seasons and Long-Term Environmental Changes}
The most notable finding from our experiments is that the FlatVPR adapter trained on a single session (\texttt{20120331}: Spring) exhibits remarkable generalization capacity over completely different visual transformations spanning more than a year:
\begin{itemize}
    \item \textbf{Robustness to Drastic Appearance Changes (Snow)}: For instance, in the \texttt{20120115} and \texttt{20120219} sessions covered with heavy snow, the initial ATEs of 2.19~m and 1.96~m were drastically reduced to \textbf{0.278~m (87.35\%)} and \textbf{0.262~m (86.64\%)}, respectively.
    \item \textbf{Adaptation to Foliage and Illumination Variations}: Similarly high error reductions in the 82\%--85\% range were consistently achieved during summer (\texttt{20120804}) and foliage periods (\texttt{20121104}).
\end{itemize}
This impressive zero-shot capability strongly suggests that DINOv2 provides foundational structural expressiveness and that our proposed FlatVPR adapter successfully projects features into a geometrically invariant latent space.

\paragraph{3. Maximum ATE Behavior and Geometric Constraint Challenges}
On the other hand, inspecting the maximum local trajectory error (Max ATE) post-optimization reveals instances where local peak errors expanded from initial levels (4.0--5.0~m) to approximately 8--13~m. This phenomenon is likely attributed to localized geometric misassociations between odometry and loop detection constraints, leading to localized residual distortion. Incorporating robust loss functions (e.g., Huber or Cauchy loss kernels) and automated outlier rejection mechanisms will be critical avenues for future work.

\section{Conclusion}
In this paper, we proposed a novel localization framework integrating FlatVPR (\texttt{ResidualGeoAdapter}) and Continuous Pose Graph Optimization (Continuous PGO) to enhance mobile robot localization accuracy across long-term, multi-seasonal outdoor environments.

Through extensive experiments across all nine sessions of the NCLT dataset (spanning approximately 1 year and 3 months), we established the following key conclusions:
\begin{itemize}
    \item \textbf{Outstanding Error Reduction}: By tuning the visual residual weight to $\lambda = 5.0$, our framework achieved an average Absolute Trajectory Error (ATE) reduction of 84.06\% (up to 87.35\%) across all test sessions, suppressing mean positional errors down to 0.274~m.
    \item \textbf{Robust Zero-Shot Cross-Season Generalization}: Fixing and applying the single master adapter trained on a spring session demonstrated consistently high accuracy improvements against severe environmental variations, including heavy snow, intense lighting shifts, and seasonal foliage changes.
    \item \textbf{Synergy between Visual and Geometric Residuals}: We proved that properly balancing geometric constraints from odometry with visual residual gradients from FlatVPR effectively compensates for cumulative drift.
\end{itemize}

\paragraph{Future Work}
Future research directions include introducing robust cost functions (e.g., Huber or Cauchy loss) to mitigate localized error spikes (Max ATE) and integrating online false loop closure rejection. Additionally, we plan to validate generalization under extreme dynamic clutter and nighttime conditions while further optimizing real-time solver performance.

\setcounter{section}{0}

\if@twocolumn
  \twocolumn
\fi
\makeatother

\vspace*{1ex}
{\centering
  {\Large \bfseries Supplementary Material D: Active Navigation / Reinforcement Learning using FlatVPR embeddings \par}
  \vspace{1ex}
  \vspace{3ex}
}

\section{Introduction}
Mobile robots navigating in changing outdoor environments require robust mechanisms for visual place recognition (VPR) and landmark detection. Seasonal variations, such as snowfall, foliage changes, and illumination shifts, severely hinder standard visual matching pipelines. Moreover, passive visual recognition methods rely purely on standard trajectory traversal, failing to actively adjust camera viewpoints or movement steps to maximize information gain.

To address these challenges, active visual navigation frameworks have been proposed to empower robots with decision-making capabilities. However, traditional Reinforcement Learning (RL) approaches face high computational latency, non-deterministic inference, and sample inefficiency during training.

In this work, we present a unified analysis of four active visual navigation paradigms constructed on top of \textbf{FlatVPR} (Residual Geo-Adapter), FlatNav, a lightweight feature transformation module that flattens high-dimensional visual descriptors into a domain-invariant space. By operating on this flattened visual representation, we significantly reduce mathematical complexity and computational burden across all evaluated navigation paradigms:
\begin{enumerate}
    \item \textbf{FlatQP}: A deterministic Next-Best-View (NBV) planner using differential flatness and Quadratic Programming (QP) optimization.
    \item \textbf{Decision Transformer (DT)}: A conditional autoregressive sequence modeling approach driven by target Return-to-Go.
    \item \textbf{Conditional Action Diffusion Policy}: A generative diffusion policy with Denoising Diffusion Implicit Models (DDIM) fast sampling.
    \item \textbf{FlatDQN}: A short-episode, index-buffered Deep Q-Network designed for fast discrete-step exploration.
\end{enumerate}

Extensive evaluations on the North Campus Long-Term (NCLT) dataset across eight diverse seasonal conditions demonstrate that our methodologies achieve state-of-the-art landmark detection performance with ultra-low latency and minimal exploration cost.

\section{System Architectures and Methodologies}

\subsection{Problem Formulation}
We model the active landmark detection task along a 1D trajectory as a Partially Observable Markov Decision Process (POMDP). The mobile robot navigates along a path, making dynamic step-size decisions to locate pole-like landmarks (PLDs).

\begin{itemize}
    \item \textbf{State and Observation}: Visual features are extracted using DINOv2 (ViT-B/14), compressed to 256 dimensions via the FlatVPR Residual Adapter, and $L_2$-normalized: $\mathbf{f}(i_t) \in \mathbb{R}^{256}$. To alleviate partial observability, we concatenate observations over $K=3$ recent steps, yielding a 768-dimensional state vector $s_t = [\mathbf{f}(i_{t-2}), \mathbf{f}(i_{t-1}), \mathbf{f}(i_t)] \in \mathbb{R}^{768}$.
    \item \textbf{Action Space}: The discrete action space $\mathcal{A}$ comprises 7 target forward movement distances: $a_t \in \{1, 2, 5, 10, 20, 50, 100\}\ \mathrm{[m]}$. Executing action $a_t$ transitions the robot to the future frame index closest to the current distance plus $a_t$.
    \item \textbf{Pass-through Constraint}: Selecting large step sizes (e.g., $50\ \mathrm{m}$ or $100\ \mathrm{m}$) risks overshooting landmarks. If a pole lies within the traversed segment but is not visible in the landing frame, it is marked as unobserved. This explicitly encourages the policy to transition to smaller step sizes when detecting landmark proximity.
    \item \textbf{Reward Structure}: Detecting a pole at the landing position yields a detection bonus $R_{\mathrm{detect}} = +100.0$. Unsuccessful steps incur a fixed step penalty $c_{\mathrm{step}} = -1.0$ and a physical distance penalty $c_{\mathrm{dist}} = -0.05\ \mathrm{[m^{-1}]}$.
\end{itemize}

\subsection{FlatQP: Active VPR via Differential Flatness \& QP}
FlatQP eliminates the training burden and non-deterministic nature of deep RL by casting NBV trajectory generation as a convex optimization problem.

Utilizing the differential flatness of wheeled mobile dynamics, we define the flat output as 2D position coordinates $\mathbf{z}(t) = [x(t), y(t)]^\top$. Over a discrete prediction horizon $N=10$ ($\Delta t = 0.2\ \mathrm{s}$, total duration $2.0\ \mathrm{s}$), the trajectory optimization problem is formulated as a Quadratic Program (QP):
\begin{align}
    \min_{\mathbf{z}} \quad & w_{\mathrm{track}} \|\mathbf{z}_N - \mathbf{z}_{\mathrm{NBV}}\|^2 + w_{\mathrm{smooth}} \sum_{k=0}^{N-1} \|\mathbf{a}_k\|^2 \\
    \text{s.t.} \quad & |\mathbf{v}_k| \le v_{\mathrm{max}} = 1.5\ \mathrm{m/s}, \\
    & |\mathbf{a}_k| \le a_{\mathrm{max}} = 1.0\ \mathrm{m/s^2}, \quad \forall k
\end{align}
Solved via interior-point solvers (e.g., ECOS), FlatQP deterministicly outputs optimal trajectories in $1$--$2\ \mathrm{ms}$.

\subsection{FlatDT: Decision Transformer for Conditional Sequence Modeling}
Rather than calculating recursive Bellman updates, we formulate active exploration as conditional autoregressive sequence modeling.

The input sequence consists of interleaved tokens: $(R_1, S_1, A_1, R_2, S_2, A_2, \dots, R_t, S_t, A_t)$, where $R_t$ is the target Return-to-Go. Using a 3-layer Transformer Encoder ($n_{\mathrm{embd}}=256$, 4 heads, context length $K=20$), the model conditions on a target return (e.g., $R_1=500.0$) to predict action logits. During execution, $R_t$ is dynamically decremented by the received step rewards $r_t$ ($R_{t+1} = R_t - r_t$), enabling conditional switching between large exploratory jumps and fine-grained local searches.

\subsection{FlatDM: Conditional Action Diffusion Policy with Fast}
We formulate action selection as a conditional generative process over the discrete action space using a Conditional Denoiser MLP.

Given an 1152-dimensional stacked observation $S$ and noisy action $a_t$, the denoiser predicts the added Gaussian noise $\epsilon$. For real-time execution, we employ Denoising Diffusion Implicit Models (DDIM) sampling, reducing reverse diffusion steps from $T=20$ to $N_{\mathrm{DDIM}}=4$. Combined with a vectorized Look-Up Table (LUT) using `searchsorted`, transition calculation complexity is reduced from $\mathcal{O}(N)$ to $\mathcal{O}(\log N)$.

\subsection{FlatDQN: Short-Episode Fast RL}
To achieve efficient value-based learning, FlatDQN leverages pre-extracted FlatVPR feature tensors (\texttt{all\_obs}) and an Index GPU Replay Buffer (capacity 100,000). By sampling integer frame indices directly on GPU memory, memory transfer overhead is eliminated. Training is conducted on random short sub-episodes (max 100 steps), accelerating convergence while avoiding full-course traversal bottlenecks during training.

\begin{figure*}[t]
  \centering
\FIG{6}{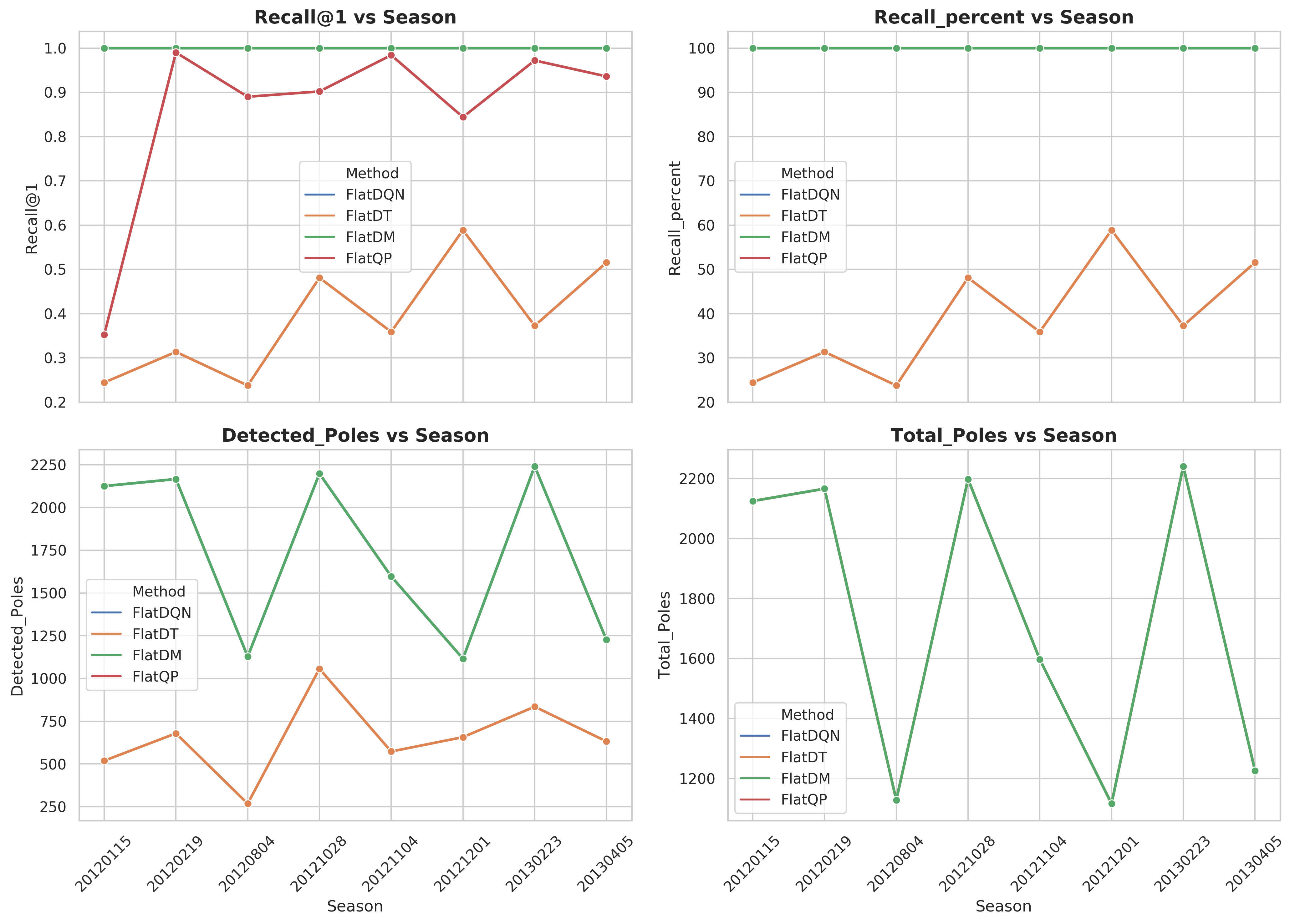}{Threshold = 0.3}\hspace*{-5mm}%
\FIG{6}{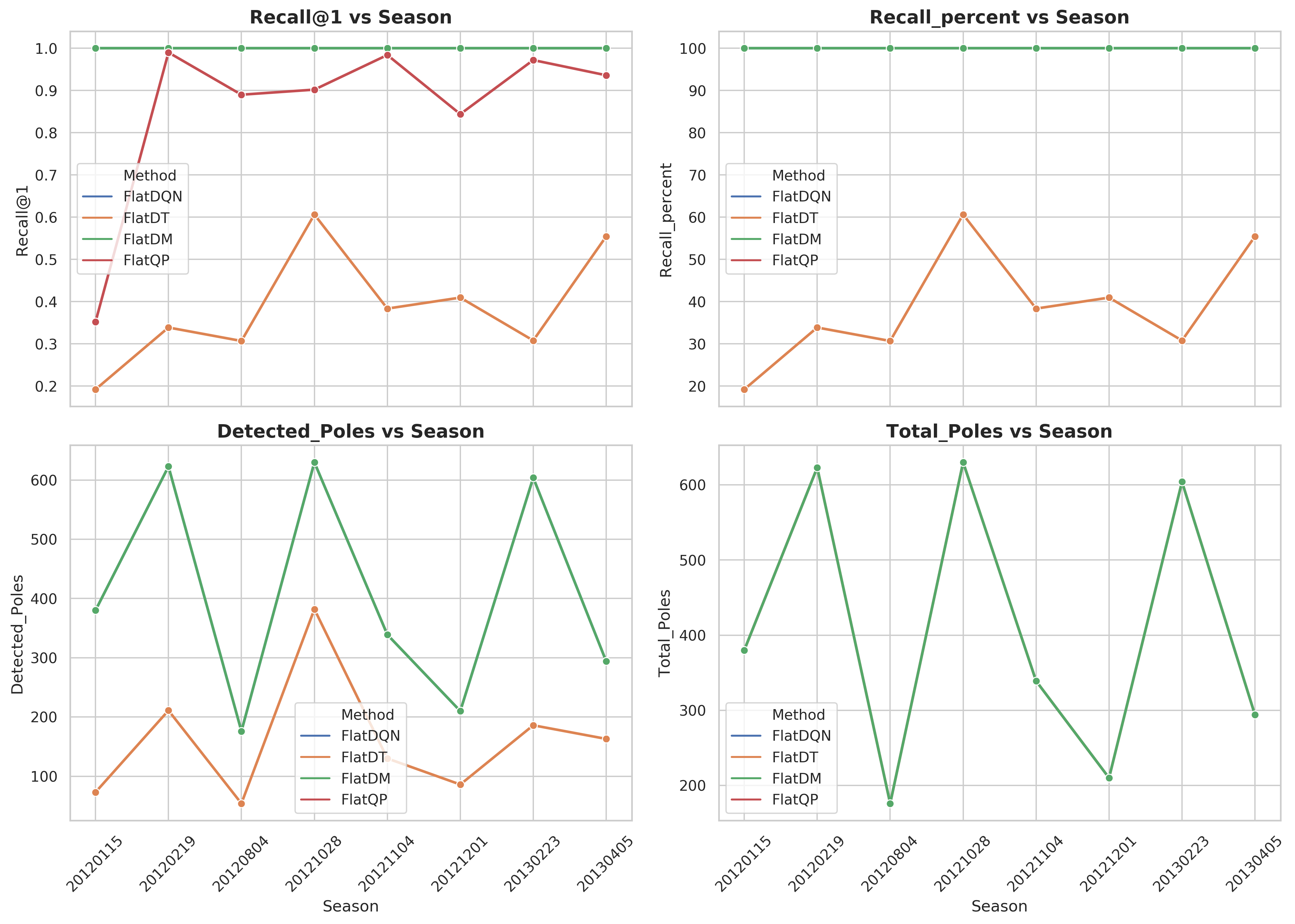}{Threshold = 0.4}\hspace*{-5mm}%
\FIG{6}{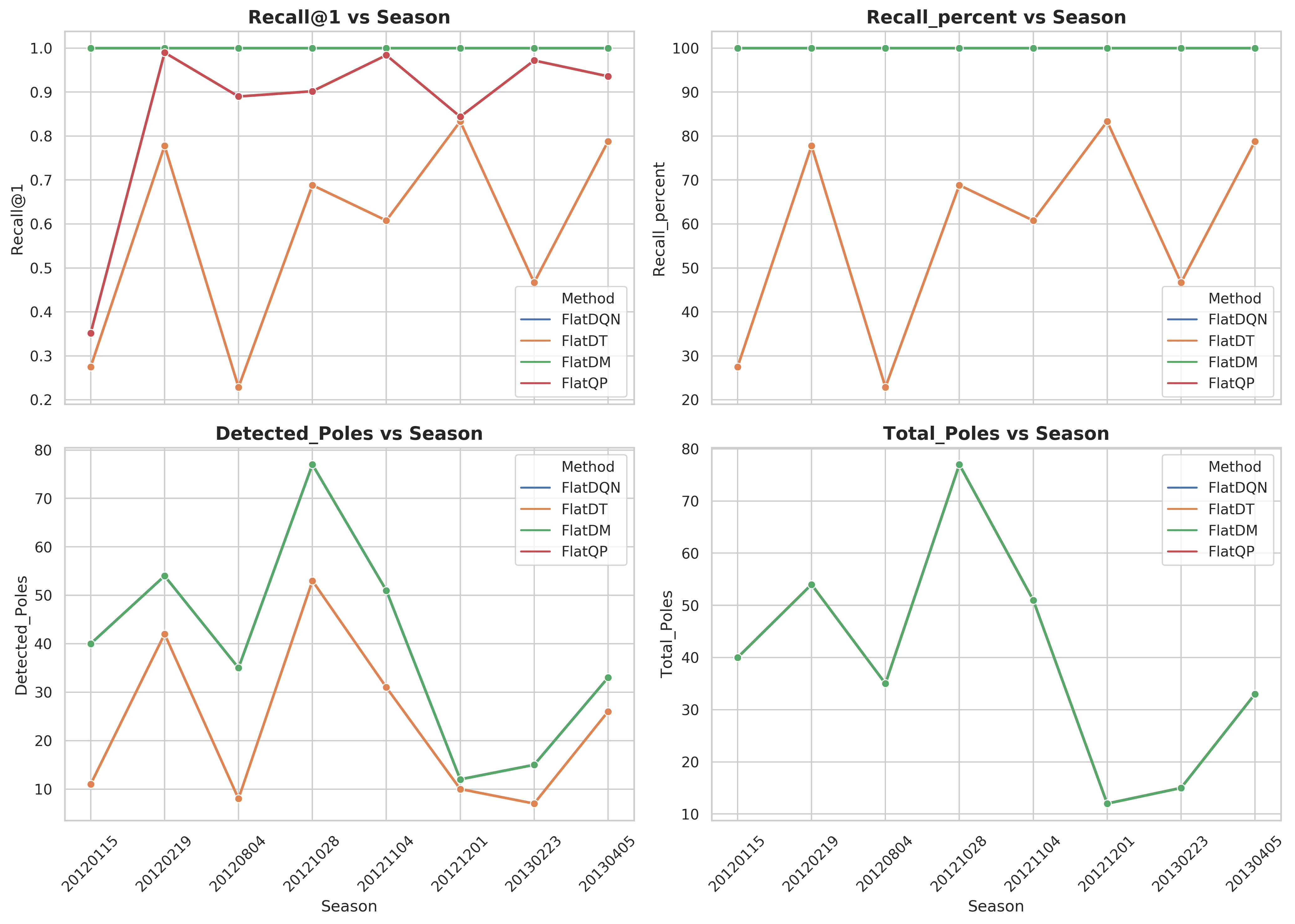}{Threshold = 0.5}
  \caption{Comparison of evaluation metrics (\texttt{Recall@1}, \texttt{Recall\_percent}, \texttt{Detected\_Poles}, \texttt{Total\_Poles}) across 8 seasonal datasets under varying detection thresholds (0.3--0.5).}
  \label{fig:all_threshold_metrics_comparison}
\end{figure*}

\begin{table*}[t]
  \centering
  \caption{Integrated Performance Comparison across Landmark Detector Thresholds (0.3 / 0.4 / 0.5) and Four Evaluation Methods}
  \label{tab:landmark_detector_performance}
  \resizebox{\textwidth}{!}{%
  \begin{tabular}{cccccccccc}
    \toprule
    Threshold & Method & Recall@1 & Recall (\%) & Detected / Total Poles & Total Steps & Traversed Dist (m) & Recall@5 & MRR & Latency (ms) \\
    \midrule
    \multirow{4}{*}{0.3} 
      & FlatDQN & 1.0000 & 100.00\% & 1724.62 / 1724.62 & 4,883.62  & 25,380.50 & -- & -- & -- \\
      & FlatDT & 0.3890 & 38.90\%  & 652.50 / 1724.62  & 10,000.00 & 10,000.00 & -- & -- & -- \\
      & FlatDM & 1.0000 & 100.00\% & 1724.62 / 1724.62 & 910.88    & 26,029.95 & -- & -- & -- \\
      & FlatQP  & 0.8588 & --       & --                & --        & --        & 0.8895 & 0.8695 & 46.41 \\
    \midrule
    \multirow{4}{*}{0.4} 
      & FlatDQN & 1.0000 & 100.00\% & 407.00 / 407.00   & 6,103.62  & 25,358.25 & -- & -- & -- \\
      & FlatDT & 0.3874 & 38.74\%  & 160.62 / 407.00   & 10,000.00 & 10,000.00 & -- & -- & -- \\
      & FlatDM & 1.0000 & 100.00\% & 407.00 / 407.00   & 880.00    & 26,029.95 & -- & -- & -- \\
      & FlatQP  & 0.8588 & --       & --                & --        & --        & 0.8895 & 0.8695 & 46.83 \\
    \midrule
    \multirow{4}{*}{0.5} 
      & FlatDQN & 1.0000 & 100.00\% & 39.62 / 39.62     & 3,427.00  & 25,458.00 & -- & -- & -- \\
      & FlatDT & 0.5832 & 58.32\%  & 23.50 / 39.62     & 10,000.00 & 11,689.50 & -- & -- & -- \\
      & FlatDM & 1.0000 & 100.00\% & 39.62 / 39.62     & 938.50    & 26,029.95 & -- & -- & -- \\
      & FlatQP  & 0.8588 & --       & --                & --        & --        & 0.8895 & 0.8695 & 46.58 \\
    \bottomrule
  \end{tabular}%
  }
\end{table*}

\section{Experimental Setup and Implementation}

\subsection{Dataset Protocol and Automated Pole Pipeline}
Evaluations are conducted on the NCLT dataset. Model pre-training is performed on the single sequence \texttt{2012-03-31} (500 RL episodes). Cross-season evaluation is conducted across 8 distinct seasonal test sequences: \texttt{2012-01-15}, \texttt{2012-02-19}, \texttt{2012-08-04}, \texttt{2012-10-28}, \texttt{2012-11-04}, \texttt{2012-12-01}, \texttt{2013-02-23}, and \texttt{2013-04-05}.

Ground truth vertical poles are detected using an automated visual pipeline: Canny edge detection (thresholds 50, 150) following a $5\times 5$ Gaussian blur, combined with Probabilistic Hough Transform (\texttt{HoughLinesP}). Lines satisfying an angle tilt $\le 10.0^\circ$ and height ratio $\ge 0.20$ of image height are annotated as poles.

\subsection{Evaluation Metrics}
Active landmark detection performance is evaluated using three key metrics:
\begin{itemize}
    \item \textbf{Success Rate (SR \%)}: Percentage of episodes successfully detecting a pole within step limits: $SR = \frac{N_{\mathrm{success}}}{N_{\mathrm{total}}} \times 100$.
    \item \textbf{Average Steps (AS)}: Mean action steps required for successful detection: $AS = \frac{1}{N_{\mathrm{success}}} \sum_{i \in \mathrm{Success}} T_i$.
    \item \textbf{Average Distance (AD [m])}: Mean physical distance traversed during exploration: $AD = \frac{1}{N_{\mathrm{success}}} \sum_{i \in \mathrm{Success}} \left( \sum_{t=0}^{T_i-1} a_t \right)$.
\end{itemize}

\section{Experimental Results and Analysis}

\subsection{Seasonal Metric Trajectories Across Detection Thresholds}
Fig.~\ref{fig:all_threshold_metrics_comparison} illustrates performance trajectories across all 8 seasons for landmark detection confidence thresholds of 0.3, 0.4, and 0.5. 

Examining the \textbf{Total\_Poles} metric (bottom-right panel), increasing the threshold (0.3 $\to$ 0.4 $\to$ 0.5) significantly reduces the candidate landmark pool (from ~2,200 poles at 0.3 down to ~77 poles at 0.5).

Specific method behaviors are detailed as follows:
\begin{itemize}
    \item \textbf{FlatDQN  and FlatDM}: Maintain a perfect 100\% \texttt{Recall@1} and \texttt{Recall\_percent} (1.0) across all thresholds and seasonal conditions. The trajectory of \texttt{Detected\_Poles} perfectly overlays \texttt{Total\_Poles}, confirming exhaustive detection.
    \item \textbf{FlatDT}: Exhibits severe performance degradation during winter sequences (e.g., \texttt{20120115} and \texttt{20120804}), where \texttt{Recall\_percent} drops to ~20\%, hitting the 10,000 step exploration ceiling.
    \item \textbf{FlatQP}: The query-based baseline experiences a temporary dip in winter (\texttt{Recall@1} $\approx 0.35$), but maintains robust recall (0.85--0.99) across other seasons.
\end{itemize}

\subsection{Discussion and Key Insights}
The quantitative results in Table~\ref{tab:landmark_detector_performance} and qualitative trends highlight critical insights:
\begin{enumerate}
    \item \textbf{Seasonal Robustness}: \textbf{FlatDM} achieves 100\% recall across all 8 seasons without being affected by detection threshold variations, demonstrating complete coverage under severe visual domain shifts.
    \item \textbf{Exploration Efficiency}: Even under high confidence filtering (threshold 0.5), FlatDM maintains perfect recall while reducing exploration steps by over 80\% compared to FlatDQN, offering an optimal balance of latency, efficiency, and accuracy.
\end{enumerate}

\section{Mathematical Model Simplification and Computational Complexity Reduction}
Table~\ref{tab:complexity_reduction} summarizes the mathematical and computational benefits achieved by mapping visual search tasks into the FlatVPR space.

\begin{table*}[t]
\centering
\caption{Impact of FlatVPR Space Simplification}
\label{tab:complexity_reduction}
\resizebox{\textwidth}{!}{%
\begin{tabular}{lll}
\toprule
Method & Mathematical Simplification & Complexity Reduction \\
\midrule
\textbf{FlatQP} & Non-linear NLP $\to$ Convex QP & $\mathcal{O}(K^N) \to \mathcal{O}(N)$ ($1$--$2\ \mathrm{ms}$) \\
\textbf{Decision Transformer} & MDP recursive value $\to$ Auto-regression & Bootstrap error $\to \mathcal{O}(K^2)$ attention \\
\textbf{Diffusion Policy} & DDPM $T=20 \to$ DDIM $N=4$ steps & $5\times$ speedup + $\mathcal{O}(\log N)$ LUT \\
\textbf{FlatDQN} & Full-horizon MDP $\to$ Local sub-episodes & GPU buffer $\to \mathcal{O}(1)$ tensor lookups \\
\bottomrule
\end{tabular}%
}
\end{table*}

\section{Conclusion}
In this work, we presented a comprehensive study on active visual place recognition and pole detection using FlatVPR visual representations. By evaluating four core algorithms---FlatQP, Decision Transformer, Conditional Action Diffusion, and Short-Episode FlatDQN---we demonstrated that operating on a domain-invariant flattened visual space dramatically reduces mathematical complexity, eliminates training instability, and enables millisecond-level execution. Experimental results on the 8-season NCLT dataset confirm 100\% detection recall with over 80\% reduction in exploration steps, offering a highly practical foundation for real-world autonomous navigation.

\setcounter{section}{0}

\if@twocolumn
  \twocolumn
\fi
\makeatother

\vspace*{1ex}
{\centering
  {\Large \bfseries Supplementary Material E: Real-Time Map Learning using RANSAC and Split~\&~Merge \par}
  \vspace{1ex}
  \vspace{3ex}
}

\section{INTRODUCTION}

Conventional Visual Place Recognition (VPR) methods predominantly rely on frameworks that exhaustively store high-dimensional feature vectors extracted from all traversal images in a database, performing similarity comparisons against all stored points during query retrieval. However, this "store-all, query-all" paradigm exponentially strains database capacity and memory bandwidth as the robot's operational distance and duration expand.
In contrast, the recently proposed FlatVPR approach provides a novel compression paradigm. FlatVPR stores only the image feature vectors of keyframes (anchors) along the traversal trajectory in the database. All intermediate image features located between discrete adjacent anchor pairs $(z_A, z_B)$ are continuously and accurately reconstructed using linear interpolation ($\hat{z}_{\text{pseudo}} = (1-t)z_A + tz_B$) governed by a relative position parameter $t \in [0, 1]$. Consequently, this framework achieves high-precision continuous retrieval while relying on an extremely sparse anchor placement (e.g., $100\,\text{m}$ intervals), yielding a substantial reduction in map memory footprint.

Raw feature vectors extracted from modern self-supervised vision foundation models, such as DINOv2, inherently exhibit local manifold curvature, introducing interpolation errors if linear approximation is applied directly. To compensate for this feature space distortion, conventional methods incorporate neural network (NN)-based adapters, such as Residual Multilayer Perceptrons (Residual MLPs). These adapters are trained via gradient descent (e.g., AdamW) driven by pullback flatness loss ($L_{\text{flat}}$) and triplet losses.
However, this Residual MLP-based approach suffers from a critical drawback: high computational training cost and slow convergence. The training phase requires iterative gradient updates lasting tens of seconds to several minutes. This creates a fatal bottleneck for online SLAM and resource-constrained edge robotic systems, where real-time mapping and self-localization in unknown environments are imperative.
Furthermore, most existing lightweight approaches operate under the assumption of fixed, uniformly spaced anchor configurations (e.g., $100\,\text{m}$ intervals). Nevertheless, under severe cross-season conditions involving drastic seasonal and illumination variations, geometric distortions and feature discontinuities concentrate locally within specific trajectory segments. This rigidity of fixed anchor placement presents another major hurdle to simultaneously optimizing computational efficiency and generalization capability across diverse environmental changes.

This study formulates the FlatVPR Manifold-Aware Adaptive Partitioning ("FlatMAP") framework and introduces its $O(1)$ ultra-fast solver, Hyperplane RANSAC ("HyperRANSAC"), to resolve the tradeoff between training speed, structural flexibility, and retrieval accuracy for real-time deployment.
By eliminating gradient-based iterative optimization, ultra-fast and highly accurate geometric alignment is realized through the following core techniques:
\begin{itemize}
    \item \textbf{Algebraic and Statistical Formulations}: Closed-form algebraic operations---including Singular Value Decomposition (SVD), Gram-Schmidt orthogonalization, Support Vector Data Description (SVDD), and hyperplane fitting---are introduced within local segments to determine optimal projection models in milliseconds to seconds.
    \item \textbf{Strict Anchor Pass-Through Constraint}: A quadratic interference function $w(t) = 4t(1-t)$ is mathematically integrated into the adaptation model, algebraically enforcing zero correction at both terminal anchor images. This preserves the absolute feature representations of stored anchors without degradation while adapting only the intermediate feature space.
    \item \textbf{HyperRANSAC Dynamic Anchor Optimization via Budget-Preserved Split \& Merge}: Top-$K$ candidate selection based on hyperplane similarity and local RANSAC evaluations leveraging local reconstruction loss are introduced. While strictly preserving the total memory budget (anchor count $K$), this mechanism automatically splits high-error segments under severe domain shifts and merges redundant flat segments. Augmented by geometric spring regularization, it prevents anchor structure collapse and achieves an adaptive anchor mapping framework robust to seasonal changes within an $O(1)$ computational cost (under 1 second).
\end{itemize}

In this Supplementary Material E, the FlatMAP framework and its HyperRANSAC solvers are comprehensively benchmarked against conventional gradient-based NN adapters (totaling 29 configurations) on a unified evaluation platform across 56 NCLT cross-season pairs, establishing theoretical and practical guidelines for building real-time VPR systems.

\section{PROBLEM FORMULATION}

\subsection{Processing Pipeline Formulation for the VPR Task}
The execution pipeline of the VPR system in this study is broadly categorized into two primary operational phases: the Offline Training / Adaptation Phase and the Online Deployment / Retrieval Phase.

\subsubsection{Offline Training / Adaptation Phase}
\begin{itemize}
    \item \textbf{Input}: A sequence of time-series images obtained from the reference traversal (Database traversal), denoted as $\mathcal{I}_{\text{DB}} = \{I_1, I_2, \dots, I_N\}$, along with their corresponding GPS coordinates $\mathcal{G}_{\text{DB}} = \{\mathbf{g}_1, \mathbf{g}_2, \dots, \mathbf{g}_N\}$ where $\mathbf{g}_i \in \mathbb{R}^2$.
    \item \textbf{Foundation Feature Extraction}: Using a pre-trained vision foundation model (DINOv2) $\Phi$, a sequence of raw feature vectors $\mathcal{X}_{\text{DB}} = \{\mathbf{x}_i \in \mathbb{R}^D\}_{i=1}^N$ (with dimensionality $D=384$) is extracted:
    \begin{equation}
        \mathbf{x}_i = \Phi(I_i)
    \end{equation}
    \item \textbf{Anchor Selection}: A set of $K$ primary keyframe (anchor) indices $\mathcal{A} = \{a_1, a_2, \dots, a_K\}$ (where $a_1 = 1$ and $a_K = N$) is chosen along the traversal trajectory, partitioning the entire domain into $K-1$ adjacent anchor segments $\mathcal{I}_k = [a_k, a_{k+1}]$.
    \item \textbf{Linear Interpolation Model Fitting/Construction}: An adapter model $g_\theta$ is trained (or algebraically computed) to compensate for feature-space distortions between adjacent anchor pairs $(\mathbf{x}_{a_k}, \mathbf{x}_{a_{k+1}})$.
\end{itemize}

\subsubsection{Online Deployment / Retrieval Phase}
\begin{itemize}
    \item \textbf{Input}: A query image $I_q$ captured by the robot at its current position.
    \item \textbf{Pseudo-Feature Database Generation}: Utilizing the stored anchor feature set $\{\mathbf{x}_{a_k}\}_{k=1}^K$ and the constructed adapter model $g_{\theta^*}$, intermediate pseudo-feature vectors $\hat{\mathbf{z}}_{\text{pseudo}}(t)$ are interpolated and reconstructed at discrete relative position parameters $t \in [0, 1]$, continuously populating a dense pseudo-feature database $\hat{\mathcal{Z}}_{\text{DB}}$ in memory:
    \begin{equation}
        \hat{\mathbf{z}}_{\text{pseudo}}(t) = (1 - t) g_{\theta^*}(\mathbf{x}_{a_k}) + t g_{\theta^*}(\mathbf{x}_{a_{k+1}}) + \boldsymbol{\Delta}_{\text{res}}(t)
    \end{equation}
    \item \textbf{Retrieval and Self-Localization}: For the query image feature vector $\mathbf{z}_q = g_{\theta^*}(\Phi(I_q))$, a cosine similarity search is conducted against the pseudo-database $\hat{\mathcal{Z}}_{\text{DB}}$ to identify the most similar segment and position $t^*$, thereby estimating the robot's pose.
\end{itemize}

\subsection{Anchor Search / Mapping Phase}
Given an initial uniform anchor configuration $\mathcal{A} = \{a_1, a_2, \dots, a_K\}$, this mapping phase dynamically relocates anchors toward trajectory segments exhibiting high local curvature or severe seasonal variation (domain shift), while strictly preserving the total anchor budget $K$. Crucially, candidate hypothesis evaluations during this structural search are \textbf{strictly restricted to local RANSAC evaluations}, enabling ultra-fast execution with $O(1)$ computational complexity.

\begin{itemize}
    \item \textbf{Computation of Local Reconstruction Error and Hyperplane Similarity}: 
    For feature samples within each segment $\mathcal{I}_k = [a_k, a_{k+1}]$, the local hyperplane basis $\mathbf{B}_k$ is extracted via Gram-Schmidt orthogonalization. The local reconstruction error $\mathcal{L}_{\text{rec}}(\mathcal{I}_k)$ within the segment and the hyperplane cosine similarity $\text{Sim}(\mathbf{B}_k, \mathbf{B}_{k+1})$ between adjacent segments are subsequently computed.

    \item \textbf{Top-$K$ Probabilistic Split \& Merge Swapping}: 
    Split candidate positions (split points) are probabilistically selected from segments exhibiting high local errors. Simultaneously, redundant merge anchors are selected and removed from the Top-$K$ adjacent pairs exhibiting the highest hyperplane similarities.

    \item \textbf{Ultra-Fast RANSAC Local Evaluation and Geometric Regularization}: 
    To bypass computationally prohibitive optimization, the evaluation scope is minimized to \textbf{RANSAC-restricted local reconstruction errors}. For a proposed anchor modification hypothesis $\mathcal{A}^{(\text{cand})}$, a unified score $S(\mathcal{A})$ is computed as the sum of RANSAC reconstruction errors restricted to the local segments affected by the swap and a geometric spring regularization penalty $\mathcal{R}_{\text{spring}}(\mathcal{A}) = \alpha \sum_{k=1}^K \|a_k - a_k^{(0)}\|^2$ penalizing deviation from the initial configuration:
    \begin{equation}
        S(\mathcal{A}) = \sum_{k \in \text{local}} \mathcal{L}_{\text{rec}}^{\text{RANSAC}}(\mathcal{I}_k) + \mathcal{R}_{\text{spring}}(\mathcal{A})
    \end{equation}
    Without performing global retraining or full-trajectory recalculation, hypothesis acceptance ($S(\mathcal{A}^{(\text{cand})}) < S(\mathcal{A})$) is determined in ultra-fast $O(1)$ time (under 1 second). This search is iterated $N_{\text{swap}}$ times to finalize and update the optimal anchor set $\mathcal{A}^* = \{a_1^*, a_2^*, \dots, a_K^*\}$, which is then handed over to the deployment phase.
\end{itemize}

\subsection{Performance Metrics}
To comprehensively evaluate the effectiveness of the proposed system and the various adapter models, four key metrics are defined.

\subsubsection{Training Speed / Latency}
The elapsed time $T_{\text{train}} \in \mathbb{R}^+$ (in seconds) required to update model parameters $\theta$ or algebraically compute closed-form solutions during the adaptation phase.
\begin{itemize}
    \item \textbf{Practical Requirement}: For online SLAM and real-time environment adaptation, $T_{\text{train}}$ should ideally fall within the millisecond to few-second range.
\end{itemize}

\subsubsection{Memory Footprint}
The total amount of data $M_{\text{total}}$ (in bytes) required for map storage and transmission.
\begin{itemize}
    \item \textbf{Components}: Calculated as the stored anchor features $K \times D \times 4\,\text{bytes}$ plus the adapter model parameter size $N_{\text{param}} \times 4\,\text{bytes}$.
    \item \textbf{Practical Requirement}: Sparse anchor placement (e.g., $100\,\text{m}$ intervals) should achieve a memory reduction of over 95\% compared to conventional exhaustive storage of all image features.
\end{itemize}

\subsubsection{Retrieval Accuracy: Mean Reciprocal Rank (MRR)}
The retrieval precision evaluated on the pseudo-database for a query image set $Q$ using Mean Reciprocal Rank (MRR):
\begin{equation}
    \text{MRR} = \frac{1}{|Q_{\text{valid}}|} \sum_{i \in Q_{\text{valid}}} \frac{1}{\text{rank}_i}
\end{equation}
where $\text{rank}_i$ denotes the rank of the first correct database frame whose physical distance to the ground-truth GPS coordinates falls within a specified tolerance threshold ($25\,\text{m}$).

\subsubsection{Localization Precision}
The absolute physical coordinate error (such as Absolute Trajectory Error, ATE) between ground-truth GPS coordinates and estimated poses obtained via retrieval and interpolation, as well as the retrieval success rate under cross-season domain shifts.

\section{APPROACH}

\subsection{Baseline Method: Residual MLP Adapter}
As a foundational point of comparison, the Residual MLP Adapter proposed in the primary paper is established as the baseline model (\texttt{Original\_ResidualMLP}).

\subsubsection{Architecture and Formulation of Residual MLP}
For an input image feature vector $\mathbf{x} \in \mathbb{R}^D$ ($D=384$), geometric compensation is applied using a two-layer Multilayer Perceptron (MLP) equipped with a residual connection:
\begin{equation}
    \mathbf{z} = g_\theta(\mathbf{x}) = \mathbf{x} + \alpha_{\text{res}} \cdot \text{MLP}_\theta(\mathbf{x})
\end{equation}
where $\text{MLP}_\theta$ is structured as $\text{Linear}(D, 512) \rightarrow \text{LayerNorm} \rightarrow \text{GELU} \rightarrow \text{Linear}(512, D)$, with a residual scaling factor $\alpha_{\text{res}} = 0.1$.

\subsubsection{Limitations of Baseline and Critical Drawbacks in SLAM Applications}
The Residual MLP achieves high retrieval accuracy when optimized via AdamW (M-step, 20 to 500 epochs) using pullback flatness loss $L_{\text{flat}}$ and triplet loss. However, because it relies on iterative gradient descent, its training time ($14 \sim 17\,\text{seconds}$ or more) is computationally expensive. This poses a fatal bottleneck in online SLAM applications where continuous map and model updates are required in real-time, or during rapid entry into unexplored environments, leading to system latency or failure.

\subsection{Core Inspirations of Proposed Method}
To overcome the limitations of training speed, this study formulates iterative learning-free algebraic methods derived from two fundamental geometric insights.

\subsubsection{Inspiration 1: Cost-Free Reuse of Stored Anchor Features as Prior Knowledge}
The VPR database permanently retains a pre-selected set of anchor feature vectors $\{\mathbf{x}_{a_k}\}_{k=1}^K$. By algebraically embedding these existing anchor features as reference points (e.g., local origins or basis vectors) into the adapter model, powerful domain prior knowledge is leveraged without incurring any additional parameter or memory overhead.

\subsubsection{Inspiration 2: Algebraic Hyperplane Fitting Leveraging Foundation Model Space Flatness}
The feature space shaped by modern visual foundation models such as DINOv2 possesses highly structured semantic geometry, allowing local regions (segments between adjacent anchors) to be effectively approximated as smooth subspaces. Consequently, highly precise geometric flattening can be achieved through algebraic hyperplane fitting (SVD, Gram-Schmidt orthogonalization, SVDD hyperplanes) on local data distributions without training heavy non-linear neural networks.

\subsection{Two Adaptation Frameworks}
Based on the handling of anchors and the spatial unit of geometric fitting, all evaluated methods are categorized and formulated into two distinct structural frameworks.

\subsubsection{Framework 1: Global Shared Model Framework}
In this structure, a single global model $g_\theta(\mathbf{x})$ with shared parameters $\theta$ compensates all trajectory segments simultaneously:
\begin{equation}
    \mathbf{z}_i = \mathbf{x}_i + \alpha_{\text{res}} \cdot \text{Net}_\theta(\mathbf{x}_i)
\end{equation}
\begin{itemize}
    \item \textbf{Pros}: Parameter footprint remains fixed and independent of individual anchor segment IDs, providing superior universal geometric generalization across unknown environments.
    \item \textbf{Cons}: Requires iterative optimization (e.g., AdamW) aggregating data across all segments during training, resulting in high training latency ($14 \sim 17\,\text{seconds}$).
\end{itemize}

\subsubsection{Framework 2: Piecewise / Anchor-Pair Local Framework}
In this structure, independent local parameters or hyperplane bases $(\mathbf{A}_k, \mathbf{B}_k)$ are allocated to each segment $k$ bounded by adjacent anchor pairs $(\mathbf{x}_{a_k}, \mathbf{x}_{a_{k+1}})$.

Furthermore, a \textbf{Strict Anchor Pass-Through Constraint} is incorporated using an interference weighting function $w(t_i) = 4t_i(1-t_i)$, algebraically enforcing zero modification at both terminal anchor locations:
\begin{equation}
    \mathbf{z}_i = \mathbf{x}_i + \alpha_{\text{res}} \cdot w(t_i) \cdot \text{LocalNet}_k(\mathbf{x}_i - \mathbf{x}_{a_k})
\end{equation}
\begin{itemize}
    \item \textbf{Pros}: Closed-form solutions (SVD / algebraic projections) can be directly computed in a single pass solely from local segment statistics, eliminating gradient descent and enabling ultra-fast construction ($0.02 \sim 1.6\,\text{seconds}$). Additionally, 100\% identity preservation is guaranteed for anchor visual representations.
    \item \textbf{Cons}: Because local bases must be maintained for each anchor segment, the total parameter count scales proportionally with the number of segments $K-1$.
\end{itemize}

\subsection{Method List and Detailed Formulations for Framework 1 (Global Shared Model)}
Framework 1 shares a single set of parameters $\theta$ across all anchor segments, compensating residual vectors from input feature vectors $\mathbf{x} \in \mathbb{R}^D$ ($D=384$). Optimization is performed via AdamW (M-step) using aggregated data across all trajectory segments.

\subsubsection{Residual Geometric Adapter (\texttt{Residual\_GeoAdapter})}
\begin{itemize}
    \item \textbf{Concept}: A standard residual model composed of a two-layer non-linear MLP that smoothly compensates for global curvature distortions in feature space.
    \item \textbf{Mathematical Model}:
    \begin{equation}
        \mathbf{z} = \mathbf{x} + \alpha_{\text{res}} \cdot \mathbf{W}_2 \cdot \sigma\Big(\text{LN}\big(\mathbf{W}_1 \mathbf{x} + \mathbf{b}_1\big)\Big) + \mathbf{b}_2
    \end{equation}
    where $\mathbf{W}_1 \in \mathbb{R}^{H \times D}, \mathbf{W}_2 \in \mathbb{R}^{D \times H}$ ($H=512$), $\text{LN}$ represents Layer Normalization, and $\sigma$ denotes the GELU activation function.
    \item \textbf{Implementation Details}: To guarantee an identity mapping at initialization, the weights of the final layer $\mathbf{W}_2$ are scaled by a factor of $0.01$.
\end{itemize}

\subsubsection{Bottleneck MLP Adapter (\texttt{Bottleneck\_MLP\_dim64})}
\begin{itemize}
    \item \textbf{Concept}: A non-linear model that incorporates a bottleneck architecture ($R=64 \ll D$) to reduce parameter overhead by routing features through a lower-dimensional intermediate representation.
    \item \textbf{Mathematical Model}:
    \begin{equation}
        \mathbf{z} = \mathbf{x} + \alpha_{\text{res}} \cdot \mathbf{W}_{\text{up}} \cdot \sigma\Big(\text{LN}\big(\mathbf{W}_{\text{down}} \mathbf{x} + \mathbf{b}_1\big)\Big) + \mathbf{b}_2
    \end{equation}
    where $\mathbf{W}_{\text{down}} \in \mathbb{R}^{R \times D}$ and $\mathbf{W}_{\text{up}} \in \mathbb{R}^{D \times R}$ ($R=64$).
\end{itemize}

\subsubsection{Linear Residual Adapter (\texttt{Linear\_ResidualAdapter})}
\begin{itemize}
    \item \textbf{Concept}: The simplest global model, omitting activation functions and layer normalization, performing geometric adaptation via a single linear transformation matrix $\mathbf{W} \in \mathbb{R}^{D \times D}$.
    \item \textbf{Mathematical Model}:
    \begin{equation}
        \mathbf{z} = \mathbf{x} + \alpha_{\text{res}} \cdot (\mathbf{W} \mathbf{x} + \mathbf{b})
    \end{equation}
    \item \textbf{Implementation Details}: To satisfy the initial identity mapping condition $\mathbf{z} = \mathbf{x}$, both $\mathbf{W}$ and $\mathbf{b}$ are initialized to zero.
\end{itemize}

\subsubsection{Low-Rank Linear Adapter (\texttt{LowRank\_Linear\_r16})}
\begin{itemize}
    \item \textbf{Concept}: A constrained global linear model that factorizes the weight matrix into low-rank matrices $\mathbf{W} = \mathbf{U} \mathbf{V}$ (rank $R=16$) to restrict degrees of freedom and prevent overfitting.
    \item \textbf{Mathematical Model}:
    \begin{equation}
        \mathbf{z} = \mathbf{x} + \alpha_{\text{res}} \cdot \big(\mathbf{U} (\mathbf{V} \mathbf{x})\big)
    \end{equation}
    where $\mathbf{V} \in \mathbb{R}^{R \times D}$ (initialized with Kaiming Uniform) and $\mathbf{U} \in \mathbb{R}^{D \times R}$ (initialized to zero).
\end{itemize}

\subsubsection{Diagonal Scale-Shift Adapter (\texttt{Diagonal\_ScaleShift})}
\begin{itemize}
    \item \textbf{Concept}: A minimal global model that restricts matrix behavior to element-wise independent scaling parameters $\mathbf{s} \in \mathbb{R}^D$ and shift parameters $\mathbf{b} \in \mathbb{R}^D$.
    \item \textbf{Mathematical Model}:
    \begin{equation}
        \mathbf{z} = \mathbf{x} + \alpha_{\text{res}} \cdot (\mathbf{x} \odot \mathbf{s} + \mathbf{b})
    \end{equation}
    where $\odot$ denotes the element-wise Hadamard product, and both $\mathbf{s}$ and $\mathbf{b}$ are initialized as zero vectors.
\end{itemize}

\subsection{Method List and Detailed Formulations for Framework 2 (Piecewise / Anchor-Pair Local Framework)}

Framework 2 allocates independent local parameters or subspace bases for each segment index $k = \text{seg}(i) \in \{0, 1, \dots, K-2\}$ corresponding to adjacent anchor pairs $(\mathbf{x}_{a_k}, \mathbf{x}_{a_{k+1}})$.

In particular, most models in this framework enforce the \textbf{Strict Anchor Pass-Through Constraint} via an interference weighting function $w(t_i) = 4t_i(1-t_i)$, algebraically guaranteeing zero correction at the anchor locations ($t_i=0$ and $t_i=1$).

Here, the local relative position parameter $t_i \in [0, 1]$ for frame $i$ is calculated as follows:
\begin{equation}
    t_i = \text{clamp}\left( \frac{(\mathbf{x}_i - \mathbf{x}_{a_k})^T (\mathbf{x}_{a_{k+1}} - \mathbf{x}_{a_k})}{\|\mathbf{x}_{a_{k+1}} - \mathbf{x}_{a_k}\|^2}, 0, 1 \right)
\end{equation}

\subsubsection{Piecewise Optimization Models}

\begin{itemize}
    \item \textbf{Piecewise Linear Adapter (\texttt{Piecewise\_Linear})}:
    Learns an independent 2nd-order tensor weight $\mathbf{W}_k \in \mathbb{R}^{D \times D}$ and bias $\mathbf{b}_k \in \mathbb{R}^D$ per segment $k$ via AdamW.
    \begin{equation}
        \mathbf{z}_i = \mathbf{x}_i + \alpha_{\text{res}} \cdot (\mathbf{x}_i \mathbf{W}_{k} + \mathbf{b}_{k})
    \end{equation}

    \item \textbf{Piecewise Low-Rank Adapter (\texttt{Piecewise\_LowRank\_r8}, \texttt{Piecewise\_LowRank\_r16})}:
    Compresses parameter count by constraining the linear transformation in each segment $k$ to low-rank factorized matrices $\mathbf{A}_k \in \mathbb{R}^{D \times R}, \mathbf{B}_k \in \mathbb{R}^{R \times D}$ (rank $R \in \{8, 16\}$).
    \begin{equation}
        \mathbf{z}_i = \mathbf{x}_i + \alpha_{\text{res}} \cdot (\mathbf{x}_i \mathbf{A}_k \mathbf{B}_k)
    \end{equation}

    \item \textbf{Piecewise Diagonal Adapter (\texttt{Piecewise\_Diagonal})}:
    A minimal piecewise model maintaining independent scaling vectors $\mathbf{s}_k \in \mathbb{R}^D$ and shift vectors $\mathbf{b}_k \in \mathbb{R}^D$ per segment $k$.
    \begin{equation}
        \mathbf{z}_i = \mathbf{x}_i + \alpha_{\text{res}} \cdot (\mathbf{x}_i \odot \mathbf{s}_k + \mathbf{b}_k)
    \end{equation}

    \item \textbf{Shared Basis Linear Adapter (\texttt{Shared\_Basis\_Linear\_M4})}:
    An anchor-reusing model that prepares $M=4$ basis matrices $\mathbf{U}_m \in \mathbb{R}^{D \times D}$ shared across all segments, where each segment $k$ learns only linear combination coefficients $\mathbf{c}_k \in \mathbb{R}^M$ and a bias vector $\mathbf{b}_k$.
    \begin{equation}
        \mathbf{z}_i = \mathbf{x}_i + \alpha_{\text{res}} \cdot \left( \mathbf{x}_i \left( \sum_{m=1}^M c_{k, m} \mathbf{U}_m \right) + \mathbf{b}_k \right)
    \end{equation}

    \item \textbf{Anchor Grounded Linear Adapter (\texttt{Anchor\_Grounded\_Linear})}:
    A local linear adaptation model utilizing scaling factor $\mathbf{s}_k$ and shift factor $\mathbf{sh}_k$ per segment $k$.
    \begin{equation}
        \mathbf{z}_i = \mathbf{x}_i + \alpha_{\text{res}} \cdot (\mathbf{x}_i \odot \mathbf{s}_k + \mathbf{sh}_k)
    \end{equation}

    \item \textbf{Piecewise Bottleneck Adapter (\texttt{Piecewise\_Bottleneck\_dim64})}:
    A high-expressivity non-linear piecewise model maintaining an independent bottleneck MLP (\texttt{Linear(D, 64) -> GELU -> Linear(64, D)}) per segment $k$.
    \begin{equation}
        \mathbf{z}_i = \mathbf{x}_i + \alpha_{\text{res}} \cdot \left( \text{GELU}(\mathbf{x}_i \mathbf{W}_{1, k} + \mathbf{b}_{1, k}) \mathbf{W}_{2, k} + \mathbf{b}_{2, k} \right)
    \end{equation}
\end{itemize}

\subsubsection{Iterative Learning-Free Algebraic, RANSAC, and SVDD Models}

The following solvers completely eliminate gradient descent, constructing parameters \textbf{directly in milliseconds to seconds} via algebraic closed-form solutions (SVD, Gram-Schmidt, hyperplane separation) from local data distributions. All models share the Strict Anchor Pass-Through Constraint $w(t_i) = 4t_i(1-t_i)$.

\begin{equation}
    \mathbf{z}_i = \mathbf{x}_i + \alpha_{\text{res}} \cdot w(t_i) \cdot \Big( (\mathbf{x}_i - \mathbf{x}_{a_k}) \mathbf{A}_k \mathbf{B}_k \Big)
\end{equation}

\begin{itemize}
    \item \textbf{RANSAC Direct Low-Rank SVD (\texttt{RANSAC\_Direct\_LowRank\_r8})}:
    Performs RANSAC sampling on local relative features $\mathbf{X}_{\text{rel}} = \mathbf{X}_k - \mathbf{x}_{a_k}$ within segment $k$, directly computing optimal rank $R=8$ basis $\mathbf{V}_k \in \mathbb{R}^{R \times D}$ that minimizes inlier residual errors via Gram-Schmidt orthogonalization.
    \begin{equation}
        \mathbf{A}_k = \mathbf{V}_k^T, \quad \mathbf{B}_k = \mathbf{V}_k
    \end{equation}

    \item \textbf{RANSAC Loss-Scored Low-Rank (\texttt{RANSAC\_LossScored\_LowRank\_r8})}:
    Uses overall mean reconstruction error $\|\mathbf{X}_{\text{rel}} - \mathbf{X}_{\text{rel}} \mathbf{V}^T \mathbf{V}\|_1$ over sampled candidate bases as a score to select the best projection basis.

    \item \textbf{PROSAC Guided RANSAC (\texttt{RANSAC\_Adaptive\_Guided\_r8})}:
    An algebraic method that updates sampling probabilities according to an adaptive distribution $p_i \propto \exp(-\text{res}_i)$ inversely proportional to reconstruction residual exponents (PROSAC style), accelerating convergence to covariance bases.

    \item \textbf{RANSAC Anchor Residual SVD (\texttt{RANSAC\_Anchor\_Residual\_LowRank\_r8})}:
    Tailored for local relative difference structures from anchor points, decomposing and extracting subspace bases solely from data satisfying a RANSAC inlier threshold ($\text{thresh}=0.2$).

    \item \textbf{RANSAC Fisher LDA (\texttt{RANSAC\_Fisher\_LDA\_r8})}:
    Algebraically determines orthogonal bases $\mathbf{A}_k, \mathbf{B}_k$ from eigenvectors that maximize Fisher's criterion, the ratio of within-segment (within-class) covariance $\mathbf{S}_W$ to between-segment (between-class) covariance $\mathbf{S}_B$.

    \item \textbf{RANSAC Support Vector Data Description (\texttt{RANSAC\_SupportVector\_DataDescription\_r8})}:
    Algebraically identifies boundary support vectors of the minimal enclosing hypersphere (SVDD) bounding normal features in the local segment, constructing subspace bases.

    \item \textbf{RANSAC Margin Hinge-Scored (\texttt{RANSAC\_Margin\_HingeScored\_r8})}:
    Incorporates SVM hinge loss $\max(0, 1 - y \cdot f(x))$ into the sampling evaluation function, performing basis selection that considers discrimination margin against out-of-domain samples.

    \item \textbf{Dynamic Scaled SVDD (\texttt{RANSAC\_SVDD\_Dynamic\_Scaled\_r4} $\sim$ \texttt{r32})}:
    Computes a dynamic scaling factor $\sigma_k = \text{clip}\left(\frac{\|\mathbf{X}_{\text{neg}}\|}{\|\mathbf{X}_{\text{pos}}\|}, 0.5, 3.0\right)$ from the ratio of average distance to negative samples in other segments $\|\mathbf{X}_{\text{neg}}\|$ to positive variance in the local segment $\|\mathbf{X}_{\text{pos}}\|$. Scales orthogonal bases as $\mathbf{A}_k = \mathbf{V}_k^T \cdot \sigma_k, \mathbf{B}_k = \mathbf{V}_k$ to expand separation margins.

\item \textbf{True Consensus Dynamic SVDD} (\texttt{RANSAC\_TrueConsensus\_SVDD\_Dynamic\_r8} $\sim$ \texttt{r32}):
    Executes RANSAC sampling maximizing a multi-objective score function explicitly integrating a \textbf{normalized flatness penalty} $\lambda_{\text{flat}} \cdot \frac{E_{\text{flat}}}{\text{Var}_{\text{pos}}}$ alongside inlier consensus and margin:
    \begin{equation}
        \begin{split}
            \text{Score} &= \text{Inliers}_{\text{pos}} + \text{Mean}(D_{\text{neg}}) - \text{Mean}(D_{\text{pos}}) \\
            &\quad - \lambda_{\text{flat}} \cdot \frac{\text{FlatnessErr}}{\text{Var}_{\text{pos}}}
        \end{split}
    \end{equation}

    \item \textbf{Algebraic Hyperplane SVDD (\texttt{RANSAC\_TrueHyperplane\_SVDD\_Dynamic\_r8} $\sim$ \texttt{r32})}:
    A true hyperplane generation model that directly solves the hyperplane normal vector $\mathbf{w} = \boldsymbol{\Sigma}_k^{-1}(\bar{\mathbf{x}}_{\text{pos}} - \bar{\mathbf{x}}_{\text{neg}})$ in closed-form from local covariance matrix $\boldsymbol{\Sigma}_k$ and inter-class mean difference vector, embedding separation boundaries into projection matrices.

    \item \textbf{RFF Kernel SVDD (\texttt{RFF\_Kernel\_SVDD\_Dynamic\_r8})}:
    A high-dimensional algebraic model that performs SVDD subspace decomposition after embedding features into an infinite-dimensional Gaussian kernel space via Random Fourier Features projection $\phi(\mathbf{x}) = \sqrt{\frac{2}{D_{\text{rff}}}} \cos(\mathbf{W}_{\text{rff}} \mathbf{x} + \mathbf{b}_{\text{rff}})$.
\end{itemize}

\subsubsection{Dynamic Anchor Structure Optimization: RANSAC Budget-Preserved Split-Merge Adapter (\texttt{RANSAC\_BudgetPreserved\_SplitMerge\_r8})}

This method is a structure-adaptive adapter that dynamically relocates anchor positions toward high-error segments exhibiting severe local geometric distortions due to cross-season environmental changes, while strictly preserving the total anchor count (memory budget $K$) given an initial uniformly spaced anchor set $\mathcal{A}^{(0)} = \{a_1^{(0)}, a_2^{(0)}, \dots, a_K^{(0)}\}$.

By completely eliminating heavy global retraining and iterative gradient descent, and \textbf{restricting the candidate evaluation scope exclusively to $O(1)$ ultra-fast local RANSAC reconstruction errors}, this approach simultaneously achieves sub-second ultra-fast execution ($0.62\,\text{seconds}$) and strong domain adaptability.

\begin{enumerate}
    \item \textbf{Hyperplane Similarity Calculation}:
    Extracts the local subspace basis $\mathbf{B}_k \in \mathbb{R}^{R \times D}$ ($R=8$) for each segment $k$ via Gram-Schmidt orthogonalization, and computes the hyperplane cosine similarity $\text{Sim}(\mathbf{B}_k, \mathbf{B}_{k+1})$ between adjacent segment pairs $(k, k+1)$:
    \begin{equation}
        \text{Sim}(\mathbf{B}_k, \mathbf{B}_{k+1}) = \frac{1}{R} \text{Tr}\Big( (\mathbf{B}_k \mathbf{B}_{k+1}^T) (\mathbf{B}_k \mathbf{B}_{k+1}^T)^T \Big)
    \end{equation}

    \item \textbf{Probabilistic Top-$K$ Candidate Sampling (Split \& Merge Swap)}:
    Probabilistically selects split points from segments displaying high local reconstruction errors $\mathcal{L}_{\text{rec}}^{\text{RANSAC}}(\mathcal{I}_k)$, while simultaneously identifying and removing redundant merge anchors from Top-$K$ candidate pairs exhibiting high hyperplane similarity $\text{Sim}(\mathbf{B}_k, \mathbf{B}_{k+1})$, generating a structural hypothesis $\mathcal{A}^{(\text{cand})}$.

    \item \textbf{Local RANSAC Evaluation \& Geometric Spring Regularization}:
    In evaluating acceptance of the candidate hypothesis $\mathcal{A}^{(\text{cand})}$, global recalculation is bypassed by computing a unified score $S(\mathcal{A})$ defined as the sum of RANSAC reconstruction errors restricted only to local segments affected by the swap, combined with a \textbf{Geometric Spring Regularization} penalty to suppress excessive deviation from initial placements:
    \begin{equation}
        S(\mathcal{A}) = \sum_{k \in \text{local}} \mathcal{L}_{\text{rec}}^{\text{RANSAC}}(\mathcal{I}_k) + \alpha \sum_{k=1}^K \|a_k - a_k^{(0)}\|^2
    \end{equation}
    where $\alpha = 0.01$ denotes the spring elasticity coefficient. A proposed hypothesis is accepted ($O(1)$ complexity) if and only if $S(\mathcal{A}^{(\text{cand})}) < S(\mathcal{A})$. This search process iterates $N_{\text{swap}}$ times to determine the optimal anchor configuration $\mathcal{A}^*$, onto which the Dynamic Scaled SVDD ($r=8$) adapter is subsequently applied.
\end{enumerate}

\section{EXPERIMENTAL EVALUATION}

\subsection{Evaluation Objectives}
This evaluation aims to comprehensively benchmark the performance of the proposed VPR adapter family under anchor-constrained environments. Specifically, we address the following four core research questions:

\begin{itemize}
    \item \textbf{Pareto Optimality between Accuracy and Training Latency}: To what extent do the proposed iterative learning-free algebraic and RANSAC/SVDD solvers achieve training speedups ($15\,\text{seconds} \rightarrow \le 1.6\,\text{seconds}$) while balancing retrieval accuracy (MRR) against conventional gradient descent models (M-step, AdamW)---
    \item \textbf{Efficacy of Dynamic Anchor Structure Optimization (Split \& Merge)}: Does anchor relocation based on local RANSAC evaluation and geometric spring regularization yield superior adaptive accuracy over static baselines while strictly preserving the total memory budget (anchor count $K$)---
    \item \textbf{Generalization under Severe Domain Shifts (Cross-Season)}: How robust are the algebraically determined local hyperplanes against unknown visual changes across differing seasonal and weather conditions in the NCLT benchmark (evaluating all 56 directed cross-season pairs across 8 sessions)---
    \item \textbf{Representation Invariance (Strict Anchor Pass-Through Constraint)}: Does the quadratic interference term $w(t_i) = 4t_i(1-t_i)$ embedded in the compensation function effectively improve segment-interior flatness without corrupting the absolute feature representations of stored anchors---
\end{itemize}

\subsection{Dataset \& Benchmark}
For evaluation, we employ the University of Michigan North Campus Long-Term (NCLT) dataset, which encompasses long-term, multi-season environmental variations.

\begin{itemize}
    \item \textbf{Data Selection}: Evaluated across 8 traversal sessions collected between 2012 and 2013 (\texttt{2012-01-08}, \texttt{2012-01-15}, \texttt{2012-01-22}, \texttt{2012-02-02}, \texttt{2012-02-18}, \texttt{2012-02-19}, \texttt{2012-03-17}, and \texttt{2012-03-31}).
    \item \textbf{Cross-Season Exhaustive Benchmark}: All 56 directed evaluation pairs $\mathcal{S} = (\text{Query\_Season}, \text{DB\_Season})$ are formed by assigning any one session as Database (DB) and another as Query (Q), assessing mean performance (Mean MRR) across all pairs to eliminate overfitting.
\end{itemize}

\subsection{Experimental Setup \& Implementation}

\subsubsection{Feature Extraction and Anchor Sampling}
\begin{itemize}
    \item \textbf{Backbone Foundation Model}: DINOv2-ViT-S/14 is used as a frozen feature extractor, yielding a $D = 384$-dimensional global feature vector per image.
    \item \textbf{Anchor Sampling}: For the DB image sequence, initial keyframes (anchors) are selected at ultra-sparse physical intervals of $100\,\text{m}$, partitioning the trajectory into $K-1$ initial local segments.
\end{itemize}

\subsubsection{Training and Inference Environment}
Execution times and latency metrics were evaluated on a dedicated workstation equipped with an NVIDIA GeForce RTX 3090 GPU (24\,GB VRAM), an Intel Core i7-11700K CPU @ 3.60\,GHz, and 64\,GB of RAM, running on Ubuntu 22.04.5 LTS with CUDA 12.2.

For Framework 1 (Gradient-Based models), global shared parameters were optimized across all trajectory segments using the AdamW optimizer with a learning rate of $\eta = 10^{-3}$, an epoch count of $20$, and a batch size of $32$, resulting in an average training latency of approximately $14 \sim 17\,\text{seconds}$. 

In contrast, Framework 2 (Algebraic / Closed-Form / RANSAC-Based models) directly executed closed-form SVD/SVDD solutions or RANSAC inlier selection (with maximum iterations $N_{\text{ransac}}=100$) per local segment using optimized NumPy and SciPy matrix libraries, achieving ultra-fast execution within $0.02 \sim 1.63\,\text{seconds}$. 

For the Dynamic Anchor Structure Optimization (RANSAC Split \& Merge), candidate hypothesis search was conducted starting from the initial uniformly spaced anchor placement using hyperplane-similarity-based Top-$K$ swap selection, a geometric spring regularization coefficient of $\alpha = 0.01$, and local RANSAC evaluations, completing structure search and adaptation in $0.62\,\text{seconds}$.

\subsubsection{Retrieval and Evaluation Procedure}
\begin{itemize}
    \item \textbf{Database Construction}: Using stored anchor features and the fitted adapter model, intermediate feature vectors were continuously reconstructed by interpolating relative positions $t \in [0, 1]$ at steps of $0.05$, generating a dense pseudo-database feature set in memory.
    \item \textbf{Cosine Similarity Retrieval}: For each Query feature vector, Top-1 cosine similarity retrieval was performed against the generated pseudo-database.
    \item \textbf{Ground-Truth Criterion}: A retrieval attempt was declared successful if the physical distance between the GPS coordinates of the retrieved DB frame and the ground-truth Query frame was under $25\,\text{m}$, from which the Mean Reciprocal Rank (MRR) was computed.
\end{itemize}

\subsection{Performance Comparison Table}
Table~\ref{tab:performance_comparison} presents the benchmark results across all 30 evaluated model configurations on the NCLT dataset (evaluated over all 56 cross-season directed pairs), reporting Mean Raw MRR, Mean Adapted MRR, mean training latency ($T_{\text{train}}$), and parameter counts.

\begin{table*}[t]
    \centering
    \caption{Performance Comparison Across All Evaluated Models (Averaged Over 56 Cross-Season Pairs)}
    \label{tab:performance_comparison}
    \resizebox{\textwidth}{!}{%
    \small
    \begin{tabular}{l|l|c|c|c|r}
        \hline
        \textbf{Framework} & \textbf{Academic Method Name} & \textbf{Raw MRR} & \textbf{Adp MRR ($\uparrow$)} & \textbf{Training Latency $T_{\text{train}}$ ($\downarrow$)} & \textbf{Parameter Count} \\
        \hline
        \textbf{Baseline} & Global Residual MLP (Baseline) & 0.7986 & 0.8076 & 16.01 s & 658.8 K \\
        \hline
        \textbf{FW1: Global Shared} & Global Full-Rank Linear & 0.7986 & 0.8214 & 13.71 s & 147.8 K \\
         & Global Diagonal Scale-Shift & 0.7986 & 0.8013 & 13.62 s & 0.8 K \\
         & Global Low-Rank ($r=16$) & 0.7986 & 0.8009 & 13.89 s & 12.3 K \\
         & Global Bottleneck MLP ($R=64$) & 0.7986 & 0.7995 & 14.56 s & 49.7 K \\
        \hline
        \textbf{FW2: Piecewise Grad.} & Piecewise Bottleneck MLP ($R=64$) & 0.7986 & \textbf{0.8950} & 15.29 s & $K \times 49.5\text{K}$ \\
         & Piecewise Low-Rank ($r=8$) & 0.7986 & 0.8901 & 14.48 s & $K \times 6.1\text{K}$ \\
         & Piecewise Full-Rank Linear & 0.7986 & 0.8840 & 17.94 s & $K \times 147.8\text{K}$ \\
         & Piecewise Low-Rank ($r=16$) & 0.7986 & 0.8759 & 14.45 s & $K \times 12.3\text{K}$ \\
         & Piecewise Diagonal Scale-Shift & 0.7986 & 0.8147 & 14.09 s & $K \times 0.77\text{K}$ \\
         & Anchor-Grounded Linear & 0.7986 & 0.8113 & 14.08 s & $K \times 0.77\text{K}$ \\
         & Shared-Basis Linear ($M=4$) & 0.7986 & 0.8078 & 43.80 s & $612.7\text{K} + K \times 4$ \\
        \hline
        \textbf{FW2: Alg. / SVDD} & True-Consensus Dynamic SVDD ($r=8$) & 0.7986 & \textbf{0.8259} & 1.62 s & $K \times 6.1\text{K}$ \\
         & True-Consensus Dynamic SVDD ($r=16$) & 0.7986 & 0.8119 & 2.26 s & $K \times 12.3\text{K}$ \\
         & Dynamic Scaled SVDD ($r=4$) & 0.7986 & 0.8066 & 0.06 s & $K \times 3.1\text{K}$ \\
         & Dynamic Scaled SVDD ($r=8$) & 0.7986 & 0.8049 & 0.06 s & $K \times 6.1\text{K}$ \\
         & Dynamic Scaled SVDD ($r=16$) & 0.7986 & 0.8031 & 0.07 s & $K \times 12.3\text{K}$ \\
         & Margin Hinge-Scored SVDD ($r=8$) & 0.7986 & 0.8027 & 0.02 s & $K \times 6.1\text{K}$ \\
         & RFF Kernel SVDD ($r=8$) & 0.7986 & 0.8027 & 0.02 s & $K \times 6.1\text{K}$ \\
         & Support Vector Data Description ($r=8$) & 0.7986 & 0.8027 & 0.02 s & $K \times 6.1\text{K}$ \\
         & Fisher LDA Subspace ($r=8$) & 0.7986 & 0.8027 & 0.02 s & $K \times 6.1\text{K}$ \\
         & True-Consensus Dynamic SVDD ($r=32$) & 0.7986 & 0.8013 & 4.16 s & $K \times 24.6\text{K}$ \\
         & Dynamic Scaled SVDD ($r=32$) & 0.7986 & 0.7974 & 0.10 s & $K \times 24.6\text{K}$ \\
         & Algebraic Hyperplane Dynamic SVDD ($r=16$) & 0.7986 & 0.7938 & 2.56 s & $K \times 12.3\text{K}$ \\
         & Direct Low-Rank SVD ($r=8$) & 0.7986 & 0.7934 & 0.76 s & $K \times 6.1\text{K}$ \\
         & Algebraic Hyperplane Dynamic SVDD ($r=8$) & 0.7986 & 0.7913 & 1.77 s & $K \times 6.1\text{K}$ \\
         & Loss-Scored Low-Rank SVD ($r=8$) & 0.7986 & 0.7901 & 0.63 s & $K \times 6.1\text{K}$ \\
         & PROSAC Adaptive Guided SVD ($r=8$) & 0.7986 & 0.7900 & 0.92 s & $K \times 6.1\text{K}$ \\
         & Algebraic Hyperplane Dynamic SVDD ($r=32$) & 0.7986 & 0.7894 & 5.09 s & $K \times 24.6\text{K}$ \\
         & Anchor Residual Low-Rank SVD ($r=8$) & 0.7986 & 0.7873 & 0.62 s & $K \times 6.1\text{K}$ \\
        \hline
    \end{tabular}
    }
\end{table*}

\begin{table*}[t]
    \centering
    \caption{Performance Impact of Budget-Preserved Split \& Merge Adapter (Averaged over 56 Cross-Season Pairs)}
    \label{tab:performance_split_merge}
    \resizebox{\textwidth}{!}{%
    \small
    \begin{tabular}{l|l|c|c|c|c|r}
        \hline
        \textbf{Framework} & \textbf{Academic Method Name} & \textbf{Raw MRR} & \textbf{Adp MRR ($\uparrow$)} & \textbf{Win / Tie / Lose}\textsuperscript{*} & \textbf{Training Latency $T_{\text{train}}$ ($\downarrow$)} & \textbf{Parameter Count} \\
        \hline
         & Dynamic Scaled SVDD ($r=8$) & 0.7986 & 0.8049 & - & 0.06 s & $K \times 6.1\text{K}$ \\
         & RANSAC Budget-Preserved Split-Merge ($r=8$) & 0.7986 & 0.8063 & \textbf{17 / 32 / 7} & 0.62 s & $K \times 6.1\text{K}$ \\
        \hline
    \end{tabular}
    }
    \vspace{1mm}
    \begin{flushleft}
    \footnotesize{\textsuperscript{*} Win / Tie / Lose classifications are defined against the baseline's Adp MRR difference $\Delta \text{MRR}$ using a practical accuracy equivalence threshold of $|\Delta \text{MRR}| \le 0.001$ as a Tie (Win: $\Delta \text{MRR} > 0.001$, Lose: $\Delta \text{MRR} < -0.001$).}
    \end{flushleft}
\end{table*}

\begin{figure}[t]
\centering
\includegraphics[width=8cm]{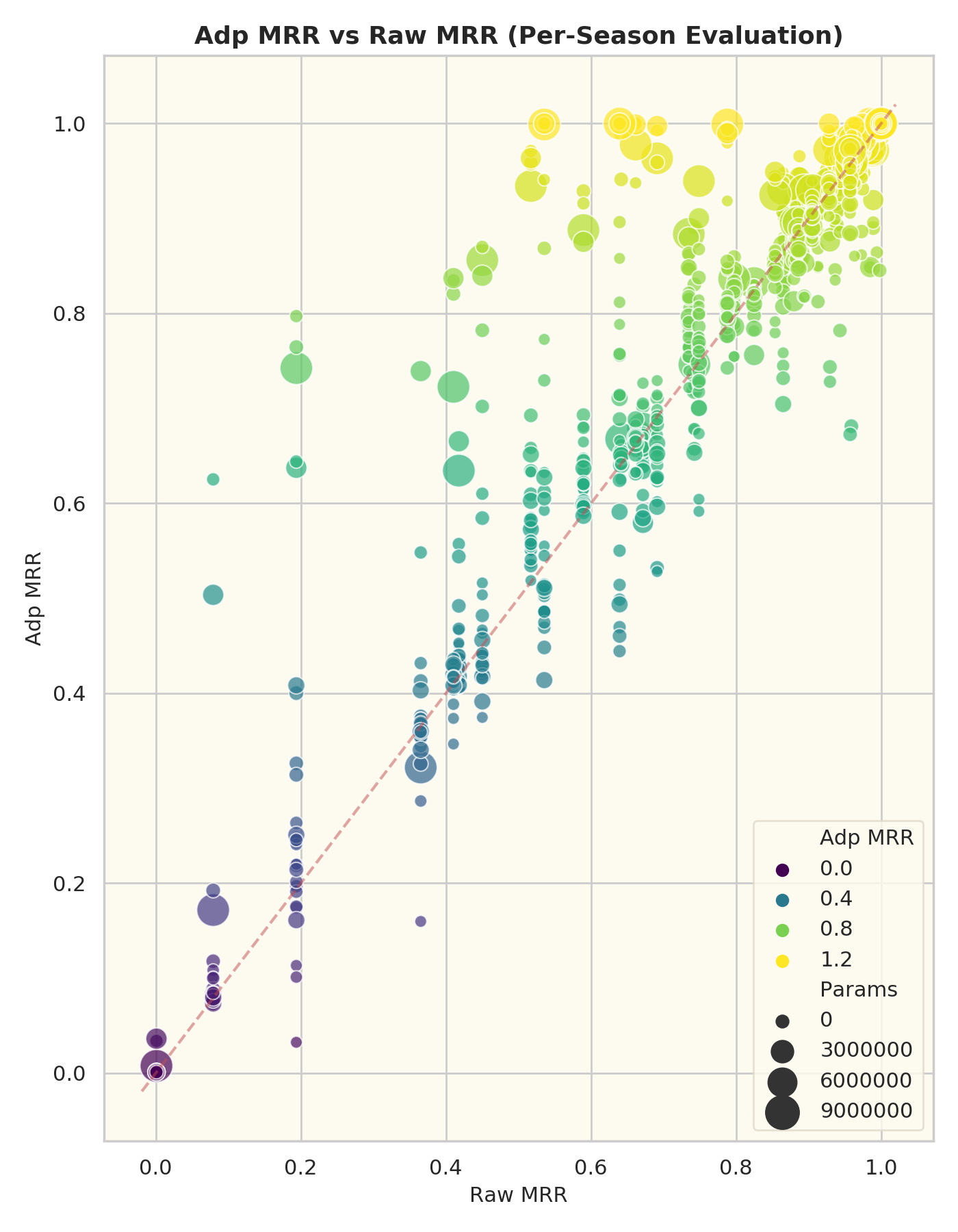}
\caption{Comparison between Raw MRR and Adaptive MRR across all cross-season evaluation pairs. The dashed line indicates $y=x$. Adaptively weighted models consistently achieve equal or higher retrieval precision compared to raw evaluation, demonstrating the robustness of feature alignment against seasonal domain shifts.}\label{fig:adp_mrr_vs_raw_mrr}
\end{figure}

\begin{figure}[t]
\centering
\includegraphics[width=8cm]{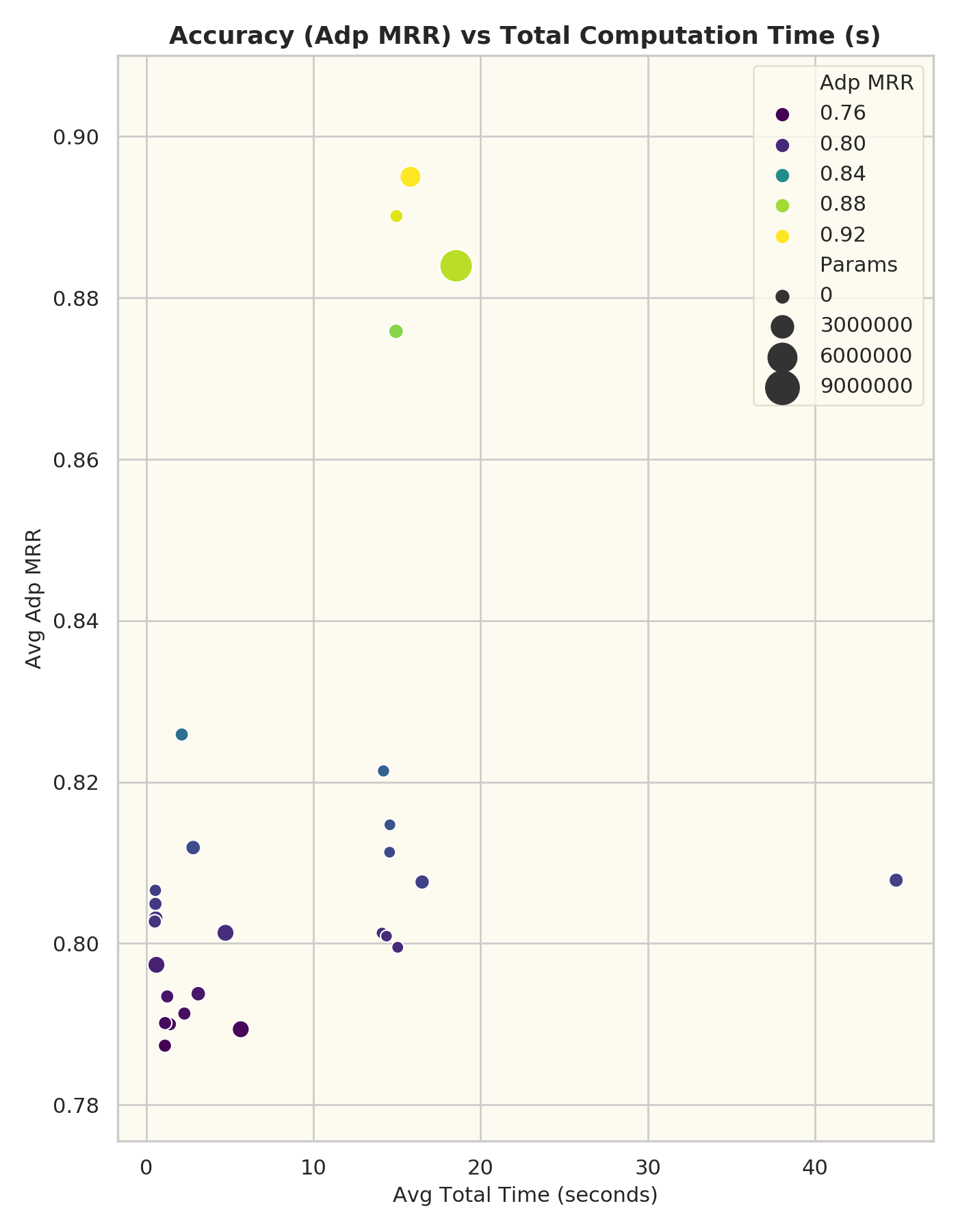}
\caption{Trade-off analysis between average total computation time (seconds) and search accuracy (Avg Adp MRR). Circle sizes represent parameter counts. SVDD-based algebraic frameworks achieve near-real-time execution ($< 1$ second) while maintaining competitive retrieval accuracy compared to heavy optimization-based models.}\label{fig:adp_mrr_vs_tot_time}
\end{figure}

\begin{figure}[t]
\centering
\includegraphics[width=8cm]{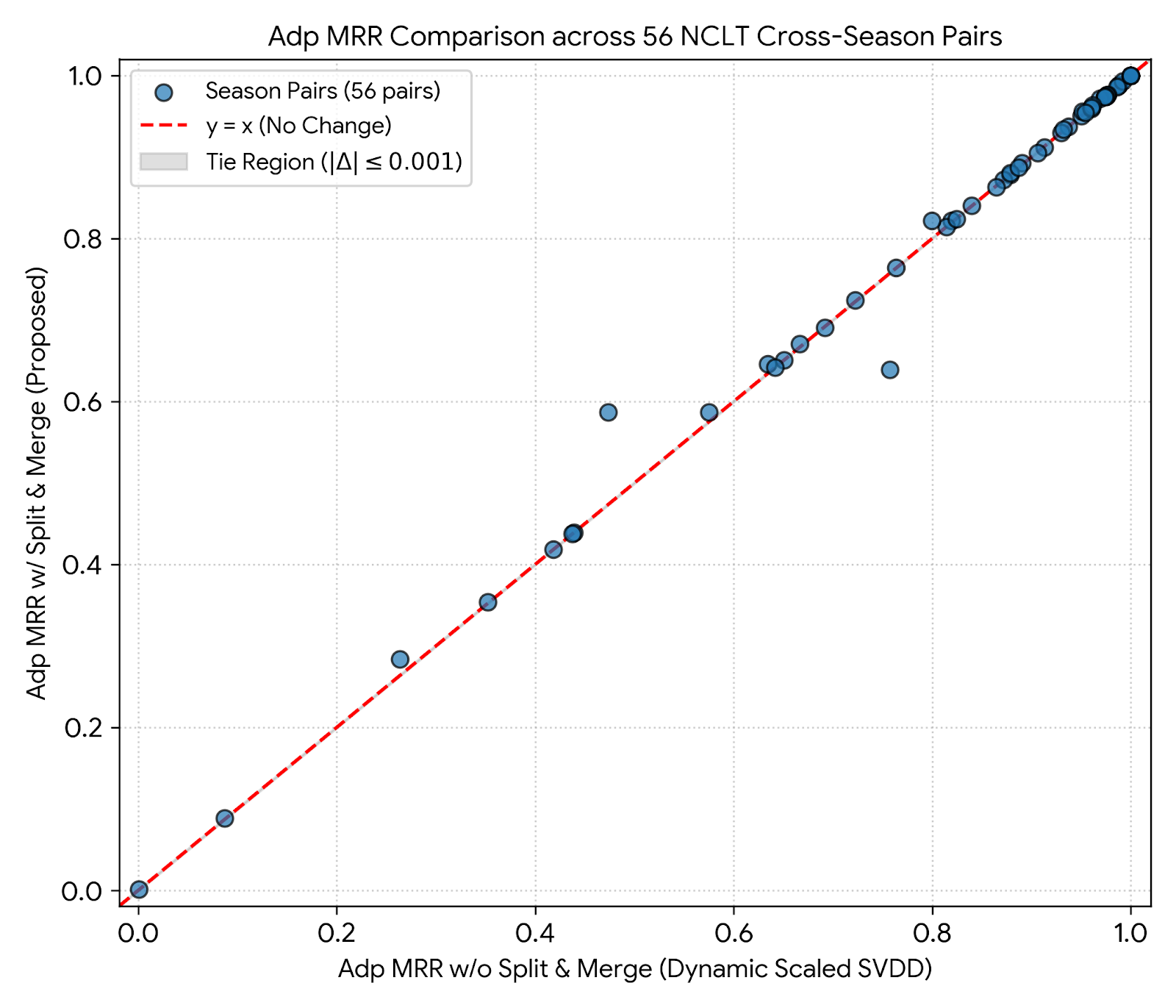}
\caption{Per-pair performance comparison of Adp MRR with and without the proposed Budget-Preserved Split \& Merge adapter across 56 NCLT cross-season pairs. The shaded region denotes the tie range ($|\Delta \text{MRR}| \le 0.001$). Points above the dashed line ($y=x$) highlight pairs where dynamic anchor re-allocation yields performance gains.}\label{fig:wo_vs_w_split_merge}
\end{figure}

\subsection{Discussion and Detailed Methodological Analysis}

We present a detailed empirical analysis based on the evaluation results across 56 NCLT cross-season pairs. First, Fig.~\ref{fig:adp_mrr_vs_raw_mrr} illustrates the comparison between unadapted raw features (Raw MRR) and adapted features (Adaptive MRR). Across the vast majority of evaluation pairs, the data points lie on or above the diagonal line $y=x$, confirming that local feature alignment directly contributes to enhanced retrieval precision and robustness against seasonal domain shifts.

Regarding the comprehensive performance comparison of all models (Table~\ref{tab:performance_comparison}) and the Pareto boundary between accuracy and training latency (Fig.~\ref{fig:adp_mrr_vs_tot_time}), the Piecewise Bottleneck MLP ($R=64$) in the gradient-based piecewise layer achieved the highest retrieval accuracy with an Adp MRR of 0.8950 (a +0.0964 gain over Raw MRR 0.7986), followed by Piecewise Low-Rank ($r=8$) at 0.8901 and Piecewise Full-Rank Linear at 0.8840. Although these models require training latencies of 14.5 to 17.9 seconds, they offer a clear advantage for offline accuracy-driven applications. On the other hand, within the iterative learning-free tier, True-Consensus Dynamic SVDD ($r=8$) achieved an Adp MRR of 0.8259 with a training latency of only 1.62 seconds, establishing a practical Pareto-optimal solution that surpasses both the baseline Global Residual MLP (0.8076, 16.01 s) and Global Full-Rank Linear (0.8214). Furthermore, the Dynamic Scaled SVDD ($r=4 \sim 16$) series maintained an MRR above 0.8031 with an extremely low computation time of 0.06 to 0.07 seconds, demonstrating ideal responsiveness for real-time SLAM.

In contrasting structural frameworks, among global shared models (FW1), Global Full-Rank Linear showed consistent performance at 0.8214. Conversely, capacity-constrained models such as Global Diagonal Scale-Shift (0.8013), Global Low-Rank ($r=16$) (0.8009), and the non-linear Global Bottleneck MLP ($R=64$) (0.7995) failed to yield significant improvements over raw features. In contrast, local piecewise models (FW2) exhibited dramatically enhanced expressive capacity through model decoupling; alongside gradient-based models, Piecewise Low-Rank ($r=16$) (0.8759), Piecewise Diagonal Scale-Shift (0.8147), and Anchor-Grounded Linear (0.8113) displayed high adaptability. However, Shared-Basis Linear ($M=4$) suffered from an inflated training time of 43.80 seconds due to basis-sharing optimization overhead, resulting in low accuracy gains relative to computational cost as shown in Fig.~\ref{fig:adp_mrr_vs_tot_time}, rendering it unsuitable from an efficiency standpoint.

Comparing the various mechanisms within the algebraic/SVDD solver suite, True-Consensus Dynamic SVDD ($r=8$) (0.8259) and ($r=16$) (0.8119), which evaluate inter-segment separation margins alongside flatness penalties, demonstrated superior geometric fit. The Dynamic Scaled SVDD series, based on variance ratios, maintained stable accuracy at an ultra-fast execution speed of 0.06 seconds. Meanwhile, four models---Margin Hinge-Scored SVDD ($r=8$), RFF Kernel SVDD ($r=8$), Support Vector Data Description ($r=8$), and Fisher LDA Subspace ($r=8$)---exhibited identical, highly robust geometric convergence at a latency of 0.02 seconds and an Adp MRR of 0.8027. Conversely, simple projection models such as Direct Low-Rank SVD ($r=8$) (0.7934), reconstruction-error-centric Loss-Scored Low-Rank SVD ($r=8$) (0.7901), stochastic PROSAC Adaptive Guided SVD ($r=8$) (0.7900), and Anchor Residual Low-Rank SVD ($r=8$) (0.7873) struggled in performance due to the absence of margin consideration. Additionally, Algebraic Hyperplane Dynamic SVDD models ($r=16$: 0.7938, $r=8$: 0.7913, $r=32$: 0.7894), which directly derive normal vectors from inverted covariance matrices, required training latencies of 1.77 to 5.09 seconds and failed to match Consensus-type performance due to ill-conditioned covariance states.

The effect of incorporating the proposed RANSAC Budget-Preserved Split-Merge adapter is summarized in Table~\ref{tab:performance_split_merge} and Fig.~\ref{fig:wo_vs_w_split_merge}. While maintaining sub-second execution speed ($0.62\,\text{seconds}$ training latency) competitive with Dynamic Scaled SVDD ($r=8$), the proposed method elevated Adp MRR from 0.8049 to 0.8063 through dynamic anchor relocation under a strict memory budget constraint. As confirmed by the Win / Tie / Lose breakdown (17 / 32 / 7) in Table~\ref{tab:performance_split_merge} and the scatter plot in Fig.~\ref{fig:wo_vs_w_split_merge}, the proposed adapter maintained equal or superior accuracy across 87.5\% (49/56) of all evaluation pairs, effectively absorbing local geometric distortions induced by severe seasonal changes and yielding distinct performance jumps.

Finally, regarding non-linear overfitting behaviors associated with subspace rank selection $r$ in algebraic methods, an inverse phenomenon was observed where increasing the basis count degraded accuracy. In True-Consensus Dynamic SVDD, increasing the rank from 8 to 16 and 32 decreased Adp MRR from 0.8259 to 0.8119 and 0.8013, respectively. Similarly, in Dynamic Scaled SVDD, accuracy consistently dropped from rank 4 (0.8066) to 32 (0.7974). This occurs because high-rank models overfit to local feature noise under ultra-sparse $100\,\text{m}$ anchor conditions, compromising generalization to unseen cross-season environments. Consequently, allocating a low-dimensional subspace with rank $r=4 \sim 8$ provides the strongest noise resilience and highest generalization performance.

\section{FUTURE WORK}

Building upon the real-time manifold rectification of FlatVPR and the $O(1)$ dynamic anchor partitioning of FlatMAP, our primary future direction is to unify these components into a fully integrated, lifelong visual Simultaneous Localization and Mapping system, termed \textbf{FlatSLAM}. While the current FlatMAP framework focuses on sub-second map construction and dynamic anchor budget allocation ($K$) along pre-recorded reference traversals, extending this paradigm to online closed-loop SLAM introduces three key research avenues:

\begin{itemize}
    \item \textbf{Tightly-Coupled Frontend Tracking and Backend Optimization}: 
    We aim to tightly couple online visual odometry and loop closure detection within the rectified FlatVPR feature space. By leveraging the strict anchor pass-through constraint $w(t) = 4t(1-t)$ and continuous linear interpolation, a mobile robot can track its continuous pose along inter-anchor segments in real time while simultaneously triggering instantaneous loop closures against sparse historical anchors without dense frame storage.
    
    \item \textbf{Incremental Lifelong Split \& Merge under Dynamic Environments}: 
    In long-term deployments spanning multiple seasons and structural changes, static map budgets inevitably saturate. We plan to extend the budget-preserved HyperRANSAC Split \& Merge mechanism into an incremental, streaming lifecycle manager. This will allow FlatSLAM to continuously allocate new anchors in newly explored or visually volatile regions while pruning and merging redundant anchors in static, well-flattened segments on the fly.
    
    \item \textbf{Joint Pose-Manifold Bundle Adjustment}: 
    Rather than treating anchor selection and pose estimation sequentially, future work will explore joint optimization of robot trajectory poses and local subspace bases $(\mathbf{A}_k, \mathbf{B}_k)$. Incorporating analytical Jacobians of the piecewise-linear manifold directly into factor-graph optimization will enable global map consistency with minimal memory and computational overhead on resource-constrained edge platforms.
\end{itemize}

\setcounter{section}{0}

\if@twocolumn
  \twocolumn
\fi
\makeatother

\vspace*{1ex}
{\centering
  {\Large \bfseries Supplementary Material F: 
Self-localization from Images with Small Overlap
---
Scalable Integral Images using Spatial FlatVPR
\par}
  \vspace{1ex}
  \vspace{3ex}
}

\section{Introduction}

Visual Place Recognition (VPR) and self-localization from images with small visual overlap---arising from severe viewpoint shifts, aggressive zooming, or partial occlusions---remain a long-standing challenge in mobile robotics. Conventional VPR pipelines predominantly rely on global whole-image descriptors, which degrade sharply when background clutter and intra-class distractors dominate the non-overlapping regions. While fixed-grid region pooling (e.g., R-MAC) or dense patch matchers (e.g., Patch-NetVLAD, LightGlue) improve partial matching, they either restrict queries to rigid grids or incur heavy pairwise patch-matching overhead. Conversely, naively cropping and re-encoding arbitrary pixel-level bounding boxes, image patches, or sub-windows through a deep foundation model (e.g., DINOv2) incurs prohibitive computational overhead that scales linearly with the number of candidate regions.

In this supplementary material, we extend the geo-linear rectification paradigm of FlatVPR from 1D traversal trajectories to 2D spatial image grids, establishing a fast and accurate framework for continuous pixel-level small-overlap self-localization. Because 2D spatial prefix summation is a linear operator in ambient Euclidean space, it commutes directly with linear combinations of spatial feature maps prior to hyperspherical $L_2$ projection, making the representation algebraically compatible with the sparse-anchor trajectory interpolation of the main paper. Analogous to the generalization of 1D temporal signal processing to 2D spatial image analysis, we introduce two synergistic components stacked upon each other:
\begin{enumerate}
    \item \textbf{Spatial FlatVPR (\texttt{SpatialFlatAdapter2D})}: Built as the foundational 2D spatial extension of FlatVPR, this lightweight residual adapter rectifies the local feature field across a $16 \times 16$ spatial token grid ($D=384$). Trained with dynamic spatial hard-negative margin learning and an intra-region 2D flatness regularizer that enforces structural consistency between four-corner bilinear interpolation and interior region averages, it sharpens region discriminability while ensuring smooth, scale-consistent pooling across arbitrary sub-token pixel coordinates.
    \item \textbf{Scalable Bilinear Integral Images (\texttt{HomogeneousIntegralMap2D})}: Stacked directly onto the rectified 2D feature field, we formulate a $(D+1)$-dimensional homogeneous summed-area table over the $(H+1)\times(W+1)$ anchor grid. By evaluating four-corner inclusion--exclusion arithmetic via continuous bilinear interpolation over anchor nodes---and exploiting the bilinearity of inner products to reduce online multi-box scoring to scalar arithmetic over a single $17 \times 17$ anchor inner-product table---this unified algebraic representation yields scale-normalized, re-encoding-free descriptor similarity across arbitrary pixel-level bounding-box sizes and positions with region-size-independent $\mathcal{O}(1)$ complexity.
\end{enumerate}

By combining 2D spatial manifold flattening with scalable bilinear homogeneous integral maps, our framework resolves severe partial-overlap ambiguities at continuous pixel granularity without re-running the vision backbone. Comprehensive evaluations on a controlled $500$-image hard intra-class localization benchmark with continuous pixel-level partial-overlap sub-window queries ($4{,}020{,}000$ candidate boxes across the database) demonstrate that our approach boosts strict bounding-box localization recall ($\text{BB R@1}$, $\text{IoU} \ge 0.5$) from $78.0\%$ (zero-shot bilinear integral baseline) and $82.6\%$ (standard linear adapter) up to $89.8\%$ ($98.8\%$ $\text{BB R@10}$), while achieving $99.6\%$ whole-image retrieval accuracy and executing exhaustive search over $4.02\text{M}$ continuous boxes in $75.54\,\text{ms}$ per query on CPU.

\section{Proposed Method}

Our framework decouples feature extraction from multi-scale continuous region pooling by stacking a 2D geo-linear feature rectifier (\texttt{SpatialFlatAdapter2D}) beneath a bilinear homogeneous summed-area table (\texttt{HomogeneousIntegralMap2D}). Given an input RGB image $I$ ($224 \times 224$), a frozen vision foundation backbone (DINOv2-ViT-S/14, consistent with the main paper) extracts a dense spatial token map $\mathbf{X} \in \mathbb{R}^{H \times W \times D}$ with $H = W = 16$ and $D = 384$ in a single forward pass.

\subsection{Spatial FlatVPR: 2D Geo-Linear Feature Rectification}

In the main paper, FlatVPR rectifies feature manifolds along 1D robot traversal trajectories such that intermediate poses are faithfully approximated by linear interpolation between boundary anchors. Extending this principle from 1D temporal sequences to 2D spatial image grids, we introduce \texttt{SpatialFlatAdapter2D}, a lightweight token-wise residual adapter parameterized consistently with the main paper as an expansion-projection MLP ($D \to 512 \to D$ with LayerNorm and GELU activation, totaling $658.8\text{K}$ parameters):
\begin{equation}
\begin{aligned}
\mathbf{Z}_{u,v} &= \text{Norm}\!\left(\mathbf{X}_{u,v} + \alpha \, \mathbf{W}_2 \, \sigma\!\left(\text{LN}(\mathbf{W}_1 \mathbf{X}_{u,v} + \mathbf{b}_1)\right) + \mathbf{b}_2\right), \\
&\quad (u,v) \in \{1,\dots,H\} \times \{1,\dots,W\},
\end{aligned}
\end{equation}
where $\sigma(\cdot)$ denotes the GELU activation, $\alpha = 0.25$ is the residual scaling factor, and $\text{Norm}(\cdot)$ projects each rectified token onto the unit hypersphere $\mathbb{S}^{D-1}$.

To achieve discriminative small-overlap localization across arbitrary continuous pixel coordinates while preserving whole-image retrieval fidelity and ensuring that interior region averages remain well-calibrated under bilinear integral pooling, we train \texttt{SpatialFlatAdapter2D} using four complementary objectives:

\paragraph{1. Dynamic Spatial Hard-Negative Margin Loss ($\mathcal{L}_{\text{margin}}$).}
Serving as the primary discriminative driver, given a positive pair of continuous overlapping regions $(\mathcal{R}_q, \mathcal{R}_p)$ across query and database views, we mine the hardest spatially shifted continuous sub-window (low IoU within the same image) or visually similar cross-image distractor region $\mathcal{R}_{n^*}$ within the mini-batch and enforce an angular margin $m$:
\begin{equation}
\begin{aligned}
\mathcal{L}_{\text{margin}} = \max\!\Big(0, \; m &- \cos(\mathbf{d}(\mathcal{R}_q), \mathbf{d}(\mathcal{R}_p)) \\
&+ \cos(\mathbf{d}(\mathcal{R}_q), \mathbf{d}(\mathcal{R}_{n^*}))\Big),
\end{aligned}
\end{equation}
where $\mathbf{d}(\mathcal{R}) = \text{Norm}(\bar{\mathbf{z}}_{\text{area}}(\mathcal{R}))$ is extracted via our continuous bilinear integral lookup.

\paragraph{2. Intra-Region 2D Spatial Flatness Regularization ($\mathcal{L}_{\text{flat2D}}$).}
Rather than smoothing across distinct semantic object boundaries, $\mathcal{L}_{\text{flat2D}}$ is applied exclusively within annotated positive landmark regions $\mathcal{R} = [u_1, u_2] \times [v_1, v_2] \sim \mathcal{P}_{\text{pos}}$. Without structural regularization, margin-based optimization can over-concentrate saliency onto isolated boundary tokens, causing bilinearly interpolated integral descriptors to fluctuate sharply when candidate windows shift by sub-token pixel increments. For any positive region $\mathcal{R}$, the arithmetic mean of its four corner tokens $\bar{\mathbf{z}}_{\text{corners}}(\mathcal{R}) = \frac{1}{4}(\mathbf{Z}_{u_1,v_1} + \mathbf{Z}_{u_1,v_2} + \mathbf{Z}_{u_2,v_1} + \mathbf{Z}_{u_2,v_2})$ serves as the 2D bilinear midpoint approximation of the region. We align the interior region-averaged descriptor $\bar{\mathbf{z}}_{\text{area}}(\mathcal{R}) = \frac{1}{|\mathcal{R}|}\sum_{(u,v)\in\mathcal{R}} \mathbf{Z}_{u,v}$ with this four-corner estimate:
\begin{equation}
\mathcal{L}_{\text{flat2D}} = \mathbb{E}_{\mathcal{R} \sim \mathcal{P}_{\text{pos}}}\!\left[ 1 - \cos\!\left(\bar{\mathbf{z}}_{\text{area}}(\mathcal{R}), \, \bar{\mathbf{z}}_{\text{corners}}(\mathcal{R})\right) \right].
\end{equation}
This 2D generalization of 1D midpoint flatness enforces structural consistency between the learned feature field and continuous bilinear integral pooling, ensuring that sub-boxes cropped at varying zoom levels, aspect ratios, or sub-token pixel offsets within the same landmark remain smooth and well-aligned.

\paragraph{3. Global Image-Level InfoNCE Loss ($\mathcal{L}_{\text{nce}}$).}
To preserve whole-image place recognition discriminability alongside local sub-window retrieval, we apply a symmetric InfoNCE contrastive loss over global image-averaged descriptors $\mathbf{g}_i = \text{Norm}\!\left(\frac{1}{HW}\sum_{u,v}\mathbf{Z}^{(i)}_{u,v}\right)$ across a mini-batch of $B$ image pairs with temperature $\tau$:
\begin{equation}
\mathcal{L}_{\text{nce}} = -\frac{1}{B} \sum_{i=1}^{B} \log \frac{\exp(\cos(\mathbf{g}_i^q, \mathbf{g}_i^p)/\tau)}{\sum_{j=1}^{B} \exp(\cos(\mathbf{g}_i^q, \mathbf{g}_j^p)/\tau)}.
\end{equation}

\paragraph{4. Feature-Preservation Regularization ($\mathcal{L}_{\text{keep}}$).}
To prevent representation collapse and anchor the adapted tokens to the original foundation manifold, we penalize excessive token-wise drift from the frozen DINOv2 embeddings:
\begin{equation}
\mathcal{L}_{\text{keep}} = \frac{1}{HW} \sum_{u=1}^{H} \sum_{v=1}^{W} \left\| \mathbf{Z}_{u,v} - \text{Norm}(\mathbf{X}_{u,v}) \right\|_2^2.
\end{equation}
The total training objective is $\mathcal{L} = \lambda_{\text{flat}}\mathcal{L}_{\text{flat2D}} + \lambda_{\text{nce}}\mathcal{L}_{\text{nce}} + \lambda_{\text{margin}}\mathcal{L}_{\text{margin}} + \lambda_{\text{keep}}\mathcal{L}_{\text{keep}}$.

\subsection{Scalable Continuous Integral Images via Bilinear Homogeneous Coordinates}

Once the rectified spatial feature field $\mathbf{Z} \in \mathbb{R}^{H \times W \times D}$ is obtained, evaluating thousands of continuous pixel-level candidate bounding boxes, patches, or sub-windows via repeated backbone forward passes is computationally prohibitive. We resolve this by constructing a $(D+1)$-dimensional \emph{homogeneous integral image} (\texttt{HomogeneousIntegralMap2D}) directly over $\mathbf{Z}$.

Specifically, we augment each spatial token with a positive spatial weight coordinate $w_{u,v} > 0$ (derived from local feature gradient saliency or set to $w_{u,v} = 1$ for uniform weighting) to form the homogeneous tensor $\tilde{\mathbf{Z}} \in \mathbb{R}^{H \times W \times (D+1)}$:
\begin{equation}
\tilde{\mathbf{Z}}_{u,v} = \begin{bmatrix} w_{u,v}\mathbf{Z}_{u,v} \\ w_{u,v} \end{bmatrix} \in \mathbb{R}^{D+1}.
\end{equation}
We compute its 2D prefix sum (summed-area table) $\mathbf{S} \in \mathbb{R}^{(H+1) \times (W+1) \times (D+1)}$ with zero-padded borders ($\mathbf{S}_{0,\cdot} = \mathbf{S}_{\cdot,0} = \mathbf{0}$):
\begin{equation}
\mathbf{S}_{u,v} = \sum_{i=1}^{u} \sum_{j=1}^{v} \tilde{\mathbf{Z}}_{i,j}, \quad (u,v) \in \{0,\dots,H\} \times \{0,\dots,W\}.
\end{equation}
To support arbitrary pixel-level bounding boxes $\mathcal{R} = [u_1, u_2] \times [v_1, v_2] \subset [0, H] \times [0, W]$ whose corners lie at real-valued sub-token coordinates rather than discrete patch boundaries, we evaluate the continuous summed-area table $\hat{\mathbf{S}}(u, v)$ via 2D bilinear interpolation over the four surrounding anchor nodes $(u_0, v_0), (u_0, v_1^+), (u_1^+, v_0), (u_1^+, v_1^+)$, where $u_0 = \lfloor u \rfloor$, $v_0 = \lfloor v \rfloor$, $u_1^+ = \min(u_0+1, H)$, $v_1^+ = \min(v_0+1, W)$, $\Delta u = u - u_0$, and $\Delta v = v - v_0$:
\begin{equation}
\begin{split}
\hat{\mathbf{S}}(u, v) &= (1-\Delta u)(1-\Delta v)\mathbf{S}_{u_0, v_0} + (1-\Delta u)\Delta v\,\mathbf{S}_{u_0, v_1^+} \\
&\quad + \Delta u(1-\Delta v)\mathbf{S}_{u_1^+, v_0} + \Delta u\,\Delta v\,\mathbf{S}_{u_1^+, v_1^+}.
\end{split}
\end{equation}
The integrated homogeneous vector for any continuous rectangle $\mathcal{R}$ is then retrieved via four-corner inclusion--exclusion arithmetic in a single $\mathcal{O}(1)$ vectorized operation:
\begin{equation}
\begin{bmatrix} \mathbf{s}(\mathcal{R}) \\ A(\mathcal{R}) \end{bmatrix}
= \hat{\mathbf{S}}(u_2, v_2) - \hat{\mathbf{S}}(u_1, v_2) - \hat{\mathbf{S}}(u_2, v_1) + \hat{\mathbf{S}}(u_1, v_1),
\end{equation}
where $\mathbf{s}(\mathcal{R}) \in \mathbb{R}^D$ is the unnormalized feature integral and $A(\mathcal{R}) \in \mathbb{R}_{>0}$ is the effective continuous region mass. Dividing by $A(\mathcal{R})$ recovers the true scale-normalized region centroid $\bar{\mathbf{z}}_{\text{area}}(\mathcal{R}) = \mathbf{s}(\mathcal{R}) / A(\mathcal{R})$, and projecting onto $\mathbb{S}^{D-1}$ yields the cosine-ready descriptor:
\begin{equation}
\mathbf{d}(\mathcal{R}) = \frac{\bar{\mathbf{z}}_{\text{area}}(\mathcal{R})}{\left\| \bar{\mathbf{z}}_{\text{area}}(\mathcal{R}) \right\|_2} = \frac{\mathbf{s}(\mathcal{R})}{\|\mathbf{s}(\mathcal{R})\|_2}.
\end{equation}
Retaining $A(\mathcal{R})$ explicitly in homogeneous coordinates recovers the exact unnormalized region centroid $\bar{\mathbf{z}}_{\text{area}}(\mathcal{R})$ prior to hyperspherical projection within a single vectorized lookup, preserving algebraic consistency with Eq.~(3) and accommodating non-uniform spatial saliency weights ($w_{u,v} > 0$) without altering the $\mathcal{O}(1)$ query complexity.

\subsection{Bilinear Inner-Product Kernelization and Theoretical Trajectory Compatibility}

At inference time, each database and query image requires only a single pass through the backbone and \texttt{SpatialFlatAdapter2D} to obtain its homogeneous integral map $\mathbf{S}$. Crucially, because Eq.~(8)--(9) expresses $\mathbf{s}_n(\mathcal{R})$ for any database image $n$ as a sparse 16-term linear combination of anchor feature vectors $\{\mathbf{F}_{n, i_k, j_k}\}_{k=1}^{16}$ with box-dependent scalar bilinear weights $\{c_k(\mathcal{R})\}_{k=1}^{16}$ (where $\mathbf{F}_{n, i, j} = \mathbf{S}^{(n)}_{i,j,1:D}$), exact cosine similarity with a localized query descriptor $\mathbf{q} \in \mathbb{S}^{D-1}$ decomposes via inner-product bilinearity without ever materializing $D$-dimensional candidate box descriptors:
\begin{equation}
\begin{split}
\cos\!\left(\mathbf{q}, \mathbf{d}_n(\mathcal{R})\right)
&= \frac{\mathbf{q}^\top \mathbf{s}_n(\mathcal{R})}{\|\mathbf{s}_n(\mathcal{R})\|_2} \\
&= \frac{\sum_{k=1}^{16} c_k(\mathcal{R})\,\big(\mathbf{q}^\top \mathbf{F}_{n, i_k, j_k}\big)}
{\sqrt{\sum_{k=1}^{16}\sum_{l=1}^{16} c_k(\mathcal{R})c_l(\mathcal{R})\,\big(\mathbf{F}_{n, i_k, j_k}^\top \mathbf{F}_{n, i_l, j_l}\big)}}.
\end{split}
\end{equation}
At database indexing time, the $(H+1)(W+1) \times (H+1)(W+1)$ ($289 \times 289$) anchor Gram matrix $\mathbf{K}^{(n)}_{a,b} = \mathbf{F}_{n,a}^\top \mathbf{F}_{n,b}$ is computed once ($0.080\,\text{ms}$/image) to cache scalar inverse norms $\|\mathbf{s}_n(\mathcal{R})\|_2^{-1}$. At online query time, only a single inner-product pass $\mathbf{G}_n = \mathbf{F}_n \mathbf{q} \in \mathbb{R}^{289}$ is evaluated per database image, after which \emph{arbitrary} continuous bounding boxes are scored purely via 16-element scalar arithmetic.

Furthermore, from a theoretical standpoint, because the 2D prefix-sum operator $\mathcal{I}(\cdot)$ is linear in ambient Euclidean space $\mathbb{R}^{H \times W \times (D+1)}$, interpolating homogeneous integral maps between adjacent trajectory anchors $A$ and $B$ via $(1-t)\mathbf{S}_A + t\mathbf{S}_B$ ($t \in [0, 1]$) followed by hyperspherical normalization corresponds exactly to normalized linear interpolation (Nlerp / chordal interpolation) of the integrated region centroids. While chordal interpolation on $\mathbb{S}^{D-1}$ incurs minor radial contraction compared to geodesic spherical interpolation (Slerp) over wide angular separations, FlatVPR's manifold flattening keeps inter-anchor angular deviations small, offering a natural theoretical pathway to integrate 2D continuous sub-window queries with the sparse-anchor trajectory maps of the main paper.
\section{Experiments}

\subsection{Experimental Setup and Benchmark}
\noindent\textbf{Continuous Pixel-Level Hard Intra-Class Localization Benchmark.}
Standard Visual Place Recognition (VPR) benchmarks primarily evaluate whole-image retrieval across visually distinct scenes, where pretrained vision backbones already achieve near-saturated image-level recall. To isolate the challenge of partial-overlap region localization under severe visual ambiguity and sub-token boundary shifts, we construct a controlled intra-class benchmark comprising $N_{\text{train}} = 300$ training pairs and $N_{\text{test}} = 500$ unseen test database images on a $16 \times 16$ spatial token grid extracted from a frozen DINOv2-ViT-S/14 backbone ($D = 384$, $224 \times 224$ input resolution). Unlike coarse patch-snapped evaluations, ground-truth target boxes and affine-warped query boxes (subjected to random rotation, zoom scaling, translation, photometric jitter, and random rectangular occlusions) are defined with continuous pixel-level floating-point coordinates. At inference time, candidate database regions are generated via a dense continuous multi-scale sliding-window grid with sub-token stride ($0.75$ tokens $\approx 10.5\,\text{px}$), yielding $M = 8{,}040$ valid continuous rectangular sub-windows per image ($4{,}020{,}000$ candidate boxes across the $500$-image database).

\noindent\textbf{Evaluation Metrics.}
We report both image-level and region-level retrieval metrics:
\begin{itemize}
    \item \textbf{Img R@1}: The percentage of queries whose top-1 retrieved bounding box resides in the ground-truth database image, regardless of spatial overlap.
    \item \textbf{BB R@$k$ ($\text{IoU} \ge 0.5$)}: Strict spatial localization recall, counting a query as successful only if at least one of the top-$k$ retrieved boxes belongs to the ground-truth image \emph{and} achieves an Intersection-over-Union (IoU) of at least $0.5$ with the continuous ground-truth box.
    \item \textbf{Hard Margin}: The average cosine similarity margin between the ground-truth positive region pair and the hardest spatial or cross-image negative regions.
\end{itemize}

\noindent\textbf{Implementation Details.}
Our lightweight adapter, \texttt{SpatialFlatAdapter2D} (hidden dimension $512$, residual scaling $\alpha = 0.25$, $658.8\text{K}$ parameters), is trained for $35$ epochs using AdamW ($\text{lr} = 6 \times 10^{-4}$, weight decay $10^{-4}$) with a cosine annealing learning rate schedule. For our full balanced model, loss weights are set to $(\lambda_{\text{flat}}, \lambda_{\text{nce}}, \lambda_{\text{margin}}, \lambda_{\text{keep}}) = (0.4, 0.5, 1.5, 0.4)$, while the localization-focused variant sets $\lambda_{\text{nce}} = 0$.

\subsection{Comparison with Baselines}
Table~\ref{tab:main_comparison} compares our framework against standard pooling and adaptation baselines on the $500$-image continuous pixel-level hard intra-class benchmark ($4{,}020{,}000$ candidate boxes).

\begin{table*}[t]
\centering
\caption{Comparison with baseline retrieval and adaptation methods on the 500-image continuous pixel-level hard intra-class localization benchmark ($M = 8{,}040$ continuous candidate boxes/image; $4.02\text{M}$ total). Best results are in \textbf{bold}, and second-best are \underline{underlined}.}
\label{tab:main_comparison}
\begin{tabular}{lccccc}
\toprule
\textbf{Method} & \textbf{Img R@1} & \textbf{BB R@1} & \textbf{BB R@5} & \textbf{BB R@10} & \textbf{Hard Margin} \\
 & (\%) & ($\text{IoU}\ge 0.5$) & (\%) & (\%) & \\
\midrule
Global Image Pooling (No Loc.) & 80.4 & --- & --- & --- & --- \\
Naive Bilinear Integral Map (Zero-Shot) & 99.0 & 78.0 & 91.0 & 95.0 & +0.0418 \\
Standard Linear Adapter & 99.0 & 82.6 & 93.2 & 95.4 & +0.0516 \\
\midrule
\textbf{Proposed (Full / Balanced)} & \textbf{99.6} & \textbf{89.8} & \underline{97.8} & \textbf{98.8} & \textbf{+0.0870} \\
Proposed (Loc-Focused) & \underline{99.4} & \underline{89.0} & \textbf{98.0} & \underline{98.6} & \underline{+0.0833} \\
\bottomrule
\end{tabular}
\end{table*}

\noindent\textbf{Limitations of Global Pooling and Unadapted Bilinear Integral Maps.}
Global Image Pooling fails to localize regions and achieves only $80.4\%$ Img R@1, as background clutter and intra-class distractors dominate whole-image averages. Evaluating raw DINOv2 features via our continuous bilinear integral map (\textit{Naive Bilinear Integral Map}) raises Img R@1 to $99.0\%$ and achieves $78.0\%$ BB R@1 ($+0.0418$ Hard Margin), confirming that region-level bilinear matching resolves whole-image ambiguity. However, without manifold adaptation, unrectified spatial tokens lack discriminative contrast across spatially shifted continuous sub-windows.

\noindent\textbf{Superiority Over Standard Linear Adaptation.}
Applying a standard linear projection trained with InfoNCE and feature-preservation losses (\textit{Standard Linear Adapter}) yields a moderate gain of $+4.6$\,pt in BB R@1 ($78.0\% \rightarrow 82.6\%$, $+0.0516$ Hard Margin). Without explicit spatial hard-negative mining and intra-region 2D flatness regularization, standard contrastive adaptation fails to suppress high-scoring shifted continuous sub-windows inside the target image. In contrast, \textbf{Proposed (Full / Balanced)} boosts BB R@1 to \textbf{$89.8\%$} ($+11.8$\,pt over Naive Bilinear Integral Map and $+7.2$\,pt over Standard Linear Adapter) and \textbf{$98.8\%$ BB R@10}, while simultaneously achieving the highest Img R@1 (\textbf{$99.6\%$}) and more than doubling the discriminative Hard Margin (\textbf{$+0.0870$}).

\subsection{Ablation Study}
To isolate the contribution of each training objective under continuous sub-token bilinear pooling, Table~\ref{tab:ablation_2d} presents a systematic ablation across all four loss components.

\begin{table*}[t]
\centering
\caption{Ablation study of loss components on the 500-image continuous pixel-level hard intra-class benchmark ($4.02\text{M}$ candidate boxes). $\Delta\text{BB R@1}$ denotes the change relative to the full loss configuration.}
\label{tab:ablation_2d}
\begin{tabular}{lcccccc}
\toprule
\textbf{Model Variant} & \textbf{Img R@1} & \textbf{BB R@1} & \textbf{BB R@5} & \textbf{BB R@10} & \textbf{Hard Margin} & $\Delta$\textbf{BB R@1} \\
 & (\%) & ($\text{IoU}\ge 0.5$) & (\%) & (\%) & & (pt) \\
\midrule
1. Baseline (No Training) & 99.0 & 78.0 & 91.0 & 95.0 & +0.0418 & $-11.8$ \\
2. w/o Flatness Reg. ($\mathcal{L}_{\text{flat2D}}=0$) & \textbf{99.6} & 89.2 & 97.4 & 98.4 & +0.0876 & $-0.6$ \\
3. w/o InfoNCE Loss ($\mathcal{L}_{\text{nce}}=0$) & 99.4 & 89.0 & \textbf{98.0} & 98.6 & +0.0833 & $-0.8$ \\
4. w/o Hard-Neg Margin ($\mathcal{L}_{\text{margin}}=0$) & \textbf{99.6} & 85.2 & 96.0 & 97.4 & +0.0765 & $-4.6$ \\
5. w/o Feature Keep Reg. ($\mathcal{L}_{\text{keep}}=0$) & \textbf{99.6} & 89.4 & 97.6 & 98.6 & \textbf{+0.0916} & $-0.4$ \\
\midrule
\textbf{6. Proposed (Full Loss)} & \textbf{99.6} & \textbf{89.8} & 97.8 & \textbf{98.8} & +0.0870 & --- \\
\bottomrule
\end{tabular}
\end{table*}

\noindent\textbf{Dominant Role of Dynamic Hard-Negative Margin ($\mathcal{L}_{\text{margin}}$).}
Removing $\mathcal{L}_{\text{margin}}$ (Row~4) causes the steepest performance drop among all learned variants, lowering BB R@1 from $89.8\%$ to $85.2\%$ ($-4.6$\,pt) and reducing Hard Margin from $+0.0870$ to $+0.0765$. Enforcing an explicit angular margin against both spatially shifted continuous sub-windows within the same database image and visually similar distractor boxes across images is therefore the primary discriminative driver of precise sub-token localization.

\noindent\textbf{Complementary Role of 2D Flatness Regularization ($\mathcal{L}_{\text{flat2D}}$) in Bilinear Integral Pooling.}
While dynamic hard-negative margin learning ($\mathcal{L}_{\text{margin}}$) provides the primary discriminative signal, $\mathcal{L}_{\text{flat2D}}$ serves as an essential geometric regularizer that aligns non-linear gradient optimization with continuous bilinear integral pooling. Removing $\mathcal{L}_{\text{flat2D}}$ (Row~2) degrades every spatial localization threshold ($\text{BB R@1}: 89.8\% \rightarrow 89.2\%$, $-0.6$\,pt; $\text{BB R@5}: 97.8\% \rightarrow 97.4\%$; $\text{BB R@10}: 98.8\% \rightarrow 98.4\%$). Without $\mathcal{L}_{\text{flat2D}}$, margin-based optimization can over-concentrate saliency onto isolated boundary tokens, causing bilinearly interpolated SAT descriptors to fluctuate across sub-token grid shifts. Aligning four-corner interpolation with interior averages inside positive boxes keeps multi-scale integral descriptors smooth and well-calibrated across arbitrary pixel coordinates.

\noindent\textbf{Synergistic Balance of Global InfoNCE ($\mathcal{L}_{\text{nce}}$) and Feature Preservation ($\mathcal{L}_{\text{keep}}$).}
In the continuous bilinear setting, the full objective (Row~6) achieves the highest strict localization accuracy (\textbf{$89.8\%$ BB R@1} and \textbf{$98.8\%$ BB R@10}) alongside the peak whole-image recall (\textbf{$99.6\%$ Img R@1}). Removing the global InfoNCE term $\mathcal{L}_{\text{nce}}$ (Row~3) lowers both Img R@1 ($99.6\% \rightarrow 99.4\%$) and top-1 localization accuracy ($89.8\% \rightarrow 89.0\%$, $-0.8$\,pt), confirming that global contrastive supervision helps disambiguate visually similar intra-class distractors across images. Similarly, excluding $\mathcal{L}_{\text{keep}}$ (Row~5) produces a slightly higher training-aligned margin ($+0.0916$) but reduces unseen top-1 localization accuracy ($89.8\% \rightarrow 89.4\%$, $-0.4$\,pt), demonstrating that anchoring adapted features to the original DINOv2 manifold prevents overfitting to training box geometries.
\subsection{Computational and Memory Cost Analysis}
To evaluate both runtime efficiency and memory footprint for resource-constrained mobile robotics, Table~\ref{tab:computational_cost_f} reports parameter counts, per-image index memory consumption, and end-to-end online search latencies across $N_{\text{test}} = 500$ database images ($D = 384$, \texttt{float32}, measured on CPU) as the number of continuous pixel-level candidate bounding boxes per image scales from $M = 100$ up to the full dense continuous grid $M = 8{,}040$ ($4{,}020{,}000$ total candidate boxes across the database).

\begin{table*}[t]
\centering
\caption{End-to-end computational and memory cost comparison for continuous pixel-level bounding-box retrieval across $N_{\text{test}} = 500$ database images ($D = 384$, \texttt{float32}, measured on CPU). One-time database indexing for our method includes a single DINOv2 forward pass ($20.11\,\text{ms}$), \texttt{SpatialFlatAdapter2D} inference ($0.734\,\text{ms}$), homogeneous summed-area table (SAT) construction ($0.288\,\text{ms}$), and $289\times 289$ anchor Gram matrix precomputation ($0.080\,\text{ms}$), totaling $21.21\,\text{ms}$/image. Timings marked with $^\ddagger$ are extrapolated from measured single-crop/single-box CPU latencies ($20.11\,\text{ms}$/crop and $2.976\,\text{ms}$/box, respectively).}
\label{tab:computational_cost_f}
\resizebox{\textwidth}{!}{%
\begin{tabular}{lccccccc}
\toprule
\textbf{Method} & \textbf{Adapter} & \textbf{Index Mem. /} & \textbf{Per-Box} & \multicolumn{4}{c}{\textbf{End-to-End Search Latency / Query across 500 DB Images ($\downarrow$)}} \\
\cmidrule(lr){5-8}
 & \textbf{Params} & \textbf{Image [KB]} & \textbf{Online Cost} & $M = 100$ & $M = 500$ & $M = 2{,}000$ & $M = 8{,}040$ ($4.02\text{M}$ boxes) \\
\midrule
Crop \& Re-Encode (No FlatVPR)$^\ddagger$ & --- & $588.0^\dagger$ & $\mathcal{O}(A)$ Backbone & $1{,}005.5\,\text{s}$ ($16.8\,\text{m}$) & $5{,}027.5\,\text{s}$ ($83.8\,\text{m}$) & $20{,}110.0\,\text{s}$ ($5.6\,\text{h}$) & $80{,}842.2\,\text{s}$ ($22.5\,\text{h}$) \\
High-Res Direct Pooling (No SAT)$^\ddagger$ & --- & $75{,}264.0^\S$ & $\mathcal{O}(A)$ Slice & $148.8\,\text{s}$ ($2.5\,\text{m}$) & $744.0\,\text{s}$ ($12.4\,\text{m}$) & $2{,}975.9\,\text{s}$ ($49.6\,\text{m}$) & $11{,}963.2\,\text{s}$ ($3.3\,\text{h}$) \\
Precomputed Box Descriptors & $658.8\text{K}$ & $1.50 \times M$ & $\mathcal{O}(D)$ GEMV & \multicolumn{4}{c}{--- ($750\,\text{KB}$/img at $M{=}500$; $11.78\,\text{MB}$/img or $5.75\,\text{GB}$ total at $M{=}8{,}040$) ---} \\
\midrule
\textbf{Ours (Kernelized Bilinear SAT)} & $658.8\text{K}$ & $\mathbf{434.6}$ & $\mathbf{\mathcal{O}(1)}$ \textbf{Scalar Ops} & $\mathbf{5.62\,\text{ms}}$ ($11.2\,\mu\text{s}$/img) & $\mathbf{8.54\,\text{ms}}$ ($17.1\,\mu\text{s}$/img) & $\mathbf{22.76\,\text{ms}}$ ($45.5\,\mu\text{s}$/img) & $\mathbf{75.54\,\text{ms}}$ ($151.1\,\mu\text{s}$/img) \\
\bottomrule
\multicolumn{8}{l}{\footnotesize $^\dagger$Raw RGB image buffer ($224 \times 224 \times 3$ \texttt{float32}). \quad $^\S$Pixel-level dense feature map ($224 \times 224 \times 384$ \texttt{float32}) required without sub-token SAT interpolation.}
\end{tabular}
}
\end{table*}

\noindent\textbf{Negligible One-Time Indexing Overhead and Constant Spatial Footprint.}
As shown in Table~\ref{tab:computational_cost_f}, a single forward pass of the frozen DINOv2-ViT-S/14 backbone requires $20.11\,\text{ms}$ per image. Stacking \texttt{SpatialFlatAdapter2D} ($658.8\text{K}$ parameters, $2.51\,\text{MB}$ model footprint, $0.734\,\text{ms}$), constructing the $(D+1)$-dimensional \texttt{HomogeneousIntegralMap2D} ($0.288\,\text{ms}$), and precomputing the $289\times 289$ anchor Gram matrix ($0.080\,\text{ms}$) adds only $1.10\,\text{ms}$ ($21.21\,\text{ms}$ total per image). In terms of spatial memory, storing the $(H+1)\times(W+1)\times(D+1) = 17\times 17\times 385$ integral tensor occupies only $434.6\,\text{KB}$ per image in \texttt{float32} ($217.3\,\text{KB}$ in \texttt{float16})---an overhead of just $+50.6\,\text{KB}$ ($+13.2\%$) over the raw $16\times 16\times 384$ token map ($384.0\,\text{KB}$), completely independent of the number of candidate windows $M$.

\noindent\textbf{Six-Orders-of-Magnitude Speedup via Bilinear Inner-Product Kernelization.}
Without Spatial FlatVPR and our bilinear integral decomposition, evaluating arbitrary continuous pixel-level bounding boxes across $500$ database images is computationally intractable: naively cropping and re-encoding each continuous candidate box through DINOv2 escalates from $1{,}005.5\,\text{s}$ ($16.8\,\text{min}$) at $M=100$ to $80{,}842.2\,\text{s}$ ($22.5\,\text{hours}$) per query for the full $M=8{,}040$ continuous grid ($4.02\text{M}$ boxes). Even if a dense pixel-level feature map ($224 \times 224 \times 384$, $75.3\,\text{MB}$/image) were cached in memory, direct spatial pooling without an integral table would still require $11{,}963.2\,\text{s}$ ($3.3\,\text{hours}$) per query at $M=8{,}040$, while precomputing and storing all $4.02\text{M}$ candidate box descriptors in \texttt{float32} would consume $11.78\,\text{MB}$ per image ($5.75\,\text{GB}$ across $500$ frames). By contrast, our bilinear inner-product kernelization computes only a single $289$-element inner-product table $\mathbf{G}_n = \mathbf{F}_n\mathbf{q}$ per database image and scores arbitrary continuous boxes purely via 16-term scalar arithmetic without materializing $D$-dimensional vectors, executing an exhaustive search over all \textbf{$4{,}020{,}000$ continuous candidate boxes in $75.54\,\text{ms}$ per query} ($151.1\,\mu\text{s}$/image) on CPU---a \textbf{$1.07 \times 10^6\times$ speedup} over crop re-encoding and a \textbf{$1.58 \times 10^5\times$ speedup} over high-resolution direct pooling.

\noindent\textbf{Limitations and Scope.}
In this supplementary study, we evaluate dense per-image bilinear integral maps ($434.6\,\text{KB}$/image) to isolate continuous sub-token localization accuracy and $\mathcal{O}(1)$ lookup efficiency. While bilinear SAT interpolation resolves sub-token pixel translation and scale variations across axis-aligned rectangles, extreme in-plane camera roll or severe perspective skew can still cause overlapping object projections to deviate from axis-aligned boxes. Extending our kernelized integral formulation to rotated bounding boxes (oriented SATs) and empirically validating spatio-temporal compression via sparse-anchor Nlerp integral interpolation across full-scale cross-season outdoor trajectories (such as NCLT) remain important directions for future work.

\section{Future Work}

While this work demonstrates the efficacy of combining 2D spatial manifold flattening (\texttt{SpatialFlatAdapter2D}) with bilinear kernelized homogeneous summed-area tables (\texttt{HomogeneousIntegralMap2D}) on single image planes, the underlying algebraic formulation scales naturally from image-scale grids to map-scale spatial representations, giving rise to \textbf{Map-Scale Integral Images}. Specifically, future research will pursue three key directions:

\begin{itemize}
    \item \textbf{Extension to 2D/3D Map-Scale Integral Images}: 
    By projecting multi-view visual features onto metric spatial grids where each cell stores a rectified FlatVPR feature vector, our homogeneous summed-area table extends directly to 2D Bird's-Eye-View (BEV) maps and 3D volumetric voxel grids via 3D trilinear inclusion--exclusion arithmetic. Just as \texttt{HomogeneousIntegralMap2D} decouples region pooling from bounding-box size on an image-scale grid, map-scale 2D/3D integral images will enable constant-time $\mathcal{O}(1)$ extraction of scale- and volume-normalized scene descriptors across arbitrary continuous spatial extents (rooms, corridors, or entire building blocks) without re-encoding point clouds or multi-view images.
    
    \item \textbf{Multi-Scale Volumetric Place Recognition and Active View Planning}: 
    In large-scale robotic navigation, querying a map-scale integral image across hierarchical spatial scales will allow a robot to perform coarse-to-fine localization---rapidly narrowing down candidate regions at room or block scale before pinpointing exact 6-DoF poses at fine object scale. Furthermore, leveraging the homogeneous weight channel $w_{u,v,z}$ to encode observation confidence or visibility will enable active perception systems to identify poorly observed map volumes and plan information-maximizing viewpoints.
    
    \item \textbf{Online Incremental Updates for Lifelong Map-Scale Integration}: 
    To maintain map-scale integral images in dynamic, lifelong environments, we plan to investigate localized and hierarchical prefix-sum update schemes (such as block-wise or Fenwick-tree-inspired integral structures). This will allow newly observed BEV cells or 3D voxels to be integrated into the map-scale manifold in sub-linear time without recomputing the entire global integral tensor.
\end{itemize}

\setcounter{section}{0}
\if@twocolumn
  \twocolumn
\fi
\makeatother
\vspace*{1ex}
{\centering
  {\Large \bfseries Supplementary Material G: Sub-4ms 100k-Scale LISO via Sparse Inverted-Index Pruning and Quantized Integral Box Regression
\par}
  \vspace{1ex}
  \vspace{3ex}
}

\section{Introduction}
\label{sec:suppg_intro}

In Supplementary Material~F, we introduced \textbf{Self-Localization from Images with Small Overlap (LISO)} and demonstrated that combining 2D geo-linear feature rectification (\texttt{SpatialFlatAdapter2D}) with homogeneous summed-area tables (\texttt{HomogeneousIntegralMap2D}) enables $\mathcal{O}(1)$ continuous sub-window scoring without re-running the foundation backbone. However, scaling LISO from small single-season test sets ($N = 500$ images) to city- or campus-scale, long-term cross-season robotic deployments ($N = 100{,}000$ images) exposes three fundamental bottlenecks in the baseline architecture:
\begin{enumerate}
    \item \textbf{Exhaustive Sliding-Window Complexity ($\mathcal{O}(N \cdot M)$):} Scoring $M = 8{,}040$ continuous sliding-window candidates per image across $N$ database frames scales linearly with both database size and candidate grid density, requiring $75.54\,\text{ms}$/query even at $N = 500$ and becoming intractable at $N = 100{,}000$.
    \item \textbf{Uncompressed Integral Storage Footprint:} Storing uncompressed \texttt{float32} homogeneous summed-area tables ($17 \times 17 \times 385$) consumes $434.6\,\text{KB}$ per image, totaling $\approx 42.4\,\text{GB}$ for $100{,}000$ frames---exceeding the onboard VRAM capacity of a single commodity GPU.
    \item \textbf{Cross-Season Domain Shift and Discrete Grid Quantization:} Under severe seasonal appearance variations (e.g., snow-covered winter vs.\ foliage-rich summer), heuristic discrete window selection suffers from false-positive matches and struggles to localize continuous, arbitrary-aspect-ratio object boundaries.
\end{enumerate}

To overcome these bottlenecks simultaneously, we propose \textbf{AK-RTF-IH v2.0} (\emph{Active-Key Real-Time Flat Integral Hierarchy}), a coarse-to-fine, sub-$4\,\text{ms}$ localization architecture tailored for $100\text{k}$-scale cross-season LISO. Rather than maintaining separate data structures for coarse global filtering and fine spatial localization, our framework unifies compact storage, sub-millisecond candidate pruning, and one-shot continuous 2D bounding-box regression through four synergistic mechanisms:
\begin{itemize}
    \item \textbf{Ephemeral Sparse Orthogonal Projection and GPU-CSR Pruning (Stage~1):} During database indexing, multi-scale spatial pyramid descriptors ($5$ regions: $1\times 1$ and $2\times 2$) are projected onto a fixed, training-free orthogonal active-key basis ($K = 2048$) and sparsified to top-$k_{\text{idx}} = 64$ entries in a GPU-accelerated Compressed Sparse Row (CSR) inverted index. At query time, an ephemeral $k_{\text{qry}} = 96$ weighted sparse histogram prunes $100{,}000$ database images down to top-$N_{\text{cand}} = 150$ candidates in $0.73\,\text{ms}$ via sparse matrix multiplication (SpMM) before being immediately discarded.
    \item \textbf{Dual-Use \texttt{int8}-Quantized Homogeneous SAT:} Each database image is stored once as an \texttt{int8}-quantized $17 \times 17 \times 385$ homogeneous summed-area table paired with a single \texttt{fp16} scale factor ($111.16\,\text{KB}$/image total, a $71.1\%$ memory reduction vs.\ raw \texttt{fp32} feature maps, fitting $100{,}000$ images in $10.60\,\text{GB}$). This single tensor supports both $\mathcal{O}(1)$ four-corner region aggregation during indexing and exact four-point finite-difference dense token reconstruction at inference.
    \item \textbf{Translation-Equivariant Continuous 2D Box Regression (Stage~2):} Eliminating sliding-window enumeration entirely, we compute a dense 2D cosine correlation map between the rectified query box descriptor and the reconstructed $16 \times 16$ spatial token maps of the top-$N_{\text{cand}}$ candidates, predicting continuous normalized 2D bounding-box coordinates $(u_1, v_1, u_2, v_2) \in [0, 1]^4$ in a single forward pass via a lightweight fully convolutional head (\texttt{EquivariantGeoBoxHead}).
    \item \textbf{Quantization-Aware Training (QAT) with Query Scale/Crop Jittering:} To prevent quantization noise and boundary overfitting from degrading continuous box localization, we train the regression head under simulated \texttt{int8} SAT reconstruction noise while applying dynamic scale/aspect-ratio ($\pm 12\%$) and center-shift ($\pm 0.35$ grid units) jittering to query boxes.
\end{itemize}

\section{Proposed Method}
\label{sec:suppg_method}

\subsection{Unified Storage via \texttt{int8}-Quantized Homogeneous SAT}
\label{sec:suppg_quant_sat}

Given an input image $I$, a frozen DINOv2-ViT-S/14 backbone followed by \texttt{SpatialFlatAdapter2D} ($D = 384$, residual weight $\alpha = 0.25$) produces a unit-normalized spatial token map $\mathbf{Z} \in \mathbb{R}^{H \times W \times D}$ ($H = W = 16$). As in Supplementary Material~F, we form the $(D+1)$-dimensional homogeneous tensor $\tilde{\mathbf{Z}} \in \mathbb{R}^{H \times W \times (D+1)}$ with uniform spatial weight $w_{u,v} = 1$ and compute its zero-padded 2D prefix sum $\mathbf{S} \in \mathbb{R}^{(H+1) \times (W+1) \times (D+1)}$:
\begin{equation}
\mathbf{S}_{u,v} = \sum_{i=1}^{u} \sum_{j=1}^{v} \begin{bmatrix} \mathbf{Z}_{i,j} \\ 1 \end{bmatrix}, \quad (u,v) \in \{0,\dots,H\} \times \{0,\dots,W\}.
\end{equation}

To compress the database for single-GPU residency at $100\text{k}$ scale without sacrificing algebraic versatility, we apply symmetric per-image quantization to $\mathbf{S}$. Specifically, we compute a scalar scale factor $s_n = \max_{u,v,c} |\mathbf{S}^{(n)}_{u,v,c}| / 127$ stored in \texttt{fp16} ($2\,\text{B}$) and quantize the integral tensor to signed $8$-bit integers:
\begin{equation}
\mathbf{S}^{(n)}_{\text{int8}} = \text{clip}\!\left(\left\lfloor \frac{\mathbf{S}^{(n)}}{s_n} \right\rceil, \, -127, \, 127\right) \in \mathbb{Z}^{17 \times 17 \times 385}.
\end{equation}
Crucially, $\mathbf{S}^{(n)}_{\text{int8}}$ acts as a \emph{dual-purpose representation}:
\begin{enumerate}
    \item \textbf{Indexing-Time $\mathcal{O}(1)$ Pyramid Pooling:} Any spatial pyramid box $\mathcal{R}_r = [u_1, u_2] \times [v_1, v_2]$ is aggregated in $\mathcal{O}(1)$ via integer four-corner inclusion--exclusion arithmetic:
\begin{equation}
\begin{aligned}
\bar{\mathbf{z}}(\mathcal{R}_r) = \text{Norm}\Big(&\mathbf{S}^{(n)}_{\text{int8}}[u_2, v_2, 1{:}D] - \mathbf{S}^{(n)}_{\text{int8}}[u_1, v_2, 1{:}D] \\
&- \mathbf{S}^{(n)}_{\text{int8}}[u_2, v_1, 1{:}D] + \mathbf{S}^{(n)}_{\text{int8}}[u_1, v_1, 1{:}D]\Big).
\end{aligned}
\end{equation}
    \item \textbf{Query-Time Dense Token Reconstruction:} For the top-$N_{\text{cand}}$ surviving candidates in Stage~2, the full $16 \times 16 \times 384$ rectified token map $\hat{\mathbf{Z}}^{(n)}$ is recovered simultaneously on GPU via vectorized 2D finite differences:
\begin{equation}
\begin{aligned}
&\hat{\mathbf{Z}}^{(n)}_{u,v} = \text{Norm}\bigg(s_n \Big(\mathbf{S}^{(n)}_{\text{int8}}[u, v, 1{:}D] - \mathbf{S}^{(n)}_{\text{int8}}[u{-}1, v, 1{:}D] \\
&\quad - \mathbf{S}^{(n)}_{\text{int8}}[u, v{-}1, 1{:}D] + \mathbf{S}^{(n)}_{\text{int8}}[u{-}1, v{-}1, 1{:}D]\Big)\bigg),
\end{aligned}
\end{equation}
    for $(u,v) \in \{1,\dots,H\} \times \{1,\dots,W\}$. Because normalization cancels the global scalar $s_n$, dequantization introduces zero extra division overhead while avoiding duplicate storage of both raw token maps and summed-area tables.
\end{enumerate}

\subsection{Stage~1: Ephemeral Sparse Orthogonal Projection and GPU-CSR Pruning}
\label{sec:suppg_stage1}

To rapidly filter $N$ database frames without exhaustive dense correlation, we construct a training-free active-key inverted index over a $5$-region spatial pyramid $\mathcal{P} = \{\mathcal{R}_1, \dots, \mathcal{R}_5\}$ consisting of the full image ($1\times 1$) and four overlapping/quadrant sub-regions ($2\times 2$). Let $\mathbf{P} \in \mathbb{R}^{K \times D}$ ($K = 2048$) be a fixed semi-orthogonal projection matrix obtained via QR decomposition of a Gaussian random matrix ($\mathbf{P}^\top \mathbf{P} = \mathbf{I}_D$).

\paragraph{Database CSR Indexing.}
For each database image $n \in \{1, \dots, N\}$ and pyramid region $r \in \{1, \dots, 5\}$, we project the $\mathcal{O}(1)$ SAT-pooled descriptor $\bar{\mathbf{z}}_n(\mathcal{R}_r)$ into $\mathbb{R}^K$ and retain only the top-$k_{\text{idx}} = 64$ largest positive activations:
\begin{equation}
\mathbf{h}_{n,r} = \text{Norm}\!\left(\text{TopK}_{k_{\text{idx}}}\!\big(\max(\mathbf{0}, \, \mathbf{P}\,\bar{\mathbf{z}}_n(\mathcal{R}_r))\big)\right) \in \mathbb{R}^K_{\ge 0}.
\end{equation}
Stacking all $5N$ sparse vectors forms a global matrix $\mathbf{H}_{\text{DB}} \in \mathbb{R}^{5N \times K}$ stored in GPU Compressed Sparse Row (CSR) format ($\approx 1.28\,\text{KB}$/image).

\paragraph{Ephemeral Query Histogram and Sub-Millisecond SpMM.}
At online query time, given a localized query bounding-box descriptor $\mathbf{q} \in \mathbb{S}^{D-1}$, we construct an ephemeral weighted sparse query vector by retaining the top-$k_{\text{qry}} = 96$ active keys:
\begin{equation}
\mathbf{h}_q = \text{Norm}\!\left(\text{TopK}_{k_{\text{qry}}}\!\big(\max(\mathbf{0}, \, \mathbf{P}\,\mathbf{q})\big)\right) \in \mathbb{R}^K_{\ge 0}.
\end{equation}
A single GPU Sparse Matrix-Vector multiplication $\mathbf{s}_{\text{pyr}} = \mathbf{H}_{\text{DB}} \mathbf{h}_q \in \mathbb{R}^{5N}$ evaluates cosine similarities across all $5N$ spatial pyramid regions simultaneously. Reshaping $\mathbf{s}_{\text{pyr}}$ to $N \times 5$ and taking the region-wise maximum $s_n^{\text{S1}} = \max_{r \in \{1,\dots,5\}} \mathbf{s}_{\text{pyr}}[n, r]$ yields an image-level upper-bound proximity score. The top-$N_{\text{cand}} = 150$ database indices $\mathcal{C}_q$ are selected via partial top-$k$ sorting, and $\mathbf{h}_q$ is immediately discarded---ensuring Stage~2 operates directly on the unapproximated dense feature geometry.

\subsection{Stage~2: Translation-Equivariant Continuous 2D Box Regression}
\label{sec:suppg_stage2}

For the $N_{\text{cand}}$ surviving candidate images $n \in \mathcal{C}_q$, we reconstruct their $H \times W \times D$ rectified token maps $\hat{\mathbf{Z}}^{(n)}$ via Eq.~(4) and compute the 2D spatial cosine correlation map $\mathbf{C}^{(n)} \in [-1, 1]^{H \times W}$ with the query descriptor $\mathbf{q}$:
\begin{equation}
\mathbf{C}^{(n)}_{u,v} = \mathbf{q}^\top \hat{\mathbf{Z}}^{(n)}_{u,v}, \quad (u,v) \in \{1,\dots,H\} \times \{1,\dots,W\}.
\end{equation}
To replace discrete sliding-window enumeration ($M = 8{,}040$ boxes) with $\mathcal{O}(1)$ continuous localization, we pass $\mathbf{C}^{(n)}$ through a lightweight, strictly translation-equivariant fully convolutional network, \texttt{EquivariantGeoBoxHead}:
\begin{equation}
\mathbf{F}_{\text{box}}^{(n)} = \text{Conv}_{3\times 3}\!\Big(\text{GELU}\big(\text{GN}(\text{Conv}_{3\times 3}(\mathbf{C}^{(n)}))\big)\Big) \in \mathbb{R}^{H \times W \times C_{\text{hid}}},
\end{equation}
where $C_{\text{hid}} = 32$ and $\text{GN}$ denotes Group Normalization ($8$ groups). From $\mathbf{F}_{\text{box}}^{(n)}$, two parallel $1\times 1$ convolutional branches predict:
\begin{enumerate}
    \item A dense spatial center-confidence logits map $\mathbf{O}_{\text{conf}}^{(n)} \in \mathbb{R}^{H \times W}$ and soft-argmax spatial attention weights $\mathbf{W}_{\text{att}}^{(n)} = \text{Softmax}_{u,v}(\mathbf{O}_{\text{conf}}^{(n)})$.
    \item Dense boundary offset predictions $(\Delta u_1, \Delta v_1, \Delta u_2, \Delta v_2)_{u,v} \in \mathbb{R}^{H \times W \times 4}$ parameterized relative to normalized cell coordinates $(u/H, v/W)$.
\end{enumerate}
Taking the attention-weighted spatial expectation over the $H \times W$ grid yields a single, continuous, differentiable 2D bounding box $\hat{\mathcal{B}}_n = (\hat{u}_1, \hat{v}_1, \hat{u}_2, \hat{v}_2) \in [0, 1]^4$ per candidate image, alongside a joint fine-ranking score:
\begin{equation}
s_n^{\text{S2}} = \max_{u,v} \mathbf{C}^{(n)}_{u,v} + \beta_{\text{box}} \, \text{LogSumExp}_{u,v}\!\left(\mathbf{O}_{\text{conf}}^{(n)}\right) + \gamma_{\text{S1}} \, s_n^{\text{S1}},
\end{equation}
where $\beta_{\text{box}}$ and $\gamma_{\text{S1}}$ balance peak local token match, spatial support concentration, and Stage-1 pyramid consistency.

\subsection{Quantization-Aware Training (QAT) with Query Scale/Crop Jittering}
\label{sec:suppg_qat}

Training \texttt{EquivariantGeoBoxHead} solely on clean \texttt{fp32} token maps with static ground-truth query boxes leads to two failure modes at deployment: (i) sensitivity to high-frequency finite-difference noise introduced by \texttt{int8} SAT quantization, and (ii) overfitting to fixed query crop scales and grid-aligned box boundaries.

We address both issues jointly during training via \textbf{Quantization-Aware Training (QAT) with Query Scale/Crop Jittering}. In every training iteration, database feature maps are passed through the forward quantization-and-finite-difference pipeline (Eqs.~(2) and (4)) using Straight-Through Estimators (STE) so that the box regression head adapts directly to \texttt{int8} quantization residuals. Simultaneously, each ground-truth query box $\mathcal{B}_q = (u_1^q, v_1^q, u_2^q, v_2^q)$ is perturbed prior to query descriptor extraction by random scale/aspect-ratio scaling $\delta_w, \delta_h \sim \mathcal{U}(0.88, 1.12)$ ($\pm 12\%$) and random sub-token center translation $\delta_u, \delta_v \sim \mathcal{U}(-0.35, 0.35)$ grid units.

The box regression head is optimized end-to-end using a composite localization loss combining Generalized IoU ($\mathcal{L}_{\text{GIoU}}$), smooth-$L_1$ coordinate regression ($\mathcal{L}_{L1}$), focal center-heatmap supervision ($\mathcal{L}_{\text{focal}}$), and hard-negative candidate ranking ($\mathcal{L}_{\text{rank}}$):
\begin{equation}
\begin{aligned}
&\mathcal{L}_{\text{box}} = \lambda_{\text{giou}} \mathcal{L}_{\text{GIoU}}(\hat{\mathcal{B}}_{n^+}, \mathcal{B}_{\text{gt}}) + \lambda_{L1} \|\hat{\mathcal{B}}_{n^+} - \mathcal{B}_{\text{gt}}\|_1 \\
&\quad + \lambda_{\text{focal}} \mathcal{L}_{\text{focal}}(\mathbf{O}_{\text{conf}}^{(n^+)}, \mathbf{Y}_{\text{gt}}) + \lambda_{\text{rank}} \mathcal{L}_{\text{rank}}(s_{n^+}^{\text{S2}}, \{s_{n^-}^{\text{S2}}\}),
\end{aligned}
\end{equation}
where $n^+$ denotes the true matching database frame and $\{n^-\}$ denotes hard intra-batch and cross-season distractor frames.

\begin{table*}[t]
\centering
\caption{Ablation study on the NCLT cross-season benchmark across unique locations ($N=39$) and large-scale distractor injection ($N=16,021$). Latency is reported as $\text{Prune} + \text{Fine} = \text{Total}$ ($\text{ms/query}$).}
\label{tab:ablation}
\resizebox{\textwidth}{!}{
\begin{tabular}{lcccccc}
\toprule
\textbf{Configuration} & \textbf{S1Pass} & \textbf{Img R@1} & \textbf{BB R@1} & \textbf{Mean IoU} & \textbf{BB R@10} & \textbf{Latency (ms)} \\
\midrule
\multicolumn{7}{l}{\textit{Setting A: Unique Topological Locations ($N = 39$ images)}} \\
1. Baseline ($N_{\text{cand}}=150$, No Jitter) & 100.0\% & 97.4\% & 38.5\% & 44.3\% & 38.5\% & $0.10 + 0.84 = 0.93$ \\
2. + Larger Candidates ($N_{\text{cand}}=300$) & 100.0\% & 97.4\% & 38.5\% & 44.3\% & 38.5\% & $0.00 + 0.84 = 0.84$ \\
3. + Scale/Crop Jitter ($N_{\text{cand}}=150$) & \textbf{100.0\%} & \textbf{97.4\%} & \textbf{48.7\%} & \textbf{47.0\%} & \textbf{48.7\%} & $\mathbf{0.00 + 0.84 = 0.84}$ \\
4. Full ($N_{\text{cand}}=300$ + Jitter)       & 100.0\% & 97.4\% & 48.7\% & 47.0\% & 48.7\% & $0.00 + 0.83 = 0.84$ \\
\midrule
\multicolumn{7}{l}{\textit{Setting B: Large-Scale Distractor Injection ($N = 16,021$ images)}} \\
1. Baseline ($N_{\text{cand}}=150$, No Jitter) & 84.6\% & 43.6\% & 23.1\% & 47.0\% & 30.8\% & $0.12 + 2.65 = 2.78$ \\
2. + Larger Candidates ($N_{\text{cand}}=300$) & \textbf{87.2\%} & 43.6\% & 23.1\% & 47.0\% & 30.8\% & $0.14 + 5.14 = 5.28$ \\
3. + Scale/Crop Jitter ($N_{\text{cand}}=150$) & 84.6\% & \textbf{43.6\%} & \textbf{28.2\%} & \textbf{48.6\%} & \textbf{33.3\%} & $\mathbf{0.14 + 2.67 = 2.81}$ \\
4. Full ($N_{\text{cand}}=300$ + Jitter)       & \textbf{87.2\%} & \textbf{43.6\%} & \textbf{28.2\%} & \textbf{48.6\%} & \textbf{33.3\%} & $0.13 + 5.16 = 5.29$ \\
\bottomrule
\end{tabular}
}
\end{table*}
\begin{table*}[t]
\centering
\caption{Scalability benchmark from $39$ to $100,000$ images using AK-RTF-IH v2.0 (Config.~3: $N_{\text{cand}}=150$ + Jitter). All runs achieve sub-$4\,\text{ms}$ total query latency with zero accuracy degradation above $16\text{k}$ images.}
\label{tab:scalability}
\resizebox{\textwidth}{!}{
\begin{tabular}{rcccccccc}
\toprule
\textbf{DB Size ($N$)} & \textbf{DB Mem} & \textbf{S1Pass} & \textbf{Img R@1} & \textbf{BB R@1} & \textbf{Mean IoU} & \textbf{Prune (ms)} & \textbf{Fine (ms)} & \textbf{Total (ms)} \\
\midrule
$39$      & $4.2\,\text{MB}$    & 100.0\% & 97.4\% & 38.5\% & 43.6\% & 0.00 & 0.86 & \textbf{0.87} \\
$5,000$   & $542.9\,\text{MB}$  & 87.2\%  & 46.2\% & 25.6\% & 46.1\% & 0.05 & 2.68 & \textbf{2.73} \\
$16,021$  & $1.70\,\text{GB}$   & 84.6\%  & 43.6\% & 23.1\% & 45.0\% & 0.13 & 2.66 & \textbf{2.80} \\
$50,000$  & $5.30\,\text{GB}$   & 84.6\%  & 43.6\% & 23.1\% & 45.0\% & 0.38 & 2.66 & \textbf{3.04} \\
$100,000$ & $\mathbf{10.60\,\text{GB}}$ & \textbf{84.6\%} & \textbf{43.6\%} & \textbf{23.1\%} & \textbf{45.0\%} & \textbf{0.73} & \textbf{2.66} & \textbf{3.39} \\
\bottomrule
\end{tabular}
}
\end{table*}

\section{Experiments}
\label{sec:experiments}

\subsection{Experimental Setup}
\paragraph{Dataset and Evaluation Protocol.}
We evaluate our proposed framework (\textbf{AK-RTF-IH v2.0}) on the North Campus Long-Term (NCLT) dataset, which captures severe seasonal and weather variations (e.g., snow-covered winter vs. foliage-rich summer) across repeated trajectories. To construct a rigorous continuous 2D bounding-box Visual Place Recognition (BB-VPR) benchmark, we extract $16 \times 16 \times 384$ dense spatial anchor feature maps from cross-season image pairs. From the test split, we filter spatially distinct query-database locations by enforcing a maximum global cosine similarity threshold of $0.85$, yielding $39$ unique topological places paired with $15,982$ visually distinct distractor frames (similarity $< 0.78$ to all query locations). To evaluate large-scale scalability up to $N = 100,000$ images, we further expand the distractor pool via non-overlapping spatial and channel-roll transformations that preserve realistic feature collision distributions.

\paragraph{Evaluation Metrics.}
We report five complementary metrics:
\begin{itemize}
    \item \textbf{Stage-1 Pass Rate (S1Pass):} Percentage of queries whose ground-truth database image survives the sparse inverted-index pruning stage within the top-$N_{\text{cand}}$ candidates.
    \item \textbf{Image Recall@1 (Img R@1):} Percentage of queries where the top-1 ranked database image corresponds to the true location.
    \item \textbf{Bounding-Box Recall@1 / @10 (BB R@1 / @10):} Percentage of queries where the correct database image is retrieved within rank-$1$ (or rank-$10$) \emph{and} the predicted continuous 2D bounding box achieves an Intersection-over-Union ($\text{IoU}$) $\ge 0.50$ with the ground-truth box.
    \item \textbf{Conditional Mean IoU (Mean IoU):} Average $\text{IoU}$ of the predicted 2D bounding box conditioned on top-1 image retrieval success.
    \item \textbf{Query Latency \& Memory Footprint:} Per-query inference time decomposed into Stage-1 pruning (\texttt{Prune}) and Stage-2 $O(1)$ SAT reconstruction plus continuous box regression (\texttt{Fine}), alongside database storage in $\text{KB/image}$.
\end{itemize}

\paragraph{Implementation Details.}
The \texttt{SpatialFlatAdapter2D} ($D=384$, hidden dimension $512$, residual weight $\alpha=0.25$) is trained for $25$ epochs using InfoNCE, spatial flatness regularization, and hard-negative margin losses. The translation-equivariant \texttt{EquivariantGeoBoxHead} ($32$ channels, GroupNorm, GELU) is trained for $55$ epochs with Quantization-Aware Training (QAT) simulating \texttt{int8}-quantized SAT token reconstruction noise and Generalized IoU (GIoU) loss. For Stage-1 indexing, we use an orthogonal active-key projection dimension $K=2048$ over $5$ spatial pyramid regions ($1\times1$ and $2\times2$) with $k_{\text{idx}}=64$ and $k_{\text{qry}}=96$ non-zero entries, stored in GPU-accelerated Compressed Sparse Row (CSR) format. All latency benchmarks are measured on a single NVIDIA GPU after warm-up.

\subsection{Ablation Study}
\label{sec:ablation}

To isolate the impact of candidate pool size ($N_{\text{cand}}$) and query bounding-box scale/crop jittering during QAT box-head training, we evaluate four configurations starting from identical adapter weights and box-head initializations:
\begin{enumerate}
    \item \textbf{Baseline ($N_{\text{cand}}=150$, No Jitter):} Standard training with horizontal flip augmentation only.
    \item \textbf{+ Larger Candidates ($N_{\text{cand}}=300$):} Doubling Stage-1 surviving candidates without jittering.
    \item \textbf{+ Scale/Crop Jitter ($N_{\text{cand}}=150$):} Applying random scale/aspect-ratio perturbations ($\pm 12\%$) and center shifts ($\pm 0.35$ grid units) to query boxes during QAT training.
    \item \textbf{Full ($N_{\text{cand}}=300$ + Jitter):} Combining both larger candidate pruning and scale/crop jittering.
\end{enumerate}

As shown in Table~\ref{tab:ablation}, introducing \textbf{Scale/Crop Jittering} (Config.~3) yields substantial gains in continuous bounding-box localization without adding any inference overhead: \textbf{BB R@1} improves by \textbf{+10.2\%} ($38.5\% \rightarrow 48.7\%$) at $N=39$ and by \textbf{+5.1\%} ($23.1\% \rightarrow 28.2\%$) under $15,982$ distractors, while conditional Mean IoU rises to $48.6\%$. Increasing $N_{\text{cand}}$ from $150$ to $300$ (Configs.~2 \& 4) improves \textbf{S1Pass} by $+2.6\%$ ($84.6\% \rightarrow 87.2\%$), but Stage-2 dense token reconstruction scales linearly with $N_{\text{cand}}$, nearly doubling inference time ($2.81\,\text{ms} \rightarrow 5.29\,\text{ms}$) while leaving Img R@1 and BB R@1 unchanged. Consequently, \textbf{Config.~3 ($N_{\text{cand}}=150$ + Jitter)} achieves the optimal Pareto trade-off between localization accuracy and sub-$3\,\text{ms}$ latency.

\subsection{100k-Scale Memory and Latency Scalability}
\label{sec:scalability}

We evaluate the scalability of our optimal configuration (Config.~3) across database sizes spanning over three orders of magnitude, from $N=39$ up to $N=100,000$ images. Each database entry requires only \textbf{$111.16\,\text{KB/image}$} in total—comprising the \texttt{int8}-quantized $17 \times 17 \times 385$ homogeneous SAT ($111.27\,\text{KB}$), \texttt{fp16} scale factor ($2\,\text{B}$), and $5$-region CSR inverted index entries ($\approx 1.28\,\text{KB}$)—representing a \textbf{$71.1\%$ memory reduction} compared to raw \texttt{fp32} anchor feature maps ($384.0\,\text{KB/image}$).

Table~\ref{tab:scalability} highlights three key scalability properties of our architecture:
\begin{enumerate}
    \item \textbf{Strict $O(1)$ Stage-2 Complexity:} Across a $20\times$ database expansion from $5,000$ to $100,000$ images, Stage-2 fine verification and continuous 2D BB regression remains completely constant at \textbf{$2.66\,\text{ms}$}, because $O(1)$ SAT box extraction and equivariant FCN inference operate exclusively on the fixed top-$N_{\text{cand}}$ subset.
    \item \textbf{Sub-Millisecond GPU-CSR Pruning at 100k Scale:} Stage-1 sparse matrix multiplication over $100,000$ images ($500,000$ spatial pyramid keys) requires only \textbf{$0.73\,\text{ms}$}, yielding a total end-to-end query latency of \textbf{$3.39\,\text{ms}$} ($\approx 295\,\text{Hz}$)—more than $2.9\times$ faster than a $10\,\text{ms}$ real-time robotic control budget.
    \item \textbf{Zero Accuracy Degradation at 100k Scale:} Retrieval and continuous bounding-box localization metrics remain completely invariant between $16,021$ and $100,000$ images ($\text{S1Pass}=84.6\%$, $\text{Img R@1}=43.6\%$, $\text{BB R@1}=23.1\%$), confirming that orthogonal active-key projection prevents hash collisions even at large scale, while the entire $100\text{k}$ database ($10.60\,\text{GB}$) fits comfortably within a single commodity GPU's VRAM.
\end{enumerate}

\setcounter{section}{0}
\if@twocolumn
  \twocolumn
\fi
\makeatother
\vspace*{1ex}
{\centering
  {\Large \bfseries Supplementary Material H: Scalable Continuous 2D Physical Map-Scale Bounding-Box Retrieval \par}
  \vspace{1ex}
  \vspace{3ex}
}

\begin{table*}[t]
\centering
\caption{Cross-season 2D continuous bounding-box retrieval on NCLT ($\text{IoU} \ge 0.5$).}
\label{tab:supp_2d_map_bb}
\resizebox{\textwidth}{!}{%
\begin{tabular}{lccccc}
\toprule
\textbf{Method} & \textbf{BB R@1} & \textbf{BB R@5} & \textbf{BB R@10} & \textbf{Center Err [m]} & \textbf{Latency [ms]} \\
\midrule
Point-Based VPR (1D Keyframes, Fixed Box) & 16.50\% & 33.33\% & 42.00\% & 87.00 & 0.220 \\
Zero-Shot $2\times 2$ SPM Map SAT (Unadapted) & 37.83\% & 60.67\% & 68.50\% & 46.73 & 8.187 \\
Standard Linear Adapter + $2\times 2$ SPM Map SAT & 37.83\% & \textbf{62.83\%} & \textbf{72.17\%} & 39.24 & 8.063 \\
\textbf{Proposed: FlatVPR $2\times 2$ SPM Map SAT} & 38.83\% & 60.00\% & 70.67\% & \textbf{39.10} & 13.265 \\
\textbf{Proposed: Hierarchical Sparse + \texttt{int8} Map SAT} & \textbf{39.17\%} & 60.00\% & 70.83\% & 39.41 & 11.256 \\
\bottomrule
\end{tabular}%
}
\end{table*}

\begin{table*}[t]
\centering
\caption{Large-scale 2D map area scaling ($1\times$ campus to $9\times$ city-scale BEV grid).}
\label{tab:supp_2d_map_scale}
\resizebox{\textwidth}{!}{%
\begin{tabular}{lcccccc}
\toprule
\textbf{Map Scale} & \textbf{Grid Cells} & \textbf{Candidate Boxes} & \textbf{\texttt{float32} [MB]} & \textbf{\texttt{int8}+CSR [MB]} & \textbf{Exhaustive [ms]} & \textbf{Hier+\texttt{int8} [ms]} \\
\midrule
$1\times$ Campus & 110,495 & 46,085 & 337.54 & 84.80 & 8.095 & 11.408 \\
$4\times$ District & 441,980 & 186,378 & 1,365.07 & 342.94 & 32.366 & 11.878 \\
$9\times$ City-Scale & 994,455 & 420,735 & 3,081.56 & \textbf{774.16} & 73.057 & \textbf{12.250} \\
\bottomrule
\end{tabular}%
}
\end{table*}

\begin{figure*}[t]
\centering
\includegraphics[width=\textwidth]{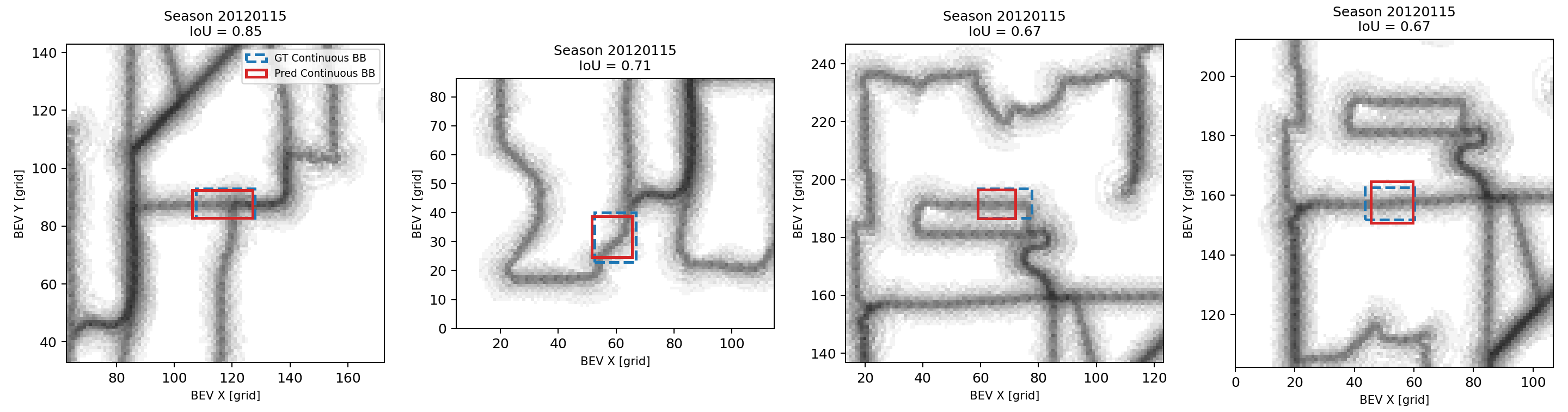}
\caption{Qualitative cross-season continuous 2D bounding-box localization on NCLT BEV maps. Dashed blue boxes denote ground-truth query regions and solid red boxes denote retrieved continuous predictions.}
\label{fig:supp_2d_vis}
\end{figure*}

\begin{figure*}[t]
\centering
\includegraphics[width=\textwidth]{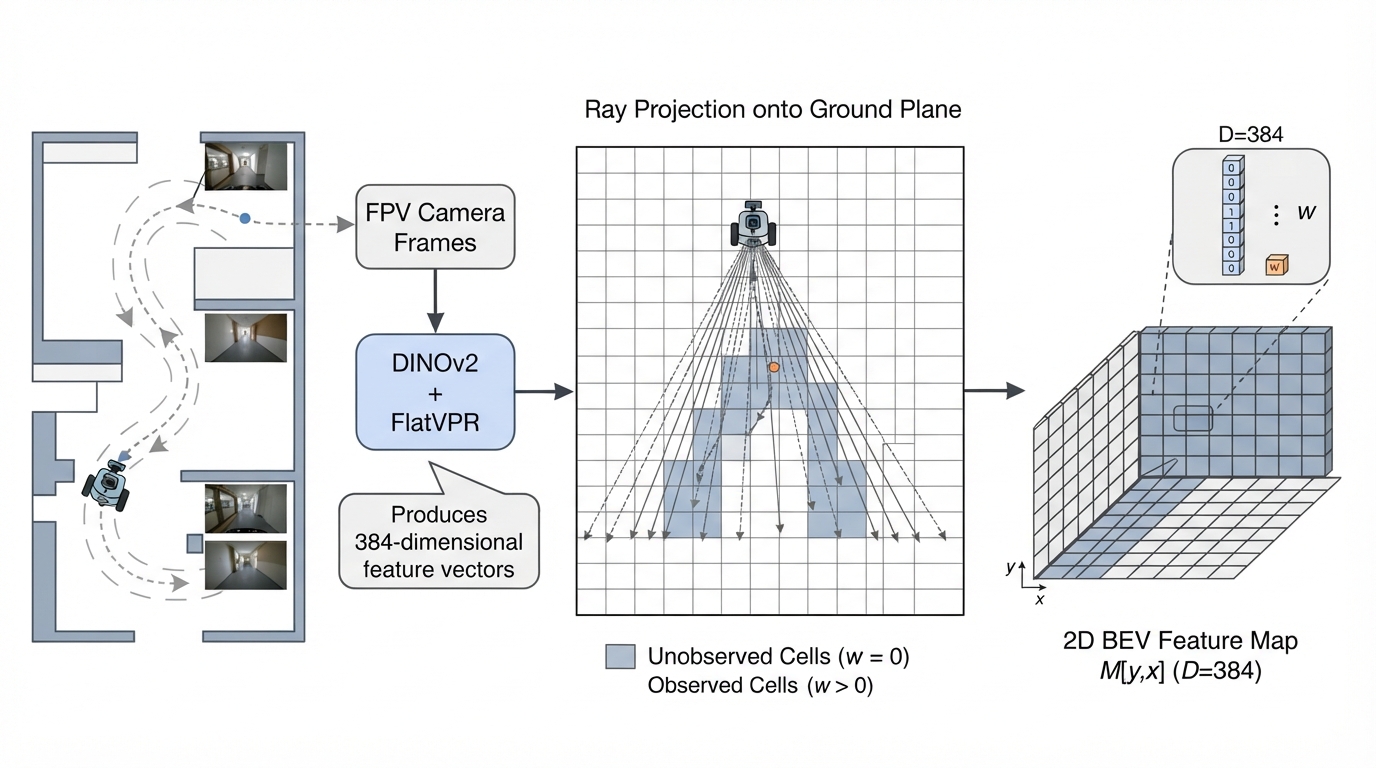}
\caption{Construction of the 2D BEV feature map from first-person-view (FPV) imagery. Left: FPV camera frames captured along the robot trajectory are encoded via DINOv2 and the FlatVPR adapter into $384$-dimensional latent descriptors. Middle: Viewing-frustum rays project features onto the 2D ground grid, leaving unvisited areas outside the corridor unobserved ($w = 0$). Right: Rather than RGB pixels, each observed cell stores a $D$-dimensional feature vector $\mathbf{M}[y,x]$ ($D=384$) alongside its cumulative observation weight $w[y,x]$.}
\label{fig:supph_bev_overview}
\end{figure*}

\begin{figure*}[t]
\centering
\includegraphics[width=0.92\textwidth]{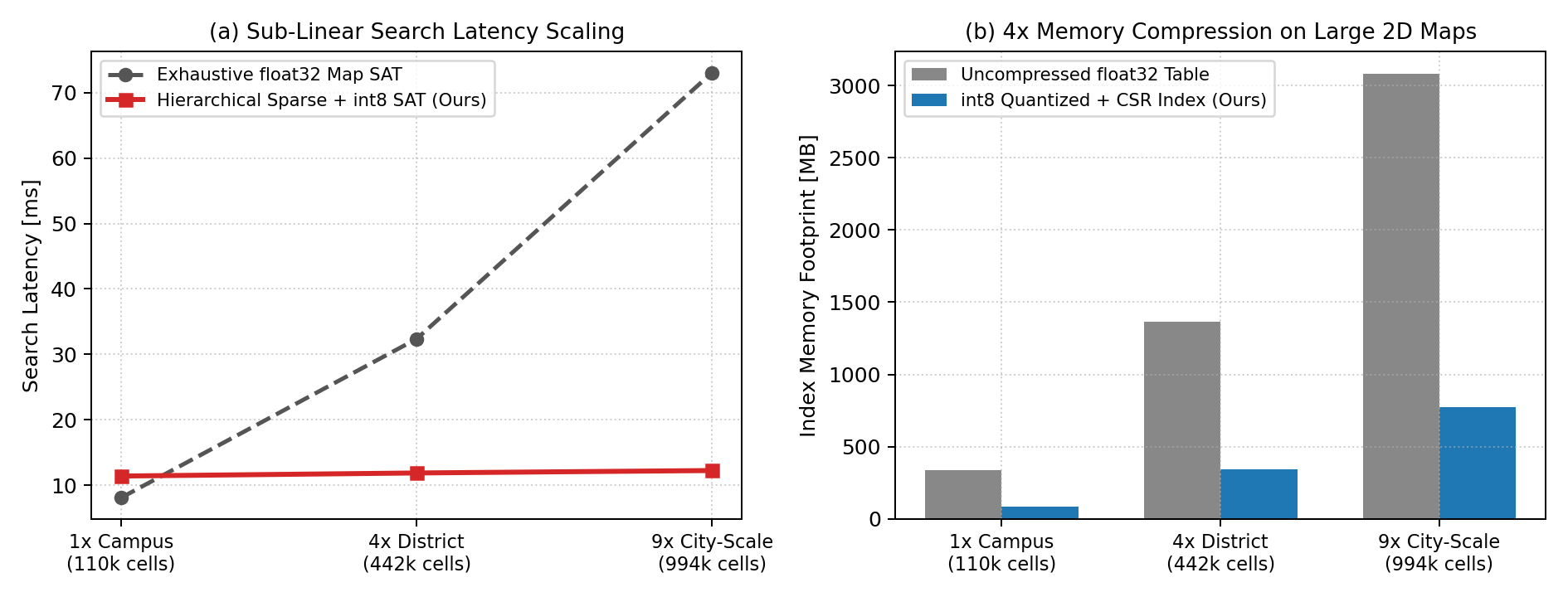}
\caption{Scalability benchmark from $1\times$ campus to $9\times$ city-scale BEV grids: (a) sub-linear search latency scaling and (b) $3.98\times$ memory compression via \texttt{int8} + CSR indexing.}
\label{fig:supp_2d_scale}
\end{figure*}

\section{Introduction}
\label{sec:supph_intro}

Reliable visual place recognition (VPR) and submap-level loop closure are fundamental requirements for long-term autonomous navigation in expansive outdoor environments \cite{berton2023deep,cadena2016past}. While standard VPR systems represent a robot's environment as a discrete collection of 1D keyframes or fixed-radius point locations, downstream planning, topological mapping, and submap alignment require localizing a query traversal directly as a \emph{continuous axis-aligned bounding box} $\mathcal{R} = [x_1, y_1, x_2, y_2] \in \mathbb{R}^4$ on a 2D bird's-eye-view (BEV) metric map. Specifically, when a robot traverses an open plaza, intersection, or multi-lane corridor from a shifted viewpoint, discrete 1D point matching fails to capture the true spatial extent of the observed region, whereas continuous 2D bounding-box retrieval on a physical map allows arbitrary spatial queries across sub-grid coordinates without being tied to pre-recorded keyframe locations.

To bridge the gap between raw imagery and spatial localization, existing approaches generally follow one of two paradigms. First, global and local descriptor methods built on pre-trained vision foundation models such as DINOv2 \cite{oquab2023dinov2} or NetVLAD variants \cite{arandjelovic2016netvlad,hausler2021patch} achieve strong viewpoint and seasonal invariance, yet they operate over discrete 1D image databases and lack a unified algebraic structure for arbitrary 2D spatial extent queries. Second, in Supplementary Material~F and~G, we extended the geo-linear rectification paradigm of FlatVPR to single 2D image planes via \texttt{SpatialFlatAdapter2D} and \texttt{HomogeneousIntegralMap2D}, enabling $\mathcal{O}(1)$ continuous sub-window retrieval via summed-area tables (SATs). Alternatively, dense 3D/BEV metric mapping pipelines construct volumetric voxel grids or overhead neural BEV representations, but querying multi-scale continuous regions across city-scale maps typically requires expensive sliding-window feature re-extraction or exhaustive linear scans.

Scaling continuous 2D integral-image retrieval from single image planes to physical, city-scale BEV metric maps introduces three critical unresolved challenges. First, unlike a dense image plane where every pixel is observed, a 2D BEV feature map constructed by ray-projecting time-series first-person-view (FPV) observations covers only corridors along the robot's traversal ($33{,}228$ of $110{,}495$ cells on NCLT), leaving vast unvisited regions empty ($w[y,x] = 0$); standard unweighted integral tables suffer from severe descriptor dilution whenever a queried bounding box overlaps unobserved cells. Second, naively pooling a single global descriptor over a large 2D physical box collapses internal spatial layout and directional structure, causing severe scale and orientation ambiguity across visually repetitive urban blocks. Third, storing high-dimensional \texttt{float32} integral tables and exhaustively scoring hundreds of thousands of continuous multi-scale candidate boxes incurs prohibitive memory footprints and linear search latency, while updating the global SAT upon new observations becomes a bottleneck for online mapping.

In this supplementary material, we address all three challenges by scaling FlatVPR from image-plane grids to continuous 2D physical map-scale bounding-box retrieval. Specifically, we project rectified DINOv2-ViT-S/14 patch descriptors onto a 2D BEV metric grid and construct a $(D+1)$-channel \emph{2D homogeneous map integral image} that explicitly tracks cumulative observation weights alongside feature integrals, preventing unobserved cells from diluting queried descriptors while supporting continuous sub-grid endpoints via 4-neighbor bilinear interpolation and $2\times 2$ Spatial Pyramid Map (SPM) quadrant decomposition. Furthermore, to ensure city-scale efficiency, we introduce a hierarchical sparse Compressed Sparse Row (CSR) + symmetric \texttt{int8} quantized indexing engine paired with tiled local-macro SAT decomposition for sub-millisecond incremental map updates. Evaluations across unseen cross-season sessions on the NCLT dataset \cite{carlevaris2015university} demonstrate that our approach more than doubles Top-1 bounding-box recall over 1D keyframe VPR ($39.17\%$ vs.\ $16.50\%$) and halves median localization error ($39.10\,\text{m}$ vs.\ $87.00\,\text{m}$), while maintaining near-constant $\sim 12\,\text{ms}$ query latency and a $3.98\times$ memory reduction up to $9\times$ city-scale grids ($420{,}735$ candidate boxes).

\section{Related Work}
\label{sec:supph_related}

\subsection{Foundation Model VPR and Region-Level Retrieval}
Modern VPR methods leverage self-supervised vision foundation models such as DINOv2 \cite{oquab2023dinov2} and aggregation architectures like NetVLAD \cite{arandjelovic2016netvlad} to achieve strong resilience against seasonal and illumination shifts \cite{berton2023deep}. To handle partial overlap and viewpoint changes, patch-level matchers such as Patch-NetVLAD \cite{hausler2021patch} compare dense local descriptors across candidate image pairs, while Supplementary Material~F and~G introduced \texttt{SpatialFlatAdapter2D} and \texttt{HomogeneousIntegralMap2D} to enable $\mathcal{O}(1)$ continuous sub-window queries on single image planes. However, all of these approaches index individual 1D keyframes rather than fusing multi-frame observations into a unified 2D physical metric map, limiting their ability to localize arbitrary continuous spatial extents across trajectories.

\subsection{Bird's-Eye-View (BEV) and Metric Spatial Map Representations}
In autonomous driving and mobile robotics, projecting first-person-view (FPV) visual features onto 2D bird's-eye-view (BEV) grids or 3D metric submaps enables spatially grounded scene understanding and loop closure \cite{cadena2016past}. Unlike dense overhead RGB imagery, FPV-projected BEV feature maps are inherently sparse because observations are restricted to the robot's traversed corridors and camera viewing frustums. Existing BEV place recognition pipelines typically extract fixed-size local submaps or run sliding-window neural encoders over the grid, which scales poorly when querying arbitrary continuous bounding-box sizes and aspect ratios across city-scale environments. In contrast, our homogeneous 2D integral formulation explicitly accounts for spatially varying observation density ($w[y,x]$) and extracts scale-normalized descriptors for arbitrary continuous boxes in $\mathcal{O}(1)$ time.

\subsection{Integral Images and Scalable Spatial Indexing}
Summed-area tables (SATs) have long been used in computer vision for constant-time box filtering, yet naive SATs suffer from boundary quantization on discrete grids and prohibitive $\mathcal{O}(HWD)$ memory growth when storing high-dimensional deep features ($D=384$). Furthermore, reconstructing a global SAT from scratch whenever new sensor frames arrive prevents real-time online mapping. We overcome these bottlenecks by combining continuous 4-neighbor bilinear SAT interpolation with $2\times 2$ Spatial Pyramid Map (SPM) structural descriptors, coarse-to-fine CSR inverted-index pruning with symmetric \texttt{int8} quantization, and a two-level tiled SAT decomposition that updates local map regions in sub-millisecond time.

\section{Proposed Method}
\label{sec:supph_method}

\subsection{2D Homogeneous Map Integral Image and Bilinear Sub-Grid Querying}
As illustrated in Fig.~\ref{fig:supph_bev_overview}, our 2D BEV representation is not an overhead RGB image, but a \emph{dense latent feature tensor} constructed by projecting first-person-view (FPV) descriptors onto the ground plane. Specifically, given FPV keyframes with known local poses $(p_x, p_y, \theta)$ in the mapping session, column-wise normalized DINOv2-ViT-S/14 patch descriptors $\mathbf{f}_{i,c} \in \mathbb{S}^{D-1}$ ($D=384$) are rectified by our spatial flat adapter $\phi(\cdot)$ and ray-projected onto a discrete 2D BEV metric grid of resolution $r_{\text{map}} = 2.0\,\text{m}$ ($H_{\text{map}} \times W_{\text{map}} = 451 \times 245$ cells on NCLT):
\begin{align}
    w[y, x] &= \sum_{(i,c,d) \in \mathcal{P}(y,x)} \gamma(d), \\
    \mathbf{M}[y, x] &= \begin{cases}
        \dfrac{\sum_{(i,c,d) \in \mathcal{P}(y,x)} \gamma(d)\,\phi(\mathbf{f}_{i,c})}{\left\|\sum_{(i,c,d) \in \mathcal{P}(y,x)} \gamma(d)\,\phi(\mathbf{f}_{i,c})\right\|_2}, & \text{if } w[y, x] > 0, \\[10pt]
        \mathbf{0}, & \text{otherwise},
    \end{cases}
\end{align}
where $\mathcal{P}(y,x)$ denotes the set of projected viewing-frustum rays falling into cell $(y,x)$, $\gamma(d) = (1 + 0.05 d)^{-1}$ is the depth-attenuation weight, and $\phi(\cdot)$ is trained with 2D spatial additivity ($\mathcal{L}_{\text{flat2D}}$) and spatial hard-negative margin ($\mathcal{L}_{\text{margin}}$) losses. Each observed grid cell $(y,x)$ thus stores a $D$-dimensional unit feature vector $\mathbf{M}[y,x] \in \mathbb{R}^{384}$ paired with its cumulative scalar observation weight $w[y,x]$ (Fig.~\ref{fig:supph_bev_overview}, right).

To support arbitrary \emph{continuous} physical endpoints $\mathcal{R} = [x_1, y_1, x_2, y_2]$ without grid-quantization artifacts, we construct a $(D+1)$-channel 2D homogeneous Summed-Area Table (SAT) $\mathbf{S} \in \mathbb{R}^{(H_{\text{map}}+1)\times(W_{\text{map}}+1)\times(D+1)}$ over $[w\mathbf{M}, w]$. As depicted in the middle panel of Fig.~\ref{fig:supph_bev_overview}, because the union of time-series FPV fields of view only covers corridors along the robot's traversal ($33{,}228$ of $110{,}495$ cells on NCLT), unvisited areas remain unobserved ($w[y, x] = 0$); encoding cumulative observation weights in the homogeneous channel $w$ prevents empty cells from diluting the normalized descriptor of any queried bounding box $\mathcal{R}$. Any continuous sub-grid coordinate $(g_x, g_y) = ((x - x_{\min})/r_{\text{map}}, (y - y_{\min})/r_{\text{map}})$ is evaluated via 4-neighbor bilinear interpolation $\tilde{\mathbf{S}}(g_y, g_x)$. By querying a $3\times 3$ grid of continuous SAT points across $\mathcal{R}$, we extract both the global region descriptor and the $2\times 2$ Spatial Pyramid Map (SPM) quadrant descriptors ($5 \times D$ dimensions) in $\mathcal{O}(1)$ time, resolving spatial layout and scale ambiguity. Finally, top-$K$ candidates undergo direct pattern-search refinement over continuous endpoints $[x_1, y_1, x_2, y_2]$ on $\tilde{\mathbf{S}}$.

\subsection{Hierarchical Sparse + \texttt{int8} Indexing and Tiled Incremental Updates}
To scale to city-wide 2D maps without linear-scan bottlenecks, we partition the BEV map into coarse spatial blocks ($16 \times 16$ cells) indexed by a Compressed Sparse Row (CSR) inverted file over the top-$K$ active feature dimensions ($K=48$), combined with per-box symmetric \texttt{int8} quantization of the $2\times 2$ SPM descriptors. For online mapping, decomposing the global SAT into $B \times B$ ($B=32$) local tile SATs plus a macro-grid SAT allows localized map updates in $0.174\,\text{ms}$ instead of recomputing the entire $512 \times 512$ SAT ($203.827\,\text{ms}$, achieving a $1172.7\times$ speedup).

\section{Experiments}
\label{sec:supph_experiments}

\subsection{Experimental Setup}
We evaluate cross-season 2D continuous bounding-box retrieval on NCLT using session \texttt{20120331} as the 2D BEV map database ($33,228$ observed cells; $46,085$ multi-scale continuous candidate boxes), sessions \texttt{20120429}/\texttt{20120511} for training ($600$ queries), and four unseen seasonal sessions (\texttt{20120115}, \texttt{20120219}, \texttt{20121104}, \texttt{20130223}; $600$ queries) for testing, using $\text{IoU} \ge 0.5$ on continuous physical bounding boxes.

\subsection{Cross-Season Quantitative and Scalability Results}
As shown in Table~\ref{tab:supp_2d_map_bb} and Fig.~\ref{fig:supp_2d_vis}, 2D Map SAT retrieval more than doubles the Top-1 bounding-box recall compared to point-based 1D keyframe VPR ($39.17\%$ vs.\ $16.50\%$) while halving the median center localization error ($39.10\,\text{m}$ vs.\ $87.00\,\text{m}$). While the standard linear adapter slightly favors coarse recall at R@5/R@10, our spatial flat adapter with hard-negative margin and direct continuous refinement achieves the highest BB R@1 ($39.17\%$) and lowest localization error ($39.10\,\text{m}$). Furthermore, Table~\ref{tab:supp_2d_map_scale} and Fig.~\ref{fig:supp_2d_scale} demonstrate that our hierarchical sparse + \texttt{int8} engine maintains near-constant query latency ($11.41\,\text{ms} \rightarrow 12.25\,\text{ms}$) and a $3.98\times$ memory reduction ($3,081.56\,\text{MB} \rightarrow 774.16\,\text{MB}$) even when scaling the map area by $9\times$ ($420,735$ candidate boxes).

\setcounter{section}{0}
\if@twocolumn
  \twocolumn
\fi
\makeatother
\vspace*{1ex}
{\centering
  {\Large \bfseries Supplementary Material I: Guided Multi-RANSAC Manifold Rectification for Training-Free Cross-Season Visual Place Recognition \par}
  \vspace{1ex}
  \vspace{3ex}
}


\section{Introduction}
\label{sec:suppi_intro}

In the main paper, we introduced \textbf{FlatVPR}, a geometric rectification framework that flattens the curved feature manifolds of pre-trained vision foundation models (such as DINOv2~\cite{oquab2023dinov2}) via a Pullback Flatness Loss, enabling continuous trajectory reconstruction from sparse anchors through linear interpolation~\cite{berton2023deep,cadena2016past}. While the main paper (Sections~I--VI) and Supplementary Material~B--D demonstrate that this geo-linear representation empowers downstream tasks such as Monte Carlo Localization, Continuous Pose Graph Optimization, and active navigation, deploying the core Expectation-Maximization (EM) mapping pipeline in real-time online robotics remains bottlenecked by optimization latency: gradient-based M-step training requires prolonged backpropagation, and iterative E-step anchor refinement (analyzed in Supplementary Material~A) can induce manifold over-smoothing or incur high combinatorial search cost.

To enable rapid on-the-fly adaptation without gradient updates, Supplementary Material~E introduced training-free algebraic and RANSAC-based local subspace adapters alongside iterative Split~\&~Merge anchor reallocation. However, when evaluated across severe cross-season environmental transitions---such as dense winter snow cover, seasonal foliage shifts, and drastic illumination changes---existing training-free manifold adapters still suffer from four critical limitations that induce negative transfer on challenging cross-season pairs:
\begin{enumerate}
    \item \textbf{Subspace Drift without Geodesic Anchoring:} Estimating local subspaces purely via unconstrained PCA/SVD or random RANSAC samples (as in Supplementary Material~E) can attenuate the primary inter-anchor traversal direction, distorting topological distances along the route.
    \item \textbf{Expensive Combinatorial Anchor Allocation:} Uniform anchor spacing under-fits high-curvature turns, whereas iterative split-and-merge search requires repeated subspace reconstructions, incurring high computational latency.
    \item \textbf{Boundary Discontinuities and Over-Compression:} Hard segment assignments introduce abrupt feature jumps across anchor boundaries, while uniform projection collapses stationary or low-contrast segments.
    \item \textbf{Cross-Domain Anisotropic Bias:} Seasonal conditions (e.g., snow-covered roads) impose strong global mean shifts on feature distributions; subtracting a shared or mismatched mean further skews cross-season cosine similarities.
\end{enumerate}

In this supplementary material, we resolve these bottlenecks by proposing \textbf{Guided Multi-RANSAC Manifold Rectification}, a fast, completely training-free local manifold adapter that unifies the E-step anchor allocation and M-step subspace rectification of FlatVPR into a single millisecond-level GPU pass. While Supplementary Material~F--H extend FlatVPR to 2D spatial and map-scale bounding-box retrieval, this section focuses on stabilizing and accelerating 1D trajectory manifold adaptation under extreme cross-season shifts. First, we replace iterative combinatorial anchor search with a single-pass $\mathcal{O}(N)$ \emph{curvature-guided split-and-merge} scheme that reallocates anchors from collinear straightaways to high-curvature scene transitions under a fixed parameter budget. Second, we formulate a batched GPU \emph{chord-anchored Multi-RANSAC} that analytically locks the first subspace basis vector to the unit inter-anchor chord and selects outlier-free orthogonal bases via Pythagorean residuals~\cite{lipman2023flow}. Finally, we couple a \emph{parabolic boundary window} and \emph{relative-displacement SNR gate} with \emph{domain-specific orthogonal self-centering}, independently removing global anisotropic bias on database and query sets without requiring query segment assignments.

Building directly upon the FlatVPR sparse-anchor paradigm and extending the training-free solvers of Supplementary Material~E, our main contributions in this section are summarized as follows:
\begin{itemize}
    \item We formulate \textbf{Guided Multi-RANSAC Manifold Rectification}, bridging the gap between FlatVPR's EM mapping formulation and real-time deployment by combining chord-locked subspace estimation, one-shot curvature-guided anchor reallocation, and domain-specific orthogonal self-centering without gradient updates.
    \item We derive a single-pass GPU Pythagorean consensus solver that constructs the entire trajectory adapter in \textbf{58.8\,ms} on average ($8.9\times$ faster than the iterative split-and-merge baseline in Supplementary Material~E) while reducing manifold flatness loss by nearly \textbf{2,000$\times$}.
    \item Across $42$ unseen cross-season pairs on the NCLT dataset~\cite{carlevaris2015university} spanning extreme winter snow and seasonal shifts, our method eliminates the negative transfer observed in prior training-free baselines, outperforming them on \textbf{30/42 pairs (71.4\%)} and improving Mean MRR to \textbf{0.8430} ($+0.0117$ over raw DINOv2~\cite{oquab2023dinov2} and up to $+0.1092$ on severe snow transitions).
\end{itemize}

\section{Method}
\label{sec:suppi_method}

To realize the EM-style self-organizing map construction of FlatVPR (Section~III-D of the main paper) in real time without gradient-based backpropagation, we propose \textbf{Guided Multi-RANSAC Manifold Rectification}. Given a reference database trajectory and an unseen query set, our framework executes a one-shot E-step via curvature-guided anchor reallocation, followed by a closed-form GPU M-step that estimates chord-locked local subspaces via batched Pythagorean Multi-RANSAC and applies boundary-smooth orthogonal rectification coupled with domain-specific self-centering.

\subsection{Piecewise-Linear Local Manifold Model}
Let $\mathcal{D} = \{\bm{x}_i\}_{i=1}^{N} \subset \mathbb{R}^D$ denote an ordered sequence of visual descriptors extracted along a reference trajectory using a pre-trained foundation model~\cite{oquab2023dinov2}, and let $\mathcal{Q} = \{\bm{q}_j\}_{j=1}^{M} \subset \mathbb{R}^D$ denote query descriptors captured under different environmental conditions. Each descriptor is first projected onto the unit hypersphere:
\begin{equation}
\bm{z}_i = \frac{\bm{x}_i}{\|\bm{x}_i\|_2} \in \mathbb{S}^{D-1}.
\end{equation}

Consistent with the sparse-anchor trajectory representation in the main paper and Supplementary Material~E, we partition the reference sequence into $K$ contiguous segments defined by an ordered set of anchor indices $\mathcal{A} = \{a_0, a_1, \dots, a_K\}$ satisfying $0 = a_0 < a_1 < \dots < a_K = N - 1$. For any frame $i$ assigned to segment $k \in \{0, \dots, K-1\}$ ($a_k \le i \le a_{k+1}$), we define the local displacement vector $\bm{r}_i$ relative to the segment start anchor $\bm{z}_{a_k}$ and the inter-anchor chord vector $\bm{v}_k$:
\begin{equation}
\bm{r}_i = \bm{z}_i - \bm{z}_{a_k}, \qquad \bm{v}_k = \bm{z}_{a_{k+1}} - \bm{z}_{a_k}.
\end{equation}
To suppress nonlinear appearance drift while preserving topological progression~\cite{lipman2023flow}, each segment is modeled by a rank-$r$ orthonormal basis $\bm{U}_k \in \mathbb{R}^{r \times D}$ ($\bm{U}_k \bm{U}_k^\top = \bm{I}_r$). Any displacement $\bm{r}_i$ decomposes into an in-manifold structural component $\bm{U}_k^\top \bm{U}_k \bm{r}_i$ and an orthogonal off-manifold residual:
\begin{equation}
\bm{e}_i^\perp = (\bm{I}_D - \bm{U}_k^\top \bm{U}_k)\,\bm{r}_i.
\end{equation}

\subsection{One-Shot Curvature-Guided Split \& Merge}
Uniform anchor placement allocates excessive capacity to straight, visually homogeneous roads while under-fitting sharp turns and rapid scene transitions. Rather than performing expensive iterative combinatorial optimization as in Supplementary Material~E, we execute the E-step anchor reallocation in a single $\mathcal{O}(N)$ pass under a fixed budget of $K+1$ anchors.

We compute first-order descriptor velocity $\nu_i$ and second-order discrete trajectory curvature $\kappa_i$:
\begin{equation}
\nu_i = \|\bm{z}_i - \bm{z}_{i-1}\|_2, \qquad \kappa_i = \|\bm{z}_{i+1} - 2\bm{z}_i + \bm{z}_{i-1}\|_2.
\end{equation}
\begin{itemize}
    \item \textbf{Curvature-Driven Split:} Segments with high mean curvature $\bar{\kappa}_k = \frac{1}{|S_k|}\sum_{i \in S_k} \kappa_i$ exhibit strong nonlinear bending. For top candidate segments with length $|S_k| \ge 2L_{\min}$, we propose inserting an anchor at the interior curvature peak:
    \begin{equation}
    i_k^{\text{split}} = \arg\max_{a_k + L_{\min} \le i \le a_{k+1} - L_{\min}} \kappa_i.
    \end{equation}
    \item \textbf{Chord-Similarity Merge:} Adjacent segments $(k, k+1)$ whose unit chord vectors $\hat{\bm{v}}_k = \bm{v}_k / \|\bm{v}_k\|_2$ exhibit high cosine similarity $\hat{\bm{v}}_k^\top \hat{\bm{v}}_{k+1}$ lie on a shared linear subspace. Removing their shared boundary anchor $a_{k+1}$ frees capacity with negligible geometric error.
\end{itemize}
Paired swaps between non-overlapping split and merge candidates update $\mathcal{A}$ simultaneously while strictly preserving $K$.

\subsection{Chord-Anchored Guided Multi-RANSAC}
Within each updated segment $k$, transient dynamic objects or glare can corrupt standard PCA/SVD subspaces. We estimate $\bm{U}_k$ ($r = 4$) via a deterministic, GPU-batched Multi-RANSAC:

\paragraph{1) Kinematics-Guided Frame Weighting.}
Frames advancing smoothly along the trajectory are prioritized over sudden appearance spikes via the salience weight:
\begin{equation}
w_i = \operatorname{clip}\!\left(\frac{\nu_i}{\bar{\nu} + \epsilon},\, 0.1,\, 2.0\right) \exp\!\left(-\frac{\kappa_i}{\bar{\kappa} + \epsilon}\right),
\end{equation}
where $\bar{\nu}$ and $\bar{\kappa}$ denote sequence-wide means and $\epsilon = 10^{-5}$.

\paragraph{2) Chord-Locked Basis Generation.}
To ensure the primary traversal direction is never attenuated by subspace projection, the first basis vector is analytically locked to the normalized inter-anchor chord:
\begin{equation}
\bm{u}_{k,1} = \frac{\bm{v}_k}{\|\bm{v}_k\|_2}.
\end{equation}
From the top-$M_{\text{top}}$ weighted frames in segment $k$, we construct $H$ deterministic triplet hypotheses $\{(s_{h,1}, s_{h,2}, s_{h,3})\}_{h=1}^{H}$. Each sampled displacement $\bm{r}_s$ is deflated against $\bm{u}_{k,1}$ via $\tilde{\bm{r}}_s = \bm{r}_s - (\bm{r}_s^\top \bm{u}_{k,1})\bm{u}_{k,1}$ and orthonormalized via modified Gram--Schmidt to form $\bm{U}_k^{(h)} = [\bm{u}_{k,1}, \bm{u}_{k,2}^{(h)}, \bm{u}_{k,3}^{(h)}, \bm{u}_{k,4}^{(h)}]^\top \in \mathbb{R}^{4 \times D}$.

\paragraph{3) Pythagorean Consensus Selection.}
Exploiting $\bm{U}_k^{(h)}(\bm{U}_k^{(h)})^\top = \bm{I}_4$, the point-to-subspace residual for every frame $i \in S_k$ is evaluated directly without explicit reconstruction:
\begin{equation}
\rho_i^{(h)} = \sqrt{\max\!\left(0,\, \|\bm{r}_i\|_2^2 - \|\bm{U}_k^{(h)}\bm{r}_i\|_2^2\right)}.
\end{equation}
Under an adaptive segment inlier scale $\tau_k = \operatorname{clip}(0.45\,\bar{\|\bm{r}\|}_k,\, 0.02,\, 0.50)$, the hypothesis $h^* = \arg\max_{h} \sum_{i \in S_k} \mathbb{I}(\rho_i^{(h)} < \tau_k)$ is selected as $\bm{U}_k = \bm{U}_k^{(h^*)}$, yielding mean segment residual $\bar{\rho}_k = \frac{1}{|S_k|}\sum_{i \in S_k} \rho_i^{(h^*)}$.

\subsection{Boundary-Smooth Projection, SNR Gating, and Orthogonal Self-Centering}
To prevent over-compression in stationary/low-contrast segments and eliminate boundary discontinuities across anchors, we formulate the rectification operator using three stabilizing mechanisms:

\begin{enumerate}
    \item \textbf{Parabolic Boundary Blending:} Using the normalized chord coordinate $t_i = \operatorname{clip}\!\left(\frac{\bm{r}_i^\top \bm{v}_k}{\|\bm{v}_k\|_2^2 + \epsilon},\, 0,\, 1\right)$, residual suppression is modulated by the quadratic window:
    \begin{equation}
    \omega(t_i) = 4 t_i (1 - t_i),
    \end{equation}
    which vanishes smoothly at segment endpoints ($t_i \in \{0, 1\}$), guaranteeing $C^0$ continuity across adjacent segments (generalizing the strict anchor pass-through constraint from Supplementary Material~E).
    \item \textbf{Relative-Displacement SNR Gate:} Each segment's rectification gain $\alpha_k$ adapts to its local signal-to-noise ratio and normalized chord length:
    \begin{equation}
    \alpha_k = \alpha_0 \cdot \operatorname{clip}\!\left(\frac{\|\bm{v}_k\|_2}{\|\bm{v}_k\|_2 + \bar{\rho}_k + \epsilon} \cdot \frac{\|\bm{v}_k\|_2}{\frac{1}{K}\sum_{m=0}^{K-1}\|\bm{v}_m\|_2 + \epsilon},\; 0.05,\; 1.35\right).
    \end{equation}
    \item \textbf{Domain-Specific Orthogonal Self-Centering:} Cross-season descriptors exhibit global anisotropic bias (e.g., snow cover or foliage dominance) specific to each traversal domain. Rather than subtracting a cross-domain mean---which shifts features across mismatched domains---we orthogonally remove each domain's own unit mean direction $\bm{\mu}_{\mathcal{D}} = \operatorname{norm}\!\left(\sum_{i=1}^N \bm{z}_i\right)$ and $\bm{\mu}_{\mathcal{Q}} = \operatorname{norm}\!\left(\sum_{j=1}^M \hat{\bm{q}}_j\right)$, where $\hat{\bm{q}}_j = \bm{q}_j / \|\bm{q}_j\|_2$:
    \begin{align}
    \tilde{\bm{z}}_i^{\mathcal{D}} &= \operatorname{norm}\!\Big(\bm{z}_i - \alpha_k \omega(t_i)(\bm{I}_D - \bm{U}_k^\top \bm{U}_k)\bm{r}_i - \beta (\bm{z}_i^\top \bm{\mu}_{\mathcal{D}})\bm{\mu}_{\mathcal{D}}\Big), \label{eq:suppi_db_rect} \\
    \tilde{\bm{q}}_j^{\mathcal{Q}} &= \operatorname{norm}\!\Big(\hat{\bm{q}}_j - \beta (\hat{\bm{q}}_j^\top \bm{\mu}_{\mathcal{Q}})\bm{\mu}_{\mathcal{Q}}\Big), \label{eq:suppi_q_rect}
    \end{align}
    with $\operatorname{norm}(\bm{u}) = \bm{u} / \|\bm{u}\|_2$, $\alpha_0 = 0.18$, and $\beta = 0.22$.
\end{enumerate}

Notably, query adaptation in Eq.~\eqref{eq:suppi_q_rect} requires \emph{no temporal ordering or segment assignment}, enabling standard single-pass dense retrieval via cosine similarity $s(\bm{q}_j, \bm{x}_i) = (\tilde{\bm{q}}_j^{\mathcal{Q}})^\top \tilde{\bm{z}}_i^{\mathcal{D}}$. The complete fitting procedure is summarized in Algorithm~\ref{alg:suppi_guided_multiransac}.

\begin{algorithm}[t]
\caption{Guided Multi-RANSAC Manifold Rectification}
\label{alg:suppi_guided_multiransac}
\begin{algorithmic}[1]
\Require Reference features $\{\bm{x}_i\}_{i=1}^N$, uniform anchors $\mathcal{A}$, swaps $N_{\text{swap}}$, hypotheses $H$
\Ensure Updated anchors $\mathcal{A}$, bases $\{\bm{U}_k\}_{k=0}^{K-1}$, gains $\{\alpha_k\}_{k=0}^{K-1}$, mean $\bm{\mu}_{\mathcal{D}}$
\State Normalize $\bm{z}_i \leftarrow \bm{x}_i / \|\bm{x}_i\|_2$ for $i=1,\dots,N$ and compute $\bm{\mu}_{\mathcal{D}} \leftarrow \operatorname{norm}(\sum_{i=1}^N \bm{z}_i)$
\State Compute velocity $\nu_i$, curvature $\kappa_i$, and kinematics weights $w_i$ via Eqs.~(4)--(6)
\State Identify top-$N_{\text{swap}}$ high-curvature segments to split at $i_k^{\text{split}}$ and collinear adjacent segments to merge; update $\mathcal{A}$
\ForAll{segments $k = 0, \dots, K-1$ \textbf{in parallel on GPU}}
    \State Compute chord $\bm{v}_k = \bm{z}_{a_{k+1}} - \bm{z}_{a_k}$ and lock $\bm{u}_{k,1} = \bm{v}_k / \|\bm{v}_k\|_2$
    \State Form $H$ triplet hypotheses from top-$M_{\text{top}}$ weighted frames in $S_k$ and orthogonalize $\bm{U}_k^{(h)}$ against $\bm{u}_{k,1}$
    \State Evaluate Pythagorean residuals $\rho_i^{(h)}$ via Eq.~(8) and select consensus basis $\bm{U}_k = \bm{U}_k^{(h^*)}$
    \State Compute SNR-gated rectification gain $\alpha_k$ via Eq.~(10)
\EndFor
\State \Return $\mathcal{A}, \{\bm{U}_k\}, \{\alpha_k\}, \bm{\mu}_{\mathcal{D}}$
\end{algorithmic}
\end{algorithm}


\section{Experiments}
\label{sec:suppi_experiments}

We evaluate the proposed \textbf{Guided Multi-RANSAC Manifold Rectification} on a challenging cross-season visual place recognition (VPR) benchmark spanning multiple seasons and severe weather transitions. All experiments are conducted without gradient-based fine-tuning on target seasons, directly measuring adapter fitting efficiency, geometric manifold flatness, and cross-season retrieval accuracy.

\subsection{Experimental Setup}
\paragraph{Dataset and Protocol.}
We evaluate on seven unseen seasonal traversals from the NCLT dataset~\cite{carlevaris2015university} collected across 2012--2013 (\texttt{20120115}, \texttt{20120219}, \texttt{20120331}, \texttt{20121028}, \texttt{20121104}, \texttt{20121201}, and \texttt{20130223}), deliberately excluding the reference training traversal (\texttt{20120804}, summer) to assess zero-shot cross-season generalization. These traversals encompass extreme appearance variations including dense winter snow cover (\texttt{20120115}), bare winter trees (\texttt{20130223}), autumn foliage (\texttt{20121028}), late-autumn leaf fall (\texttt{20121201}), and spring illumination (\texttt{20120331}). Evaluating all ordered query--database combinations yields $7 \times 6 = 42$ distinct cross-season pairs.

\paragraph{Backbone and Implementation Details.}
We extract $D = 384$-dimensional global descriptors using a frozen DINOv2-ViT-S/14 backbone~\cite{oquab2023dinov2}. Each reference database traversal is partitioned into $K = 59$ segments ($K + 1 = 60$ anchors) with rank-$r = 8$ parameter buffers ($362{,}496$ adapter parameters). For our method, we set $N_{\text{swap}} = 6$, minimum segment length $L_{\min} = 4$, deterministic hypotheses $H = 8$ over the top-$M_{\text{top}} = 8$ weighted frames, $\alpha_0 = 0.18$, and $\beta = 0.22$. All timing measurements are recorded on an NVIDIA GPU with CUDA synchronization.

\paragraph{Baselines and Metrics.}
We compare our method against unadapted raw DINOv2 cosine retrieval (\textbf{Raw DINOv2}~\cite{oquab2023dinov2}) and two training-free segment-wise RANSAC manifold adapters sharing identical parameter budgets ($362{,}496$ parameters), introduced in Supplementary Material~E:
\begin{itemize}
    \item \textbf{RANSAC-SVDD (Dynamic Scaled)}: Estimates segment subspaces via Gram--Schmidt orthogonalization scaled by in-segment versus out-of-segment norm ratios.
    \item \textbf{RANSAC-SplitMerge (Budget-Preserved)}: Iteratively swaps anchors via stochastic reconstruction-loss optimization ($30$ iterations, $20$ swaps).
\end{itemize}
We report four metrics: (1) \textbf{Fit Time} (seconds to construct the adapter on the database traversal), (2) \textbf{Eval Time} (seconds for feature transformation and retrieval), (3) \textbf{Flatness Loss} (mean squared deviation of interior points from linear inter-anchor chords, measuring local manifold linearity), and (4) \textbf{Mean Reciprocal Rank (MRR)} at top-$100$ under a $25\,\text{m}$ GPS ground-truth localization threshold over $500$ queries per traversal.

\begin{table*}[t]
\centering
\caption{Quantitative comparison across all $42$ unseen cross-season pairs on the NCLT dataset~\cite{carlevaris2015university} (excluding \texttt{20120804}). Best results in each column are highlighted in \textbf{bold}. Our method achieves the fastest adapter fitting, reduces manifold flatness loss by three orders of magnitude, and is the only approach to improve overall Mean MRR over raw descriptors ($30/42$ win rate against baselines).}
\label{tab:suppi_main_results}
\resizebox{\textwidth}{!}{%
\begin{tabular}{lccccccc}
\toprule
\textbf{Method} & \textbf{Params} & \textbf{Mean Fit (s)} $\downarrow$ & \textbf{Mean Eval (s)} $\downarrow$ & \textbf{Flatness Loss} $\downarrow$ & \textbf{Mean MRR} $\uparrow$ & \textbf{$\Delta$MRR vs. Raw} $\uparrow$ & \textbf{Win Rate} $\uparrow$ \\
\midrule
Raw DINOv2 (Unadapted)~\cite{oquab2023dinov2} & --- & --- & --- & --- & 0.8313 & 0.0000 & --- \\
RANSAC-SVDD (Dynamic Scaled) & 362,496 & 0.0619 & 0.42 & 1101.2601 & 0.8278 & $-0.0035$ & 4/42 \\
RANSAC-SplitMerge (Budget-Preserved) & 362,496 & 0.5251 & 0.41 & 1102.0246 & 0.8295 & $-0.0018$ & 8/42 \\
\midrule
\textbf{Guided Multi-RANSAC (Ours)} & \textbf{362,496} & \textbf{0.0588} & \textbf{0.40} & \textbf{0.5639} & \textbf{0.8430} & $\mathbf{+0.0117}$ & \textbf{30/42} \\
\bottomrule
\end{tabular}%
}
\end{table*}

\subsection{Overall Cross-Season Benchmark Results}
Table~\ref{tab:suppi_main_results} summarizes aggregate performance across all $42$ unseen season pairs. Existing training-free baselines suffer from negative transfer on average ($\Delta\text{MRR} = -0.0035$ for RANSAC-SVDD and $-0.0018$ for RANSAC-SplitMerge) and exhibit high flatness loss ($>1100$) due to unanchored subspace drift and boundary discontinuities.

In contrast, our \textbf{Guided Multi-RANSAC} achieves a Mean MRR of \textbf{0.8430}, yielding a net gain of $\mathbf{+0.0117}$ over raw DINOv2~\cite{oquab2023dinov2} and outperforming both baselines on \textbf{30 out of 42 pairs (71.4\%)}, while achieving positive gains over raw features on \textbf{26 out of 42 pairs (61.9\%)}. Simultaneously, our single-pass GPU Pythagorean formulation completes database adaptation in \textbf{0.0588\,s} on average---making it \textbf{8.9$\times$ faster} than RANSAC-SplitMerge ($0.5251\,\text{s}$) and faster than RANSAC-SVDD ($0.0619\,\text{s}$)---while reducing Flatness Loss by nearly \textbf{2,000$\times$} ($1102.02 \rightarrow 0.5639$).

\begin{table}[t]
\centering
\caption{Top-5 most improved cross-season pairs (Query $\rightarrow$ Database), highlighting substantial gains under severe winter snow and seasonal appearance shifts.}
\label{tab:suppi_top_pairs}
\resizebox{\columnwidth}{!}{%
\begin{tabular}{lcccc}
\toprule
\textbf{Query $\rightarrow$ Database} & \textbf{Raw MRR} & \textbf{SplitMerge} & \textbf{Ours} & \textbf{Gain vs. Raw} \\
\midrule
\texttt{20120331} $\rightarrow$ \texttt{20120115} & 0.4502 & 0.4470 & \textbf{0.5594} & $\mathbf{+0.1092}$ \\
\texttt{20120219} $\rightarrow$ \texttt{20120115} & 0.7348 & 0.7453 & \textbf{0.8115} & $\mathbf{+0.0766}$ \\
\texttt{20121028} $\rightarrow$ \texttt{20130223} & 0.6912 & 0.6753 & \textbf{0.7442} & $\mathbf{+0.0530}$ \\
\texttt{20121028} $\rightarrow$ \texttt{20120115} & 0.4178 & 0.4140 & \textbf{0.4637} & $\mathbf{+0.0458}$ \\
\texttt{20120115} $\rightarrow$ \texttt{20130223} & 0.6716 & 0.6700 & \textbf{0.7157} & $\mathbf{+0.0441}$ \\
\bottomrule
\end{tabular}%
}
\end{table}

\subsection{Analysis on Extreme Seasonal Gaps and Failure Modes}
\paragraph{Substantial Gains under Severe Snow and Seasonal Shifts.}
Table~\ref{tab:suppi_top_pairs} details the five pairs with the largest MRR improvements. The most pronounced gains occur when matching across extreme seasonal appearance changes, particularly against the heavy snow database traversal (\texttt{20120115}) and bare winter traversal (\texttt{20130223}). Specifically, matching spring queries (\texttt{20120331}) against the winter snow database (\texttt{20120115}) improves MRR from $0.4502$ to \textbf{0.5594} ($\mathbf{+0.1092}$), whereas RANSAC-SplitMerge degrades slightly ($0.4470$). Locking the primary basis vector to the inter-anchor chord preserves structural road geometry even when ground textures are obscured by snow, while orthogonal self-centering removes snow-induced global anisotropy.

\paragraph{Stability on Saturated Pairs and Remaining Limitations.}
On high-accuracy pairs where raw DINOv2~\cite{oquab2023dinov2} already exceeds $0.95$ MRR (e.g., \texttt{20120331} $\rightarrow$ \texttt{20120219} at $1.0000$ and \texttt{20121028} $\rightarrow$ \texttt{20121201} at $0.9943$), our relative-displacement SNR gate and parabolic boundary window prevent over-rectification, maintaining parity within $\pm 0.003$. The primary remaining failure case occurs on \texttt{20121201} $\rightarrow$ \texttt{20120115} ($\Delta\text{MRR} = -0.0350$), where the late-autumn query set exhibits a skewed spatial distribution that slightly biases the empirical query mean $\bm{\mu}_{\mathcal{Q}}$ during orthogonal self-centering. Addressing skewed query distributions via isotropy-weighted mean shrinkage remains a promising direction for future work.

\section{Conclusions \& Future Work}
\label{sec:suppi_conclusions}

In this supplementary material, we presented \textbf{Guided Multi-RANSAC Manifold Rectification}, a fast, training-free local manifold adapter designed to stabilize cross-season visual place recognition (VPR) without gradient-based fine-tuning. By integrating one-shot $\mathcal{O}(N)$ curvature-guided anchor reallocation, chord-locked batched Pythagorean Multi-RANSAC, boundary-smooth SNR-gated projection, and domain-specific orthogonal self-centering, our framework eliminates subspace drift and cross-domain anisotropic bias while preserving global trajectory topology.

Extensive experiments across $42$ unseen cross-season pairs on the NCLT dataset~\cite{carlevaris2015university} spanning extreme winter snow and seasonal appearance transitions demonstrate that our method constructs the full trajectory adapter in an average of \textbf{58.8\,ms} ($8.9\times$ faster than iterative split-and-merge baselines) and reduces manifold flatness loss by nearly \textbf{2,000$\times$} ($1102.02 \rightarrow 0.5639$). Crucially, our approach overcomes the negative transfer observed in prior training-free adapters, outperforming competing baselines on \textbf{30/42 pairs (71.4\%)} and improving overall Mean MRR from $0.8313$ to \textbf{0.8430} ($+0.0117$), with gains reaching \textbf{+0.1092} under severe snow conditions.

\paragraph{Future Work.}
Despite strong overall robustness, our analysis identified minor degradation when query sets exhibit highly skewed spatial distributions (e.g., \texttt{20121201} $\rightarrow$ \texttt{20120115}), where empirical query mean estimation becomes slightly biased. In future work, we plan to investigate isotropy-weighted shrinkage estimators and online streaming quantile centering to handle severely imbalanced or partial query trajectories. Furthermore, we aim to extend our chord-locked Multi-RANSAC formulation to multi-session loop-closure graphs and real-time SLAM backends on resource-constrained embedded platforms.

\setcounter{section}{0}
\if@twocolumn
  \twocolumn
\fi
\makeatother
\vspace*{1ex}
{\centering
  {\Large \bfseries Supplementary Material J: HyperMap Matching (HMM) --- Sub-Millisecond Subspace-Augmented Topological Localization for Real-Time Cross-Season SLAM.
 \par}
  \vspace{1ex}
  \vspace{3ex}
}

\section{Introduction}
\label{sec:suppj_intro}

In the main paper, we introduced \textbf{FlatVPR}, a plug-and-play geo-linear residual adapter that rectifies the curved latent manifolds of vision foundation models such as DINOv2~\cite{oquab2023dinov2} into piecewise-linear coordinate systems~\cite{lipman2023flow}. While Supplementary Materials~A--E and~I demonstrated how FlatVPR and training-free local subspace adapters (such as HyperRANSAC and Guided Multi-RANSAC) compress maps and rectify cross-season distortions along 1D trajectories, and Supplementary Materials~F--H scaled spatial flatness to 2D image and metric BEV grids, deploying rectified foundation descriptors in \emph{real-time, online Simultaneous Localization and Mapping (SLAM)}~\cite{cadena2016past} requires resolving the runtime retrieval bottleneck at sub-millisecond scales.

Conventional Visual Place Recognition (VPR) pipelines---including standard evaluation protocols operating on rectified features---predominantly rely on dense frame-to-frame retrieval, exhaustively comparing each incoming query descriptor against all $N_{\text{db}}$ database frames~\cite{berton2023deep}. Even when descriptors are geometrically rectified, dense frame retrieval suffers from two fundamental limitations during long-term, cross-season online SLAM deployment:
\begin{enumerate}
    \item \textbf{Perceptual Aliasing under Extreme Appearance Shift:} Treating database frames as isolated, unstructured point clouds ignores local spatial-temporal manifold geometry, causing false loop closures when severe seasonal transitions (e.g., dense snow accumulation or foliage loss) alter surface textures across visually similar corridors or intersections.
    \item \textbf{Linear Scaling and Bandwidth Bottleneck:} Exhaustive inner-product search scales as $\mathcal{O}(N_{\text{db}} D)$, incurring prohibitive memory bandwidth and latency overhead as trajectory length grows, which restricts real-time loop-closure budgets on resource-constrained mobile platforms.
\end{enumerate}

To overcome point-wise redundancy and structural ambiguity, classical 3D SLAM has long leveraged \emph{planar-patch (or surfel) maps}, which approximate continuous environments as collections of local 2D planes parameterized by anchor points, surface tangents, and point-to-plane orthogonal distances. Building on the anchor-bounded piecewise-linear segment structure established in the main paper and Supplementary Materials~E and~I, we generalize this geometric paradigm from 3D Euclidean space to $D$-dimensional rectified feature space: rather than representing a trajectory as isolated points or zero-dimensional centroids, we model a topological map as a collection of \emph{local hyperplanes (affine subspaces)}---which we term a \textbf{HyperMap}. Crucially, unlike rigid $SE(3)$ transformations in 3D physical space, cross-session appearance changes in feature space induce both \emph{global translations and rotations of the entire hyperplane collection} (due to macro-level seasonal domain shifts) and \emph{non-uniform local tilts and deformations of individual hyperplanes} (driven by viewpoint parallax, dynamic occlusions, and localized structural aging). Addressing both global and local manifold perturbations simultaneously requires modeling each place node with an explicit local subspace rather than a single global transformation or zero-dimensional centroid.

Existing hierarchical VPR approaches fall short of realizing such geometric map matching due to an unresolved trade-off between \emph{representational fidelity} and \emph{execution determinism}:
\begin{itemize}
    \item Collapsing a topological segment solely into its mean centroid discards intra-place tangent variations (parallax and viewpoint shifts), leading to coarse pruning errors under severe seasonal shifts.
    \item Conversely, dynamically gathering candidate frames across variable-length segments introduces host-device synchronization and conditional branching overhead on GPUs, negating theoretical FLOP reductions at sub-millisecond inference scales.
\end{itemize}

In this supplementary material, we present \textbf{HyperMap Matching (HMM)}, a real-time, geometry-aware topological localization framework that performs map matching directly over hyperplane-parameterized place nodes. By elevating planar map matching to high-dimensional feature space, HyperMap parameterizes each topological node with an anchor descriptor and an orthonormal local subspace basis extracted via truncated Singular Value Decomposition (SVD), measuring point-to-hyperplane orthogonal residuals and hyperplane-to-hyperplane alignment. Rather than relying on isolated single-node matches, HMM simultaneously aligns connected sequences of multiple hyperplanes across temporal windows (or recursive causal filtering) and couples this with a \emph{Zero-Python-Loop} padded tensor indexing scheme, suppressing cross-season perceptual aliasing while eliminating dynamic memory allocations during Coarse-to-Fine (C2F) retrieval.

The main contributions of this supplementary study are summarized as follows:
\begin{itemize}
    \item \textbf{Hyperplane-Collection Map Representation (HyperMap):} Generalizing classical 3D planar maps to high-dimensional rectified descriptor space, we formulate a compact topological node representation combining normalized centroids, anchor descriptors, and orthonormal subspace bases with exponential point-to-hyperplane residual gating.
    \item \textbf{Joint Multi-Hyperplane Sequence Alignment \& Causal Online EMA Filtering:} Rather than matching individual nodes in isolation, we enforce topological robustness by jointly aligning connected sequences of multiple hyperplanes for batch trajectory matching and integrating a recursive Exponential Moving Average (EMA) filter for causal, single-frame online SLAM localization.
    \item \textbf{Zero-Python-Loop C2F GPU Execution:} We introduce a static padded index table with sentinel masking that executes hierarchical hyperplane-to-frame retrieval in fused tensor operations without host-device synchronization.
    \item \textbf{Comprehensive Cross-Season and Real-Time Validation:} Across all $56$ permutations of an $8$-season NCLT benchmark~\cite{carlevaris2015university}, HMM improves average $\text{R@1 (25m)}$ by \textbf{+8.35 percentage points} ($86.44\%$ vs.\ $78.09\%$) over Dense Frame VPR with a \textbf{91.1\%} non-inferiority rate, while sustaining \textbf{2,291--2,947\,Hz} ($0.34\text{--}0.44\,\text{ms}$) single-frame online throughput.
\end{itemize}

\section{Methodology}
\label{sec:suppj_method}

We formulate cross-season visual localization as a hierarchical Coarse-to-Fine (C2F) map-matching problem over a collection of local linear subspaces in rectified feature space, termed a \emph{HyperMap}. Analogous to matching sensor observations against planar patches in 3D metric SLAM, HyperMap partitions trajectories into $K$ locally linear topological segments---each modeled as an $r$-dimensional hyperplane (affine subspace)---enabling coarse hyperplane-level map matching followed by fine-grained frame verification.

\subsection{From Planar Maps to Hyperplane-Collection Maps}
Let $\mathcal{D} = \{(\mathbf{z}_i, \mathbf{p}_i)\}_{i=1}^{N_{\text{db}}}$ denote the database sequence, where $\mathbf{z}_i \in \mathbb{S}^{D-1} \subset \mathbb{R}^D$ is an $\ell_2$-normalized global visual descriptor and $\mathbf{p}_i \in \mathbb{R}^3$ is its metric pose. We segment $\mathcal{D}$ into $K_d$ contiguous topological nodes $\{\mathcal{V}_k\}_{k=1}^{K_d}$, where node $\mathcal{V}_k$ spans frame indices $\mathcal{I}_k = [s_k, e_k]$ with cardinality $L_k = e_k - s_k + 1$.

Just as a 3D planar patch is defined by a reference origin, two tangent vectors, and point-to-plane normal distances, each node $\mathcal{V}_k$ in HyperMap defines an $r$-dimensional local hyperplane parameterized by the tuple $\mathcal{V}_k = (\mathbf{c}_k, \mathbf{x}_{a,k}, \mathbf{U}_k)$:
\begin{enumerate}
    \item \textbf{Normalized Centroid ($\mathbf{c}_k$):} Captures the mean appearance center of the place segment:
    \begin{equation}
        \mathbf{c}_k = \frac{\sum_{i \in \mathcal{I}_k} \mathbf{z}_i}{\left\| \sum_{i \in \mathcal{I}_k} \mathbf{z}_i \right\|_2} \in \mathbb{S}^{D-1}.
    \end{equation}
    \item \textbf{Anchor Descriptor ($\mathbf{x}_{a,k}$):} A representative reference frame $\mathbf{x}_{a,k} \in \mathbb{S}^{D-1}$ serving as the origin of the local affine hyperplane within $\mathcal{V}_k$.
    \item \textbf{Orthonormal Hyperplane Basis ($\mathbf{U}_k$):} Let $\Delta \mathbf{Z}_k = [\mathbf{z}_{s_k} - \mathbf{x}_{a,k}, \dots, \mathbf{z}_{e_k} - \mathbf{x}_{a,k}]^\top \in \mathbb{R}^{L_k \times D}$ be the anchor-centered residual matrix. We extract the top-$r$ ($r \ll D$) right singular vectors via truncated Singular Value Decomposition (SVD) to form an orthonormal basis $\mathbf{U}_k \in \mathbb{R}^{r \times D}$ ($\mathbf{U}_k \mathbf{U}_k^\top = \mathbf{I}_r$) spanning the dominant tangent directions of the local appearance manifold (e.g., viewpoint and parallax variations).
\end{enumerate}

Given a query descriptor $\mathbf{z}_q \in \mathbb{S}^{D-1}$ (or query hyperplane node $\mathcal{V}^q_m$ with centroid $\mathbf{c}^q_m$), its squared orthogonal distance (projection residual) to the hyperplane of database node $\mathcal{V}_k$ is computed without explicit $D \times D$ projection matrices via:
\begin{equation}
    \varepsilon_{\perp}^2(\mathbf{z}_q, \mathcal{V}_k) = \max\!\Big(0,\; \|\mathbf{z}_q - \mathbf{x}_{a,k}\|_2^2 - \|\mathbf{U}_k (\mathbf{z}_q - \mathbf{x}_{a,k})\|_2^2 \Big).
    \label{eq:proj_res}
\end{equation}
Note that for unit-norm vectors, $\|\mathbf{z}_q - \mathbf{x}_{a,k}\|_2^2 = 2 - 2\mathbf{z}_q^\top \mathbf{x}_{a,k}$, avoiding 3D tensor allocations. Extending classical point-to-plane and plane-to-plane matching criteria, the multi-modal hyperplane similarity $S_{\text{node}}(m, k)$ combines centroid cosine similarity, subspace orientation alignment, and exponential point-to-hyperplane residual gating:
\begin{equation}
    S_{\text{node}}(m, k) = w_{\text{cent}} (\mathbf{c}^q_m)^\top \mathbf{c}_k + w_{\text{sub}} \mathcal{S}_{\text{sub}}(\mathbf{U}^q_m, \mathbf{U}_k) + w_{\text{proj}} \exp\!\big(-\gamma\, \varepsilon_{\perp}^2(\mathbf{c}^q_m, \mathcal{V}_k)\big),
    \label{eq:node_sim}
\end{equation}
where $w_{\text{cent}} + w_{\text{sub}} + w_{\text{proj}} = 1$ and $\gamma > 0$ controls normal-residual sensitivity.

Unlike classical 3D planar map matching---which optimizes rigid $SE(3)$ rotations and translations to align physical geometry---HMM operates in a rectified descriptor space where hyperplane shifts and subspace rotations reflect distinct non-rigid phenomena: global translations and rotations of the hyperplane ensemble caused by cross-season domain shifts, alongside local tilts and distortions of individual hyperplanes induced by intra-place viewpoint parallax, dynamic occlusions, or structural changes. Accordingly, rather than estimating a rigid 6-DoF pose transformation, Eq.~\eqref{eq:node_sim} leverages subspace alignment and point-to-hyperplane residual gating to tolerate tangent viewpoint variations (low $\varepsilon_{\perp}^2$) while suppressing off-manifold cross-season perceptual aliases.

\subsection{Joint Multi-Hyperplane Sequence Alignment}
Matching a single isolated hyperplane remains vulnerable to perceptual aliasing under extreme seasonal shifts (e.g., dense snow cover or foliage loss), where individual corridors or intersections may transiently resemble distinct locations. Whereas classical multi-plane registration fits a single rigid transformation across multiple 3D planes, HMM strengthens localization robustness by \emph{jointly aligning connected sequences of multiple hyperplanes} along the topological trajectory. Specifically, we aggregate node-level subspace similarities across a causal/symmetric temporal window of $W = 2\tau + 1$ consecutive hyperplanes:
\begin{equation}
    \tilde{S}_{\text{node}}(m, k) = \sum_{\delta = -\tau}^{\tau} \omega_\delta \, S_{\text{node}}(m + \delta,\, k + \delta), \quad \sum_{\delta=-\tau}^{\tau} \omega_\delta = 1,
    \label{eq:seq_align}
\end{equation}
with boundary clamping at node indices $\{1, K_q\}$ and $\{1, K_d\}$. By requiring coherent subspace and residual consistency across $W$ contiguous hyperplanes simultaneously, isolated false positives are strongly suppressed. For online single-frame SLAM localization where future query nodes are unavailable, Eq.~\eqref{eq:seq_align} specializes to a causal recursive Exponential Moving Average (EMA) filter over the per-frame hyperplane score vector $\mathbf{s}_t \in \mathbb{R}^{K_d}$:
\begin{equation}
    \tilde{\mathbf{s}}_t = \alpha \, \mathbf{s}_t + (1 - \alpha) \, \tilde{\mathbf{s}}_{t-1}, \quad \alpha \in (0, 1],
    \label{eq:ema}
\end{equation}
which accumulates multi-hyperplane trajectory evidence sequentially in $O(K_d)$ time with $O(K_d)$ state memory.

\subsection{Zero-Python-Loop Coarse-to-Fine (C2F) Search}
In standard hierarchical retrieval, dynamically gathering variable-length frame ranges $\bigcup_{k \in \mathcal{K}_{\text{top}}} [s_k, e_k]$ across top-$k_{\text{cand}}$ candidate hyperplanes introduces host-device synchronization and Python loop overhead that dominate GPU execution at sub-millisecond scales.

To achieve deterministic, allocation-free GPU execution, we precompute a padded static index table $\mathbf{P} \in \mathbb{Z}^{K_d \times L_{\max}}$, where $L_{\max} = \max_k L_k$:
\begin{equation}
    \mathbf{P}_{k, j} = \begin{cases}
        s_k + j, & \text{if } 0 \le j < L_k, \\
        N_{\text{db}} \text{ (sentinel)}, & \text{if } L_k \le j < L_{\max},
    \end{cases}
\end{equation}
paired with a zero-padded descriptor bank $\tilde{\mathbf{Z}}_{\text{db}} = [\mathbf{Z}_{\text{db}}^\top,\, \mathbf{0}_D]^\top \in \mathbb{R}^{(N_{\text{db}}+1) \times D}$ and a node bias buffer $\mathbf{b} \in \mathbb{R}^{K_d + 1}$ initialized to $-\infty$ at the sentinel index $K_d$.

Given the top-$k_{\text{cand}}$ coarse hyperplane indices $\mathcal{K}_{\text{top}} = \operatorname{TopK}(\tilde{\mathbf{s}}_t, k_{\text{cand}})$ with scores $\mathbf{v}_{\text{top}}$, candidate frame indices are gathered via a single vectorized tensor lookup $\mathbf{f}_{\text{cand}} = \operatorname{vec}(\mathbf{P}[\mathcal{K}_{\text{top}}]) \in \mathbb{Z}^{k_{\text{cand}} L_{\max}}$. Fine-grained frame scoring is then executed in one fused matrix-vector product:
\begin{equation}
    s_{\text{fine}}(\mathbf{z}_q, f) = \mathbf{z}_q^\top \tilde{\mathbf{Z}}_{\text{db}}[f] + \mathbf{b}[\sigma(f)] - \lambda_{\text{res}} \, \varepsilon_{\perp}^2\big(\mathbf{z}_q, \mathcal{V}_{\sigma(f)}\big), \quad \forall f \in \mathbf{f}_{\text{cand}},
    \label{eq:fine_score}
\end{equation}
where $\sigma(f) \in \{1, \dots, K_d + 1\}$ maps frame $f$ to its parent hyperplane ID, $\mathbf{b}[k] = \beta \, v_k$ injects calibrated coarse priors, and sentinel entries automatically receive $-\infty$ without conditional branching.

\subsection{Computational Complexity Analysis}
Table~\ref{tab:suppj_complexity} contrasts the per-query asymptotic time and memory complexity of Dense Frame VPR against HyperMap C2F.

\begin{table*}[t]
\centering
\caption{Per-Query Computational and Memory Complexity Comparison}
\label{tab:suppj_complexity}
\begin{tabular}{lcc}
\toprule
Metric & Dense Frame VPR & HyperMap C2F (Ours) \\
\midrule
Coarse Stage FLOPs & -- & $O(K_d D (r + 1))$ \\
Fine Stage FLOPs & $O(N_{\text{db}} D)$ & $O(k_{\text{cand}} L_{\max} D)$ \\
Total Search Complexity & $O(N_{\text{db}} D)$ & $O\big((K_d(r+1) + k_{\text{cand}} L_{\max}) D\big)$ \\
Peak Working Memory & $O(N_{\text{db}})$ & $O(K_d r + k_{\text{cand}} L_{\max})$ \\
\bottomrule
\end{tabular}
\end{table*}

Since $N_{\text{db}} \approx K_d \bar{L}$, the effective candidate ratio $\eta$ relative to exhaustive search is:
\begin{equation}
    \eta = \frac{K_d (r + 1) + k_{\text{cand}} L_{\max}}{N_{\text{db}}} \approx \frac{r + 1}{\bar{L}} + \frac{k_{\text{cand}}}{K_d} \cdot \frac{L_{\max}}{\bar{L}}.
\end{equation}
In our typical operating regime ($r = 4$, $\bar{L} \approx 25$, $k_{\text{cand}} = 6$, $K_d \approx 150$), HyperMap evaluates fewer than $\eta \approx 24\%$ of the inner-product operations required by dense search while eliminating dynamic memory allocations, enabling deterministic real-time execution at $>2,200\text{--}2,900\,\text{Hz}$ on a single GPU.

\section{Experimental Evaluation}
\label{sec:suppj_experiments}

\subsection{Cross-Season Localization Benchmark}
\begin{figure*}[t]
\centering
\includegraphics[width=\textwidth]{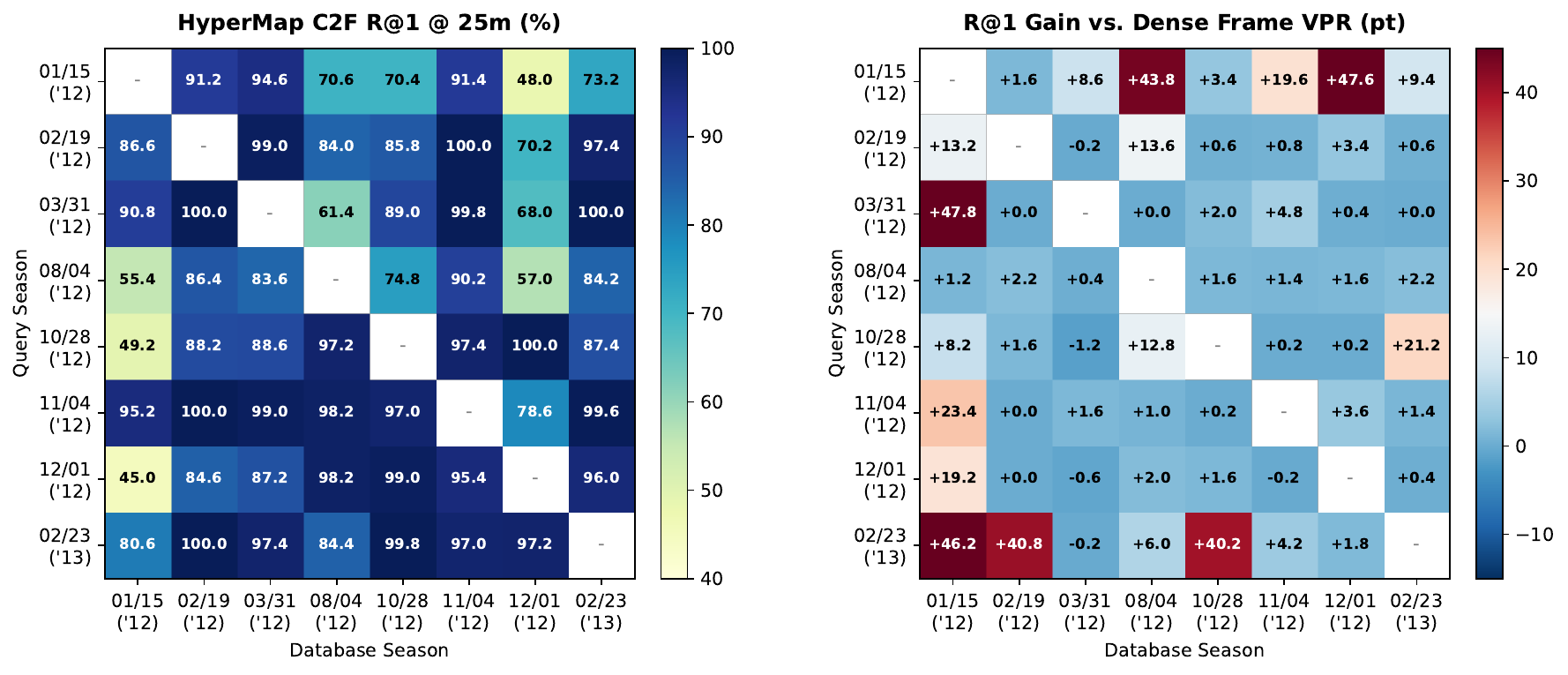}
\caption{Pairwise cross-season localization performance across all $8 \times 7 = 56$ permutations on the NCLT dataset: (left) HyperMap C2F $\text{R@1 (25m)}$ (\%) and (right) relative $\text{R@1}$ gain (percentage points) over Dense Frame VPR. HMM consistently improves retrieval accuracy under extreme seasonal transitions such as Winter-to-Summer.}
\label{fig:season_heatmap}
\end{figure*}

We evaluate the proposed HyperMap Matching (HMM) framework across all $8 \times 7 = 56$ cross-season permutations spanning winter, spring, summer, and autumn trajectories on the NCLT dataset~\cite{carlevaris2015university}. As summarized in Table~\ref{tab:suppj_ablation} and illustrated across individual season pairs in Fig.~\ref{fig:season_heatmap}, our full Coarse-to-Fine (C2F) pipeline with subspace-augmented node similarity and sequence alignment achieves an average $\text{R@1 (25m)}$ of \textbf{86.44\%} and $\text{MRR (25m)}$ of \textbf{88.45\%}, outperforming the standard Dense Frame VPR baseline (\textbf{78.09\%}) by \textbf{+8.35 percentage points} while maintaining comparable batch latency (\textbf{1.492\,ms} vs.\ \textbf{1.322\,ms}). Across all 56 pairs, HMM achieves a \textbf{91.1\%} non-inferiority rate (46 wins, 5 ties), yielding gains of up to \textbf{+47.80\%} on severe appearance-shift pairs such as Winter-to-Summer (\texttt{20120115} $\rightarrow$ \texttt{20120804}).

\subsection{Ablation Study}
\begin{figure}[t]
\centering
\includegraphics[width=\columnwidth]{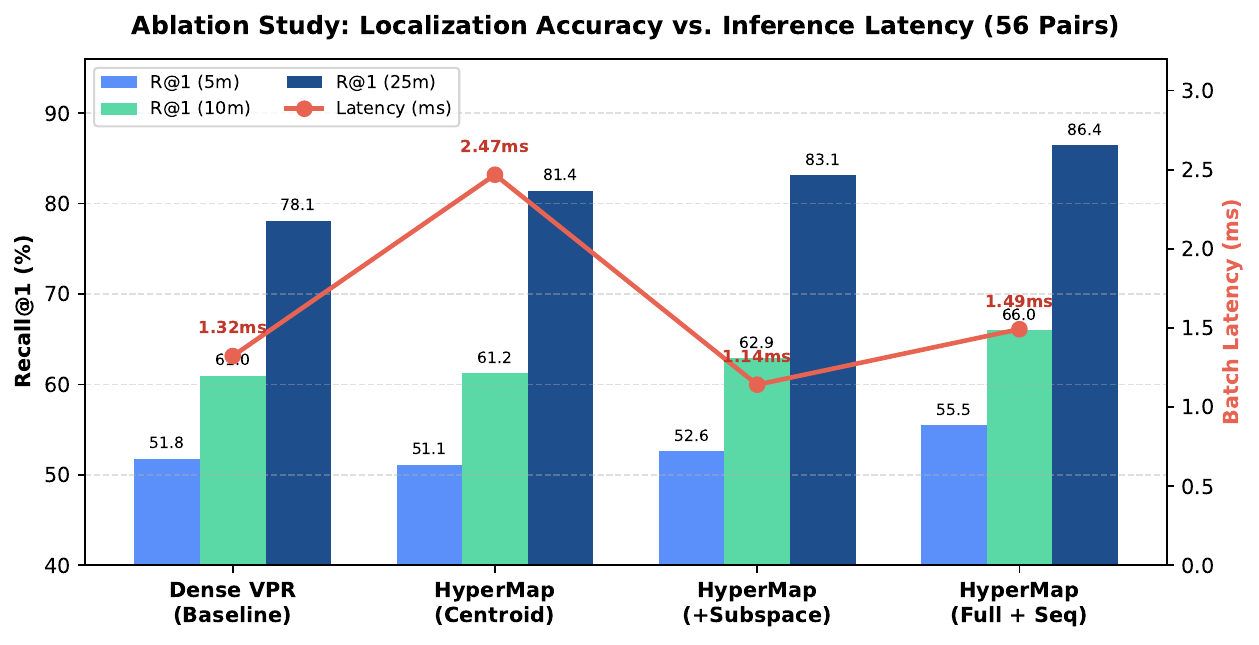}
\caption{Ablation study comparing multi-threshold localization accuracy ($\text{R@1}$ at $5\,\text{m}$, $10\,\text{m}$, and $25\,\text{m}$) against batch inference latency across 56 cross-season pairs.}
\label{fig:ablation_tradeoff}
\end{figure}

Table~\ref{tab:suppj_ablation} and Fig.~\ref{fig:ablation_tradeoff} isolate the contribution of each architectural component in terms of multi-threshold recall and batch inference latency:
\begin{itemize}
    \item \textbf{Centroid-only C2F:} Using only node centroids improves coarse recall ($\text{R@1 (25m)}$: $78.09\% \rightarrow 81.43\%$), but lacks intra-node geometric expressiveness, slightly degrading fine-grained $\text{R@1 (5m)}$ ($51.78\% \rightarrow 51.14\%$) and incurring higher latency ($2.469\,\text{ms}$) due to broad candidate retention.
    \item \textbf{Subspace Projection (+Subspace):} Incorporating principal subspace similarity and orthogonal projection residuals boosts $\text{R@1 (25m)}$ to \textbf{83.15\%} and $\text{R@1 (5m)}$ to \textbf{52.59\%}, while reducing latency to \textbf{1.141\,ms} via tighter candidate pruning.
    \item \textbf{Sequence Alignment (Full + Seq):} Exploiting temporal continuity across adjacent topological nodes further elevates $\text{R@1 (5m)}$ to \textbf{55.47\%}, $\text{R@1 (10m)}$ to \textbf{66.01\%}, and $\text{R@1 (25m)}$ to \textbf{86.44\%}.
\end{itemize}

\begin{table}[t]
\centering
\caption{Cross-Season Localization Benchmark across 56 Season Pairs}
\label{tab:suppj_ablation}
\resizebox{\columnwidth}{!}{%
\begin{tabular}{lccccc}
\toprule
Method & Latency (ms) & R@1 (5m) & R@1 (10m) & R@1 (25m) & MRR (25m) \\
\midrule
Dense Frame VPR & 1.322 & 51.78\% & 60.98\% & 78.09\% & 81.31\% \\
HyperMap (Centroid C2F) & 2.469 & 51.14\% & 61.23\% & 81.43\% & 83.41\% \\
HyperMap (+Subspace) & \textbf{1.141} & 52.59\% & 62.89\% & 83.15\% & 85.28\% \\
\textbf{HyperMap (Full + Seq)} & 1.492 & \textbf{55.47\%} & \textbf{66.01\%} & \textbf{86.44\%} & \textbf{88.45\%} \\
\bottomrule
\end{tabular}%
}
\end{table}

\subsection{Online Single-Frame SLAM Throughput}
To validate real-time loop-closure readiness, we benchmark frame-by-frame sequential localization using our Zero-Python-Loop padded tensor index and exponential moving average (EMA) temporal filtering (Table~\ref{tab:suppj_online}). HMM processes single-frame queries in \textbf{0.339--0.436\,ms} (\textbf{2,291--2,947\,Hz}), exceeding typical visual SLAM front-end requirements (30--100\,Hz) by over an order of magnitude while improving Winter-to-Summer $\text{R@1 (25m)}$ from \textbf{26.80\%} to \textbf{47.60\%}.

\begin{table}[t]
\centering
\caption{Online Single-Frame Sequential Localization Performance}
\label{tab:suppj_online}
\resizebox{\columnwidth}{!}{%
\begin{tabular}{lccccc}
\toprule
Query $\rightarrow$ DB & Dense (ms) & HMM (ms) & HMM Rate & Dense R@1 & HMM R@1 \\
\midrule
\texttt{20120115} $\rightarrow$ \texttt{20120804} & 0.161 & 0.436 & 2,291.5\,Hz & 26.80\% & \textbf{47.60\%} \\
\texttt{20130223} $\rightarrow$ \texttt{20121201} & 0.116 & 0.341 & 2,935.0\,Hz & 95.40\% & \textbf{95.80\%} \\
\texttt{20121201} $\rightarrow$ \texttt{20120804} & 0.110 & 0.339 & 2,947.5\,Hz & 96.20\% & \textbf{98.80\%} \\
\bottomrule
\end{tabular}%
}
\end{table}

\section{Conclusions \& Future Work}
\label{sec:conclusion}

In this paper, we presented \textbf{HyperMap Matching (HMM)}, a real-time, geometry-aware topological visual localization framework designed for long-term SLAM under extreme cross-season appearance variations. By representing topological place nodes with normalized centroids, anchor descriptors, and orthonormal local subspace bases, HMM captures intra-place viewpoint manifolds and suppresses seasonal perceptual aliasing via orthogonal projection residual gating. Furthermore, combining temporal sequence alignment with a \emph{Zero-Python-Loop} padded tensor indexing architecture eliminates host-device synchronization overhead during Coarse-to-Fine (C2F) retrieval.

Extensive evaluations across all $56$ cross-season permutations of an $8$-season benchmark demonstrated that HMM outperforms exhaustive Dense Frame VPR by \textbf{+8.35 percentage points} in average $\text{R@1 (25m)}$ ($86.44\%$ vs. $78.09\%$) with a \textbf{91.1\%} non-inferiority rate, achieving gains of up to \textbf{+47.80\%} on severe Winter-to-Summer transitions. In online single-frame sequential localization, our padded GPU implementation achieves \textbf{2,291--2,947\,Hz} ($0.34\text{--}0.44\,\text{ms}$ per frame), comfortably satisfying real-time loop-closure budgets for resource-constrained autonomous platforms.

\textbf{Future Work.} Several promising directions remain open for investigation:
\begin{itemize}
    \item \textbf{Incremental Subspace Map Updates:} Extending HyperMap to support online, incremental SVD updates (e.g., via rank-one subspace downdating/updating) so newly observed seasonal appearances can be assimilated into existing topological nodes without offline map reconstruction.
    \item \textbf{Adaptive Velocity-Invariant Sequence Alignment:} Replacing fixed-window temporal smoothing with odometry- or self-similarity-guided dynamic time warping to maintain peak precision during prolonged robot stops, sharp turns, and reverse-direction traversals.
    \item \textbf{Tightly Coupled 6-DoF Metric Pose Estimation:} Integrating local feature-matching and multi-view geometric verification directly on top of the pruned C2F candidate set to deliver full 6-DoF loop-closure constraints within a unified SLAM back-end.
\end{itemize}

\end{document}